\documentclass[10pt]{article}
\usepackage{nexilume-report}
\usepackage{longtable}
\usepackage{url}
\usepackage{placeins}
\reporttitle{TokenBank: Financial Infrastructure for AI Services}
[]
\reportmeta{NX-2026-002}{v1}{September 2026}{Nexilume Research}{research@nexilume.com}{Research Manuscript}
\newcommand{\system}{\textsc{TokenBank}}

\hypersetup{pdftitle={TokenBank: Building Financial Markets for AI Services},pdfauthor={Nexilume Research}}
\begin{document}
\makereporttitle{%
AI services incur inference costs during execution, while the revenue needed to cover those costs may arrive later. Changing API prices, limited upfront capital, and service failures can therefore constrain an operator's ability to sustain or expand a service. Reducing per-request costs addresses only part of this problem: operators also need to plan future spending, fund execution before revenue arrives, and obtain compensation for specified losses. Supporting these needs requires clear agreements across services with different pricing and execution conditions. These agreements must distinguish the right to consume a service from the right to receive payments, define obligations despite uncertain costs and income, and specify which failures qualify for compensation and how much can be paid. We present \system{}, a financial infrastructure that represents these commitments through structured contracts. It supports service-consumption rights, agreements that settle API-price differences in cash (forwards), financing through limited rights to future service revenue, and protection claims for specified service failures. Each contract records its participants, covered services, validity period, ownership, fulfillment conditions, and settlement rules. We evaluate \system{} through replay of 899,441 API requests, real model-driven agent execution, and contract API tests. In a zero-discount rising-price resampling scenario, forwards reduce mean expenditure by USD~304.88, while its standard deviation increases from USD~1,152.45 to USD~1,190.82. A controlled replication with five portfolios per capital condition finds mean contribution differences between financing and self-funding of +1.0635, -0.1406, and -0.2962 experimental USD under low, baseline, and ample capital, respectively. The evaluation distinguishes contract correctness from economic effectiveness under declared economic and failure assumptions; supplier invoices and commercial revenue are unavailable.\footnote[1]{https://github.com/Nexilume-AI/nexus-cloud}\footnote[2]{https://cloud.nexilume.com}
}

\section{Introduction}

Large language models are increasingly delivered through APIs and embedded in Agents, applications, and automated workflows. These systems may call multiple Providers, switch among models, and accumulate substantial inference costs over long-running tasks. As their use expands, model inference is becoming an economic resource as well as a computational one: Agent operators purchase API capacity, Providers monetize model access, and applications turn model calls into downstream revenue. Managing the associated cost, financing, revenue, and service risk is therefore becoming increasingly important.

Most existing AI infrastructure, however, focuses on making model execution cheaper and more reliable. Caching and batching reduce repeated computation, model routers select lower-cost models, and retries or failover recover failed requests. Systems such as FrugalGPT and RouteLLM show that better model selection and request execution can substantially reduce inference cost \cite{frugalgpt2023,routellm2024}. Yet these techniques mainly optimize \emph{how} requests are executed. They do not specify how heterogeneous AI services should be contracted, how future API spending can be hedged, how future Agent or Provider revenue can finance current usage, or how losses from unrecovered service failures should be shared.

These questions become increasingly important as Agents move from short-lived demonstrations to continuously running services. \textbf{First, AI services are not interchangeable.} Different providers offer different models, capabilities, prices, latency, reliability, and usage conditions. Therefore, a contract for ``AI service'' must specify what service is promised, by whom, for what period, and under what conditions it is considered delivered. \textbf{Second, future API cost is uncertain.} An Agent may know that it will need a large amount of inference next month, but it does not know what that inference will ultimately cost. \textbf{Third, payment and revenue often occur at different times.} An Agent may need to pay API bills before receiving money from its own users. If part of its future revenue could be contractually promised to a financier, that future income could potentially support today's API expenditure. \textbf{Finally, API failures can still create monetary losses after retries and failover have been attempted.} A failed request may lead to a refund, a lost customer transaction, or other business loss. Technical recovery determines whether the request can still be completed; a separate mechanism is needed to determine who pays for the loss when it cannot.

Financial mechanisms for computing services already have substantial
precedents. Service agreements describe what providers promise, cloud
forward contracts manage future price risk, revenue-based financing
exchanges current funding for future payments, and outage protection
compensates eligible failures
\citep{wsagreement2007,dynamicforward2013,clarke2025revenue,mastroeni2019sla}.
These mechanisms provide a starting point for AI-service markets.
Our question is how to specify their different obligations and connect
their payments to the services, prices, revenue, and failures they concern.
This connection matters because the same operating workload affects
several contracts: changing models changes API expenditure, available
funding affects how much demand can be served, and recovery changes
the losses remaining after a failure.

\begin{figure*}[t]
    \centering
    \includegraphics[width=\textwidth]{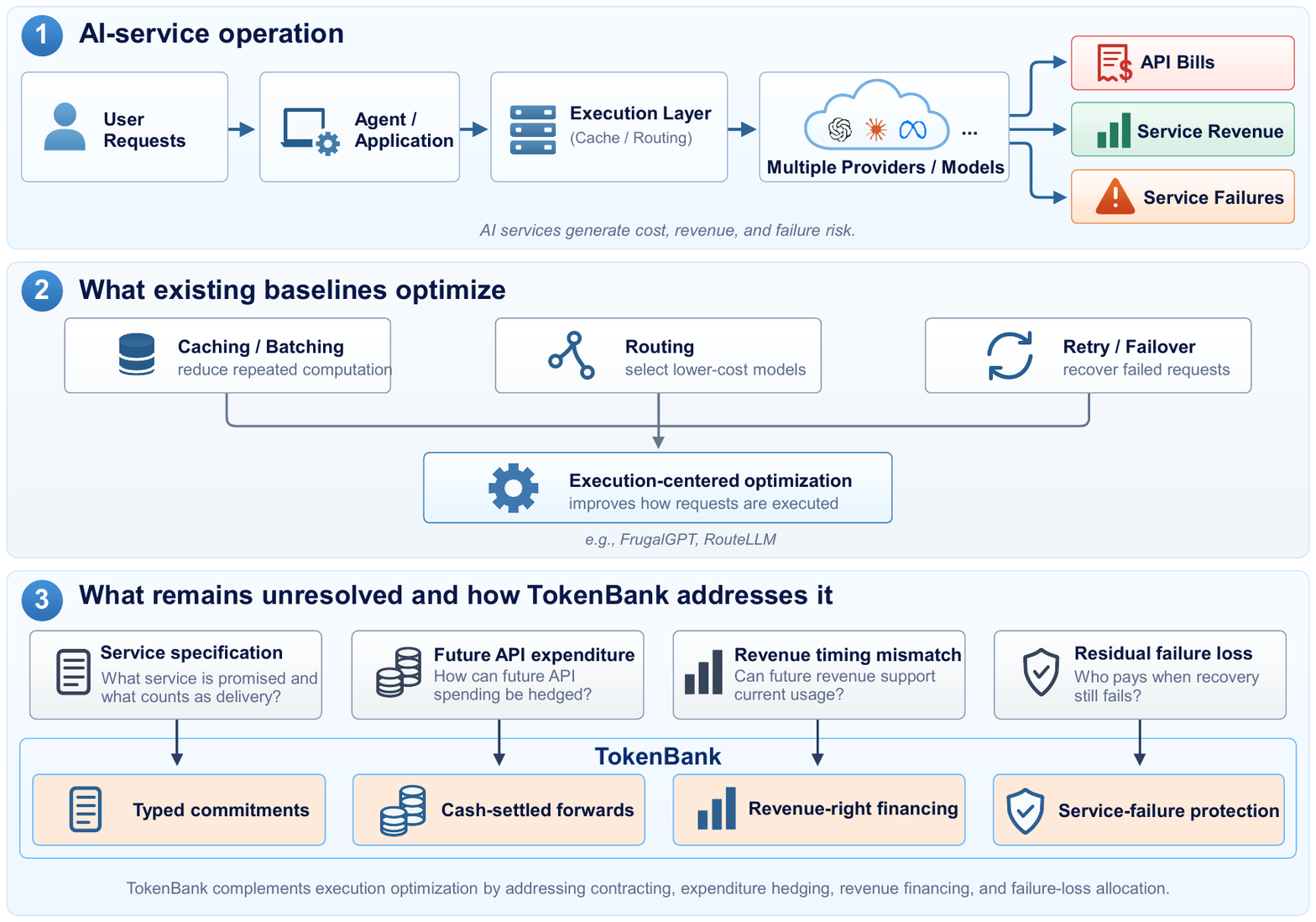}
    \caption{
        Caching, batching, model routing, and failover improve request
        execution. \system{} complements these approaches with service
        commitments that specify what is purchased, API-price forwards,
        revenue-right financing, and protection against covered service
        failures. These mechanisms address the financial consequences
        of AI-service use rather than replacing execution optimization.
    }
    \label{fig:introduction}
\end{figure*}

Applying these mechanisms to AI services raises four design challenges. Figure~\ref{fig:introduction} summarizes how the four challenges
extend beyond request execution.
\textbf{First, contracts must distinguish what is being promised and
what evidence determines payment.} Access to a model, a payment based
on a price difference, a share of revenue, and compensation for failure
are different rights, even when they concern the same service.
\textbf{Second, price contracts must account for differences between
planned and actual API spending.} Changes in model selection or request
volume can leave a contract poorly matched to the eventual bill.
\textbf{Third, revenue financing must connect additional funding to
both recorded receipts and the income retained by each participant.}
Serving more requests is beneficial only when the additional earnings
outweigh the associated costs and revenue sharing.
\textbf{Fourth, protection must distinguish service recovery from
compensation.} A contract must specify which remaining failures qualify,
how much is payable, and which funds are available to cover the payment.

We present \system{}, a financial market infrastructure for AI services that addresses these four problems. \textbf{First, to describe precisely what is being bought or sold,} \system{} defines typed service commitments. Each commitment records the service being promised, the parties involved, the period in which it is valid, the conditions for successful delivery, and how payment is determined. Depending on the product, such commitments can be created through direct agreements, requests for quotation, public listings, or order matching. \textbf{Second, to reduce uncertainty in future API spending,} \system{} supports cash-settled forward contracts. An operator can choose how much future API expenditure to cover, agree on a reference price in advance, pay the associated fee, and settle the price difference in cash when the contract expires. \textbf{Third, to connect future revenue with current financing,} \system{} supports revenue-right contracts, through which an Agent or Provider can promise a defined share of future revenue in exchange for capital today. These contracts grant rights only to the specified revenue and do not represent ownership of the underlying company, model, or service. \textbf{Fourth, to cover losses that remain after technical recovery,} \system{} provides SLA-like protection contracts. These contracts specify which service failures are eligible for compensation, how much loss is covered, how much the buyer pays for protection, and how reserves are used to make payments. Across these products, \system{} records ownership, contract status, service and revenue events, and the resulting transfers of money, providing a common basis for issuing, transferring, and settling AI-service contracts.

We evaluate \system{} through request-level API replay, real Qwen-driven
Agent execution, and actual contract APIs. The separate real-model groups
comprise a ten-trial retail baseline, 70 retail intervention cases, 84 initial
controlled cases, 624 controlled repetition and boundary cases, and a prospectively fixed 540-case financing-capital replication. Their
workloads and denominators remain distinct. \textbf{For service commitments,}
we verify ownership, fulfillment and settlement, including an observed
rounding-carry defect and its subsequent correction. \textbf{For forwards,}
paired price and demand scenarios identify favorable and unfavorable
expenditure and risk outcomes; settlement is real, but future indices remain
assumed. \textbf{For revenue-right financing,} repeated portfolios and freshly
executed capital, receipt and payment-delay boundaries show that additional
capital can enable service, while revenue sharing can outweigh that growth.
\textbf{For service protection,} local recovery changes completed service;
protection redistributes eligible remaining loss subject to premiums,
coverage and funded reserves. The shared upstream limits the recovery claim,
and commercial prices, customer receipts and credit redemption remain
explicit assumptions. The unified evaluation therefore reports both useful
operating conditions and adverse outcomes.

This paper makes the following contributions:

\begin{itemize}

\item \textbf{Contractible AI-service claims.}
We introduce a typed representation that makes heterogeneous AI services explicit economic claims with defined participants, service attributes, validity periods, fulfillment conditions, and settlement rules. The representation distinguishes rights to consume AI services from forward contracts, revenue rights, and service-protection claims, providing a common contractual foundation for their issuance, transfer, and settlement.

\item \textbf{Hedging future API expenditure.}
We design and implement cash-settled forward contracts that allow Agent operators to hedge a specified portion of future API-price exposure. Our request-level replay evaluates how hedge ratio, price movements, contract fees, and basis mismatch affect realized expenditure, showing when forward contracts reduce cost and when the hedge becomes unfavorable.

\item \textbf{Revenue-right financing for AI services.}
We design contractual revenue rights that allow Agents and Providers to exchange a bounded share of future revenue for current capital, without representing these rights as ownership of the underlying model, service, or company. Our economic replay evaluates how additional capital, served demand, and revenue sharing jointly affect participant contribution profit, identifying when financing-enabled growth is sufficient to offset the reduction in retained revenue.

\item \textbf{Financial protection against service failures.}
We design SLA-like protection contracts that compensate eligible losses remaining after technical recovery, using explicit failure conditions, coverage limits, premiums, and dedicated reserves. By evaluating failover and financial protection separately, we show that failover improves request recovery, whereas protection determines how the remaining economic loss is distributed among participants.

\end{itemize}

\section{Related Work}
Reducing the cost of executing a task and deciding how its payments and losses
are shared are related but distinct problems. FrugalGPT and RouteLLM address
the former by selecting sequences of models or routing requests between
stronger and cheaper models \citep{frugalgpt2023,routellm2024}. Their savings
come from changing which model calls are made. Financial contracts can instead
change an operator's payments even when those calls remain unchanged. This
distinction connects \system{} to research on service agreements, cloud
procurement, revenue financing, and outage compensation.

\subsection{Service Agreements and Agent Markets}

A service market must distinguish choosing a provider from specifying what that
provider promises. WS-Agreement already provides a common structure for service
terms and guarantees, but deliberately leaves application-specific measurements
to the systems using it \citep{wsagreement2007}. A common contract format can
therefore describe a quality requirement without establishing how that quality
will be verified. Agent Exchange studies the complementary allocation problem:
using capability descriptions and bids to select agents and distribute task
rewards \citep{agentexchange2025}. Descriptions useful for selecting an agent,
however, are not by themselves evidence that a promised result was delivered.
For example, the number of tokens generated can establish billed usage without
establishing the correctness of an answer.

Connecting payment to delivery is also an explicit research problem. The
Agentic Settlement Protocol proposes holding a buyer's funds until delivery is
confirmed, with defined rules for releasing payments and handling refunds
\citep{asp2026}. \system{} addresses a different contract-design question: the
same AI service can support several agreements that require different evidence.
Access to a model is checked against service terms; a payment based on a future
price depends on a price observation; revenue sharing depends on recorded
receipts; and compensation depends on an eligible failure. Its proposed
contract representation distinguishes these obligations and their payment
rules, rather than treating every transaction as the purchase of interchangeable
tokens or requiring a single matching mechanism.

\subsection{Cloud Procurement and Future API Spending}

Cloud-finance research shows that lower purchasing costs and protection against
price changes need not come from the same mechanism. In the brokerage model of
Rogers and Cliff, an intermediary purchases discounted long-term reservations
and resells shorter contracts \citep{cloudbroker2012}. The potential gain comes
from the gap between purchasing and resale terms, but the broker must still pay
for reservations that customers do not use. Cartlidge and Clamp show how
correcting reservation accounting and incorporating competing purchasing
options removes the modeled commercial opportunity under the conditions they
study \citep{brokerreplication2014}. Dynamic forward pricing addresses a
different question: what future prices a provider and customer will accept
given uncertain demand and their willingness to bear risk
\citep{dynamicforward2013}.

These studies distinguish the source of a discount from the value of sharing
risk. For AI APIs, a further issue is whether a contract follows the costs an
operator actually incurs: changing models or serving fewer requests can leave
the contracted price or quantity poorly matched to the eventual bill.
\system{} uses cash-settled forwards: the parties settle the difference
between an agreed contract price and a specified API reference price observed
later, multiplied by the covered amount. These contracts transfer money rather
than deliver reserved inference capacity. Its replay evaluates specified
contract terms; it does not infer the prices that independent buyers and sellers would agree
on. Accordingly, savings must be compared under consistent execution and
purchasing conditions, with fees and payments by the other party included.
A lower buyer bill alone does not establish that a market has created
additional value.

\subsection{Revenue Sharing and Financing}

Earning revenue, dividing it, and using it to fund current work are different
operations. Poe's monetization API connects charges to service delivery by
authorizing costs before processing and collecting payment afterward
\citep{poemonetization2026}. Revenue-sharing research asks how dividing those
receipts changes business decisions. Cachon and Lariviere show that, under
specified conditions, sharing can lead a retailer to choose prices and purchase quantities that maximize combined
supplier--retailer profit. This result can fail when sales depend on costly
effort by the retailer \citep{cachon2005}. Thus, a percentage split is not merely an
accounting rule: it changes how much each participant earns from additional
business and can change their incentive to generate it.

Financing adds a separate requirement: future revenue must be sufficiently
observable to support repayment. Clarke et al. study financing repaid from a
share of subsequent platform sales and find that firms can divert transactions
to other payment channels to delay repayment \citep{clarke2025revenue}. This
evidence distinguishes accurate distribution of recorded receipts from complete
observation of the revenue covered by a contract. \system{} applies defined
revenue shares to Agent and Provider receipts and examines whether additional
funding permits enough otherwise-unserved demand to offset the income shared
with financiers. Its analysis also distinguishes customer sales from API
payments passed from an Agent to a Provider, avoiding counting an internal
payment as a second external sale. These mechanisms support explicit revenue
contracts, while verifying revenue outside the platform and establishing that
financing causes demand growth remain separate questions.

\subsection{Service Failures and Financial Protection}

An outage, a breached service promise, and a customer's monetary loss are not
the same quantity. Amazon Bedrock's service-level agreement ties eligibility to
specified availability conditions and supporting request logs, while calculating
credits from service charges rather than the customer's downstream business
losses \citep{bedrocksla2026}. Consequently, a credit can follow the provider's
contract correctly without covering every loss caused by a failed application
transaction. Cloud-insurance research examines the other side of this promise:
Mastroeni et al. relate outage models and compensation rules to the price of
insurance and the amount of compensation a provider can sustainably offer
\citep{mastroeni2019sla}. Broader compensation therefore raises a funding
question, not just a question of detecting failures.

This distinction matters when protection is combined with retries and failover.
Recovery changes which requests still fail; protection changes how the remaining
losses are divided. If several services fail together, compensation obligations
can also accumulate together, so successful recovery in ordinary cases does not
establish that sufficient funds will be available during a larger incident.
\system{} compares the same recovery settings with and without financial
protection, reporting both the buyer's remaining loss and the balance of
funds set aside for compensation. Its implementation issues billing credits,
whose value depends on subsequent service use. This keeps request recovery,
contractual compensation, and the ability to fund that compensation distinct,
rather than treating a payment to the user as an improvement in availability.

Across these areas, the contracts depend on the same operating workload.
Changing models alters the bill a forward is intended to cover; admitting more
work changes the revenue to be shared; and recovery changes the losses left
for compensation. These dependencies motivate \system{}'s common contract and
accounting design: payments must be traced across products and participants,
rather than interpreting each module's buyer-side saving in isolation.

\section{Overview}
\begin{figure*}[t]
    \centering
    \includegraphics[width=\textwidth]{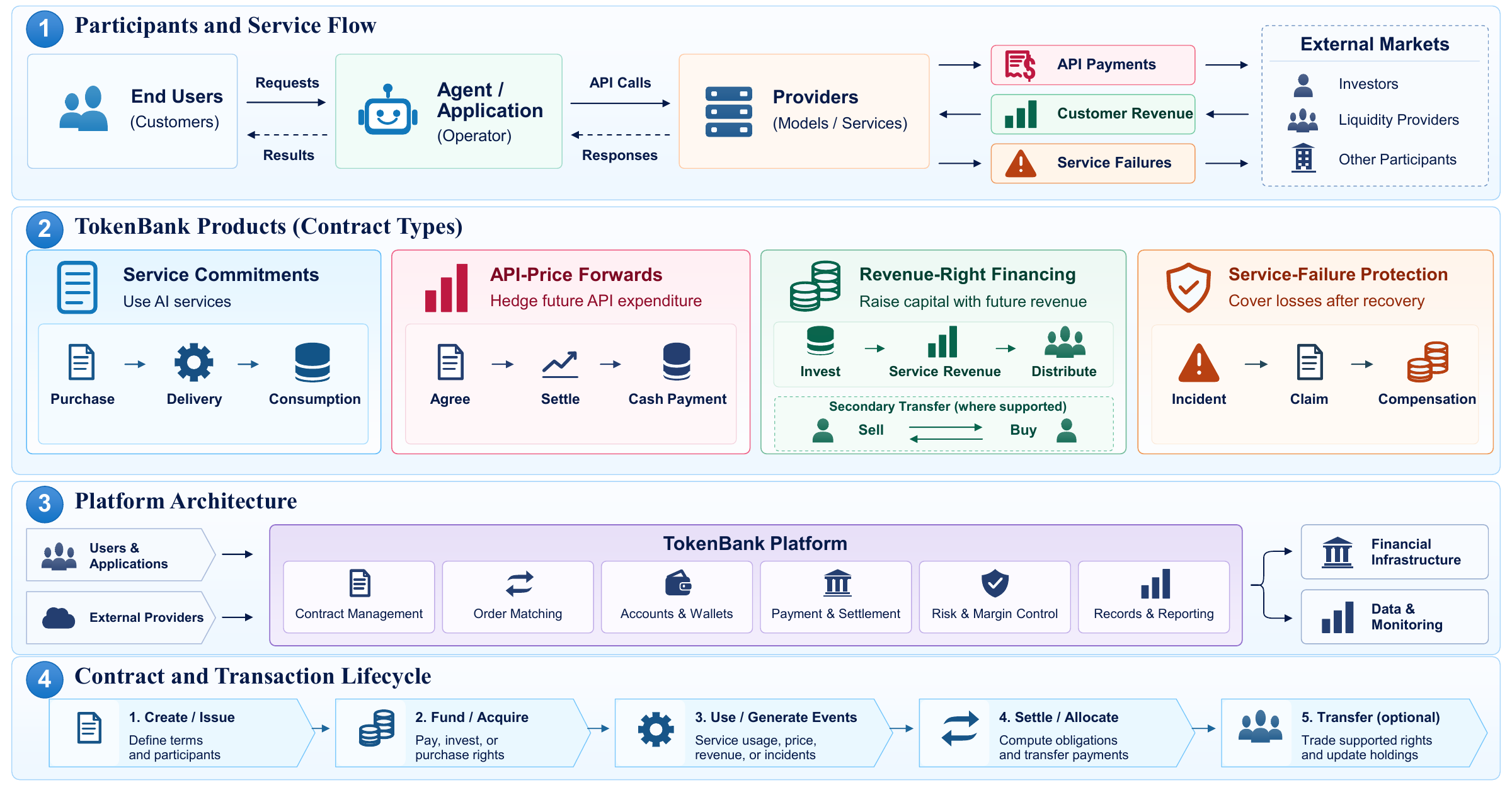}
    \caption{
        Overview of \system{}.
        Service commitments, API-price forwards, revenue-right
        financing and transfer, and service-failure protection use
        product-specific workflows supported by shared accounts,
        contract records, and settlement functions. Usage, price,
        revenue, and incident records provide the corresponding
        evidence for fulfillment and payment. Supported revenue-right
        transfers update the holders considered for subsequent
        distributions. The lifecycle summarizes available operations,
        not a mandatory sequence shared by all products.
    }
    \label{fig:overview}
\end{figure*}

An Agent operator may purchase API services before receiving customer
payments, face changing API prices, and incur losses during service
interruptions. These activities motivate the four challenges discussed
above: specifying what is purchased, managing future prices, financing
current usage, and compensating covered failures. \system{} implements
four corresponding components. Each product defines when money should move and
who should receive it. Figure~\ref{fig:overview} presents the participants, contract types,
and shared financial infrastructure of \system{}. 

\paragraph{Contractual representation.}
The first component distinguishes the rights obtained through different
transactions. A prepaid balance can be spent on API calls, whereas a
revenue right pays its holder a share of recorded service revenue.
Buying the latter does not itself provide API access. Similarly, a
forward promises a price-dependent payment rather than future service
delivery, while protection provides compensation only under specified
conditions.

These differences are represented through separate, linked records.
Prepaid records track purchased and remaining balances, expiration, and
refund eligibility. Preorder records connect purchases to subsequent
delivery. Forward contracts record pricing and payment terms; funding
agreements specify capital contributions and revenue shares; and
protection subscriptions identify the covered account and coverage period.
For transferable rights, holding records identify their owners and
distinguish available units from locked units, which are temporarily
unavailable for use or transfer. Thus, each product retains the records
and checks appropriate to what it promises, while account and transaction
references connect it to the shared financial system.

\paragraph{Model API-price forwards.}
The second component lets an operator arrange payments that offset
changes in a selected API reference price. The operator first requests
a quotation: a proposal specifying the fixed price and other contract
terms. Confirmation of a valid, unexpired quote is followed by the
required approval and activation steps. The contract identifies the
two parties, the reference price to observe, the fixed price, the
amount covered, the contract dates, and the fees.

When payment is due, the system compares the observed reference price
with the fixed price. If the observed price is higher, the seller pays
the buyer the difference multiplied by the covered amount; if it is
lower, the payment direction reverses. Funds held as margin---money
set aside to support this obligation---are applied before checking
whether the payer can cover the remaining payment. Successful settlement
records the transfer and releases unused margin; insufficient funding
produces a default record. The operator still purchases API services
separately. The contract neither supplies tokens nor reserves capacity,
so its usefulness depends on how closely its price and amount match
actual API spending.

\paragraph{Revenue-right financing.}
The third component allows a service publisher, meaning the operator
of an Agent or Provider service, to obtain funding in exchange for
a specified share of future revenue. An investment plan defines the
shares assigned to the publisher, investors, and platform. Funding
agreements record investor contributions, and the corresponding workflow
transfers capital to the publisher's account once funding and activation
conditions are satisfied.

Distribution begins when the system receives a revenue event: a record
of service income to be divided. For illustration, an investor assigned
10\% of a recorded \$100 revenue amount receives \$10. The other
participants receive their configured shares, with rounding handled
so that the allocations preserve the total amount. Provider distributions
use the net amount supplied in the revenue record.

The system also checks who is entitled to each investor share. When
a funding agreement has linked revenue-right assets, their holdings
determine the recipients; otherwise, the original investor remains the
recipient. This permits the specified revenue entitlement to follow
its holder without transferring ownership of the model or company.
Whether financing benefits the publisher depends on whether the
additional business it supports earns enough to offset the revenue shared.

\paragraph{Service-failure protection.}
The fourth component determines whether a service incident qualifies
for compensation and how that compensation is funded. A protection plan
specifies the covered Provider or model, eligible event types, coverage
periods, claim deadlines, and protection fees. It also defines the amount
left to the buyer before compensation applies, the compensation rate,
and the maximum payment.

Monitoring records identify incidents, and gateway logs show the requests
affected during each incident. The impact calculation includes failed
requests and requests that used a fallback service. Recorded costs, or
an amount derived from request counts when positive costs are unavailable,
provide the starting amount to which the plan's compensation rules apply.
For example, a request that succeeds through a backup Provider may still
contribute fallback activity to this calculation. Impact records can
generate pending claims, but payment additionally requires coverage
validation and approval.

Approved claims are paid from the plan's reserve, an account containing
funds set aside for compensation, subject to its budget and available
balance. The buyer receives service credit in its billing wallet rather
than a cash withdrawal. This workflow therefore compensates covered
incidents without performing retries or recovering requests itself.
Its log-based impact calculation is also distinct from the experimental
model of losses remaining after technical recovery.
\section{Contractual Representation}
\label{sec:representation}

An AI-service transaction must specify both what a participant acquires
and what establishes fulfillment. A prepaid balance and a right to future
revenue may use the same accounting unit, but only the former permits
eligible service consumption. \system{} therefore distinguishes service
consumption, price-dependent payments, revenue participation, and failure
compensation. We call the right or payment obligation established by each
product a \emph{contractual claim}. The representation connects its terms,
current state, and supporting records without treating the existence of
a right as evidence that it has already been fulfilled.

\subsection{Contract Types and Terms}
\label{sec:claim-principles}
\label{sec:typed-contract-model}

A claim's type determines what can be received. A consumption claim permits
use of a prepaid balance or delivery of a purchased service. A forward
specifies a payment based on a reference price, the recorded price selected
for the contract. A revenue right assigns part of specified service income,
while a protection subscription provides compensation under defined failure
conditions. These types share descriptive fields but retain different
fulfillment rules.

Each claim identifies the participants and their roles, the service or
agreement concerned, the units of its quantities and payments, and the
applicable terms. Roles are not reducible to a single owner: a service
purchase identifies a buyer and seller, a forward identifies two potential
payers or recipients, and revenue sharing identifies several recipients.
Service references specify the Provider and model, or the Agent whose
revenue is shared. A common representation identifies these services;
it does not establish that different models provide equivalent results.

Terms remain separate from current state. For a \emph{preorder}, a purchase
for later delivery, the purchased amount is a term and the delivered amount
is state. The same distinction separates a promised revenue share from a
recorded distribution and an approved compensation amount from its payment.
Time conditions also depend on the product: prepaid balances may expire,
forwards record contract dates, and protection defines coverage periods
and claim deadlines. Revenue policies, the rules identifying receipts and
their recipients, do not require a finite end date. Quantities retain their
units, so a service quantity, a transferable right, and a monetary payment
are not interchangeable merely because their numerical amounts coincide.

This common description is realized through linked product-specific
records, not a single contract table or a shared sequence of processing
steps. Appendix~\ref{app:claim-details} gives the complete record notation
and detailed validation rules.

\subsection{Evidence and Fulfillment}
\label{sec:claim-entitlements}

The same observation can have different consequences for different claims.
A usage record can establish consumption, whereas a revenue-sharing
payment requires recorded income. Let $\tau$ identify a claim's type,
$\Theta_{\tau}$ its terms, $x_{\tau}$ its current state, and $e$ the
supporting input or record. We summarize the product-specific rule as
\begin{equation}
\Phi_{\tau}(\Theta_{\tau},x_{\tau},e)=o_{\tau},
\label{eq:typed-fulfillment}
\end{equation}
where $o_{\tau}$ is the resulting allocation or payment obligation.
This notation describes how the product routines interpret evidence;
it is not a generic executable rule stored in every contract.

\paragraph{Consumption.}
Prepaid use requires sufficient valid balance after excluding amounts
locked against other use. A \emph{prepaid lot} records a credited amount
with its validity conditions; consumption draws from eligible lots and
reduces their remaining balances. For a preorder, delivery records increase
the fulfilled quantity without exceeding the purchased amount. These
checks establish recorded consumption or delivery, not the semantic
correctness of a model's output.

\paragraph{Price-dependent payment.}
A forward combines its agreed price and contractual amount with the
selected reference-price observation to determine which participant owes
payment and how much. This calculation establishes an obligation rather
than delivering API capacity. Payment is completed separately, subject
to sufficient funding.

\paragraph{Revenue participation.}
A \emph{revenue event} records service income submitted for distribution.
The applicable policy assigns each recipient a share of the event's
recorded net amount. For example, a 10\% share of a 100-unit distribution
base gives an exact entitlement of 10 units. Actual allocations use the
supported decimal precision and retain rounding differences for subsequent
processing. The right concerns the specified receipts, not ownership of
the Agent, model, or company.

\paragraph{Failure compensation.}
An \emph{incident} records a service disruption, and a compensation claim
requests payment under a protection subscription. Coverage checks relate
the incident to the protected service, coverage period, eligible event
type, and submission deadline. The plan determines compensation, review
authorizes an amount, and settlement checks the \emph{reserve}, the account
and budget set aside to fund approved claims. The resulting service credit
can pay service charges rather than provide a cash withdrawal. Coverage,
approval, and payment thus establish different stages of the obligation.

\subsection{State Changes and Transfer}
\label{sec:claim-lifecycle}
\label{sec:claim-transfer}
\begin{figure*}[t]
    \centering
    \includegraphics[width=\textwidth]{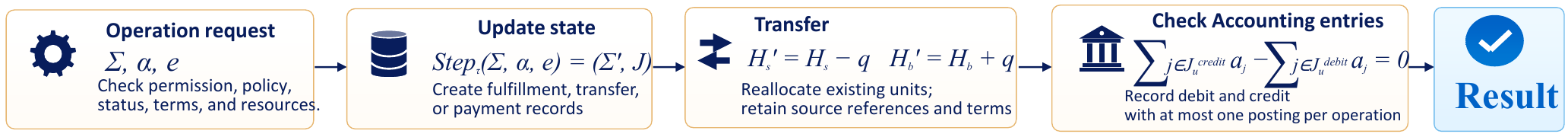}
    \caption{
    Supported transfers change
    ownership while preserving source references, and financial effects
    are recorded through balanced accounting entries.
    }
    \label{fig:contractual-representation}
\end{figure*}
An operation changes a claim only when its applicable conditions hold.
For requested operation $\alpha$, supporting input $e$, and relevant system
state $\Sigma$, write
\begin{equation}
\mathsf{Step}_{\tau}(\Sigma,\alpha,e)=(\Sigma',\mathcal{J}),
\label{eq:claim-transition}
\end{equation}
where $\Sigma'$ is the updated state and $\mathcal{J}$ contains the records
of fulfillment, transfer, or payment. The applicable checks concern
permission, product policy, current status, contract terms, and available
resources. Their role is to distinguish a valid obligation from an operation
that can presently execute: a positive forward amount does not establish
that its payer has sufficient funds, just as compensation approval does
not establish that payment has occurred.

Transfer changes a holder rather than fulfilling the underlying service
or revenue obligation. It is supported for specified objects, including
prepaid balances and Agent or Provider market assets, which record units
of transferable rights. For a valid purchase of $q$ existing units, with
sufficient seller units reserved, total seller and buyer holdings satisfy
\begin{equation}
H_s'=H_s-q,\qquad H_b'=H_b+q,
\label{eq:holding-transfer}
\end{equation}
where primes denote holdings after transfer. Thus, $H_s'+H_b'=H_s+H_b$:
transfer reallocates existing units rather than creating new ones.
Settlement connects this change to the buyer-to-seller payment.

Source references preserve the connection between the transferred right
and its original terms. A receiving prepaid lot retains its source expiry
and refund eligibility; transfer does not restart its validity. A derived
revenue-right holding retains references to its funding agreement and
revenue policy. Before distribution, \emph{holder resolution}, the routine
identifying recipients from these linked assets, updates the investor
participants from the selected current holdings. If no matching asset
exists, it uses the original investor. Recipient selection therefore
follows the holdings used at settlement, not necessarily ownership when
an earlier revenue event occurred. Historical allocation requires separate
rules rather than merely recorded transfer timestamps. 

\subsection{Accounting Consistency}
\label{sec:claim-realization}

Product rules determine why a financial change is required; accounting
records establish the resulting debit and credit. For an ordinary transfer
in unit $u$, let $\mathcal{J}_u^{\mathrm{credit}}$ and
$\mathcal{J}_u^{\mathrm{debit}}$ be its credit and debit entries, and $a_j$
the amount of entry $j$. Balanced posting requires
\begin{equation}
\sum_{j\in\mathcal{J}_u^{\mathrm{credit}}}a_j
-\sum_{j\in\mathcal{J}_u^{\mathrm{debit}}}a_j=0.
\label{eq:ledger-balance}
\end{equation}
Transfers spanning organizations may require checking several linked
records together. This relation concerns accounting entries, not raw
balance changes that may also include credit use or repayment. The workflow is shown in Figure~\ref{fig:contractual-representation}.

Product-level checks complement this accounting relation by bounding
consumption by available balances, deliveries by purchased quantities,
and compensation by available reserve funding. Recorded revenue must
also be conserved across recipient allocations. Transaction controls
coordinate concurrent updates and associate repeated requests with an
existing operation, with the intended effect of at most one committed
posting for the same operation identifier within its scope.

The representation therefore connects three questions: what is promised,
which evidence determines the obligation, and which records establish its
fulfillment or payment. The following product sections develop the forward,
financing, and protection mechanisms on this basis, keeping contractual
rights, ownership changes, and completed financial transfers distinct.

\section{Model API-Price Forwards}
\label{sec:api-price-forwards}

Future API expenditure varies with prices, request volume, and model
selection, even under a fixed execution policy. \system{} addresses its
price-related component through \emph{cash-settled forwards}: contracts
that exchange a payment based on an agreed price and a later reference
price, rather than deliver inference capacity. Separating this payment
from API procurement allows us to analyze its effect on expenditure
without changing the requests executed.

\subsection{Contract Formation}
\label{sec:forward-contract-model}
\label{sec:forward-formation}

A contract specifies two participants, a reference-price index identifying
the service price to observe, a fixed price, a contractual amount, and
fee, funding, and time conditions. The \emph{long} participant receives
payment when the reference price exceeds the fixed price; the \emph{short}
participant receives payment when it falls below that price. The parties
establish these terms through a \emph{request for quotation}, in which one
participant requests terms and another supplies an offer. Acceptance must
precede the quote-validity deadline, which is distinct from the contract's
recorded expiry. Activation then requires both parties to fund their
service fees and \emph{margin}, money reserved to support payment.

The contractual amount must be interpreted with the quotation units.
For example, a price per thousand tokens requires a quantity in thousands
of tokens. To express the payment in expenditure units, let
$P^{\mathrm{abs}}$ and $K^{\mathrm{abs}}$ be the observed and fixed prices
in the original units, $q$ the corresponding service quantity, and $P_0>0$
a baseline price. Define
\begin{equation}
P=\frac{P^{\mathrm{abs}}}{P_0},\qquad
K=\frac{K^{\mathrm{abs}}}{P_0},\qquad N=qP_0.
\end{equation}
Then $q(P^{\mathrm{abs}}-K^{\mathrm{abs}})=N(P-K)$.
Throughout the analysis, $P$ and $K$ are dimensionless, and the
\emph{notional} $N$, the amount scaling the price difference, represents
baseline-price expenditure. This normalization relates the analytical
notation to the contract's quotation units; product inputs and charges
must follow the corresponding unit convention.

\subsection{Valuation and Settlement}
\label{sec:forward-valuation}
\label{sec:forward-settlement}
\begin{figure*}[t]
    \centering
    \includegraphics[width=\textwidth]{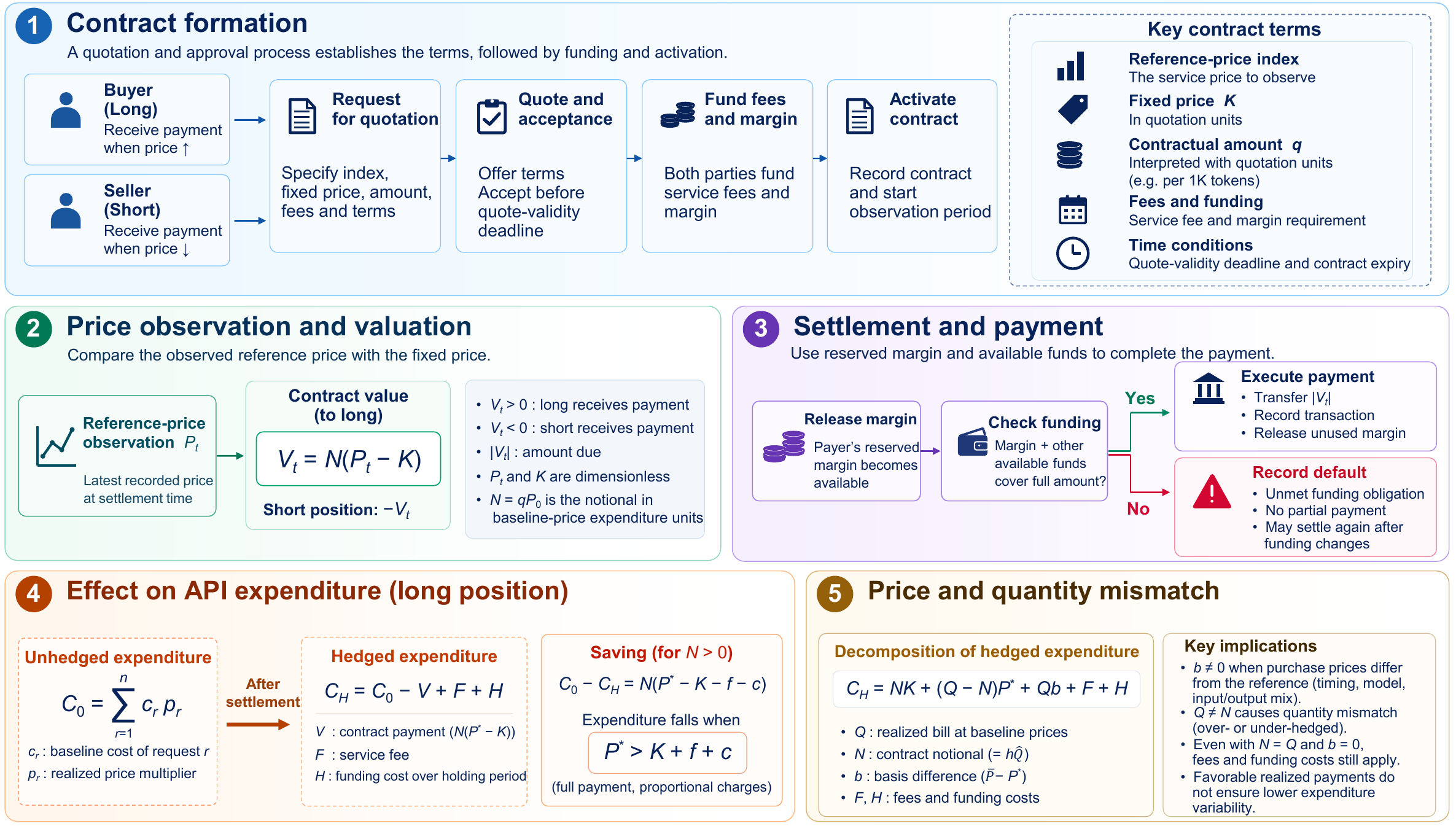}
    \caption{
    API-price forward mechanism in \system{}. Participants agree on
    contract terms, fund fees and margin, and settle the resulting
    price-difference payment against a reference price. The forward is
    financially separate from API procurement. Its expenditure effect
    depends on contract charges and on how closely the contracted amount
    and reference price match realized API purchases.
    }
    \label{fig:api-forward}
\end{figure*}
Figure~\ref{fig:api-forward} summarizes the forward workflow and its
relationship to API expenditure. The contract creates a cash payment
from the difference between an agreed price and a reference price,
while API procurement remains unchanged.

For reference-price observation $P_t$, the signed amount payable to the
long participant is
\begin{equation}
V_t=N(P_t-K),
\label{eq:forward-mtm}
\end{equation}
and the short participant has the opposite amount. The sign determines
payment direction and $|V_t|$ the amount due. This valuation identifies
the contractual obligation; settlement executes its payment. The reference
is the latest recorded price at the authorized settlement time, which need
not coincide with the recorded expiry.

At settlement, the payer's reserved margin becomes available for payment.
Since margin is already part of the account balance, releasing it does
not create additional funds. Payment completes only when the released
margin and other available funds cover the full obligation. Otherwise,
the contract records \emph{default}, an unmet funding obligation, without
partial payment and may be settled again after funding changes. Successful
settlement releases unused margin; a zero obligation requires no transfer.
The platform coordinates payments and collects fees without assuming
either participant's payment obligation. Detailed margin and transaction
controls are given in Appendix~\ref{app:forward-details}.

\subsection{Effect on API Expenditure}
\label{sec:forward-expenditure}
\label{sec:forward-scope}

Consider an operator holding the long position. For the same $n$ purchased
requests and execution settings, let $c_r\geq0$ be request $r$'s baseline
cost and $p_r$ its realized price multiplier. Unhedged expenditure is
\begin{equation}
C_0=\sum_{r=1}^{n}c_rp_r.
\label{eq:forward-unhedged-cost}
\end{equation}
Let $Q=\sum_r c_r$ denote the realized bill at baseline prices and
$\widehat{Q}$ its forecast before contract formation. The replay sizes
the contract as
\begin{equation}
N=h\widehat{Q},
\label{eq:forward-hedge-ratio}
\end{equation}
where the \emph{hedge ratio} $h$ scales the amount covered relative to
forecast expenditure. This is the replay's sizing convention, not an
implemented forecasting or contract-selection algorithm.

For settlement reference price $P^{*}$ and full payment
$V=N(P^{*}-K)$, expenditure becomes
\begin{equation}
C_H=C_0-V+F+H,
\label{eq:forward-hedged-cost}
\end{equation}
where $F$ is the operator's service fee and $H$ the modeled cost of
funding the position over its holding period. Refundable margin is not
itself an expense, although reserving funds may incur funding costs.
For $N>0$ and proportional charges $F+H=N(f+c)$, where $f$ is the fee
and $c$ the holding-period funding charge per unit of normalized notional,
the saving is
\begin{equation}
C_0-C_H=N(P^{*}-K-f-c).
\label{eq:forward-saving}
\end{equation}
Thus, expenditure falls exactly when
\begin{equation}
P^{*}>K+f+c.
\label{eq:forward-break-even}
\end{equation}
This condition concerns realized savings under full payment, not a
reduction in the amount of inference performed.

The same comparison must account for both participants. If $F_L$ and
$F_S$ are their service fees, the long and short parties receive $V-F_L$
and $-V-F_S$, with negative values denoting outgoing payments. The
platform receives $F_L+F_S$; these amounts sum to zero before separately
modeled funding and operating costs. A favorable contract payment to the
operator is therefore funded by its counterparty rather than newly
created service revenue.

\subsection{Price and Quantity Mismatch}
\label{sec:forward-basis}

The contract's reference price and amount need not match actual purchases.
For $Q>0$, define the purchase-price average and its difference from the
settlement reference as
\begin{equation}
\overline{P}=\frac{\sum_r c_rp_r}{Q},\qquad
b=\overline{P}-P^{*}.
\end{equation}
The \emph{basis difference} $b$ captures price mismatch, while $Q-N$
captures quantity mismatch in baseline-expenditure units. Substitution
into Equation~\eqref{eq:forward-hedged-cost} yields
\begin{equation}
C_H=NK+(Q-N)P^{*}+Qb+F+H.
\label{eq:forward-cost-decomposition}
\end{equation}
The first term is the fixed-price component for the contracted amount;
the next two retain the effects of unmatched quantity and purchase prices.
Fees and funding costs complete the expenditure.

Model selection, input/output-token mix, and purchase timing can produce
$b\neq0$. In particular, a time-average reference gives equal weight to
sampled prices, whereas requests concentrated in expensive periods yield
a higher purchase-weighted price. Thus, using the same underlying price
path does not ensure that the reference matches the bill.

When $N=Q$ and $b=0$, expenditure reduces to $C_H=QK+F+H$.
However, $h=1$ covers the forecast $\widehat{Q}$ rather than the realized
$Q$. If $N>Q$, the position is \emph{overhedged}: its contractual amount
exceeds the baseline cost of actual purchases. Covering all forecast
spending therefore need not fix realized expenditure. This decomposition
motivates evaluating contract size and price alignment together with
fees. Favorable realized payments alone do not establish lower expenditure
variability. Appendix~\ref{sec:forward-variability} provides the corresponding
variance relation and its assumptions.

\section{Revenue-Right Financing}
\label{sec:revenue-right-financing}

An Agent or Provider operator may need funds to serve available demand
before receiving customer payments. We call this operator the
\emph{publisher}. \system{} connects current funding to rights over
specified service revenue and supports subsequent transfers of these
rights. The design separates capital contributed to the publisher,
revenue earned from service delivery, and payments between holders.
This distinction allows us to analyze whether funding supports enough
additional business to compensate the publisher for sharing its revenue.
\begin{figure*}[t]
    \centering
    \includegraphics[width=\textwidth]
    {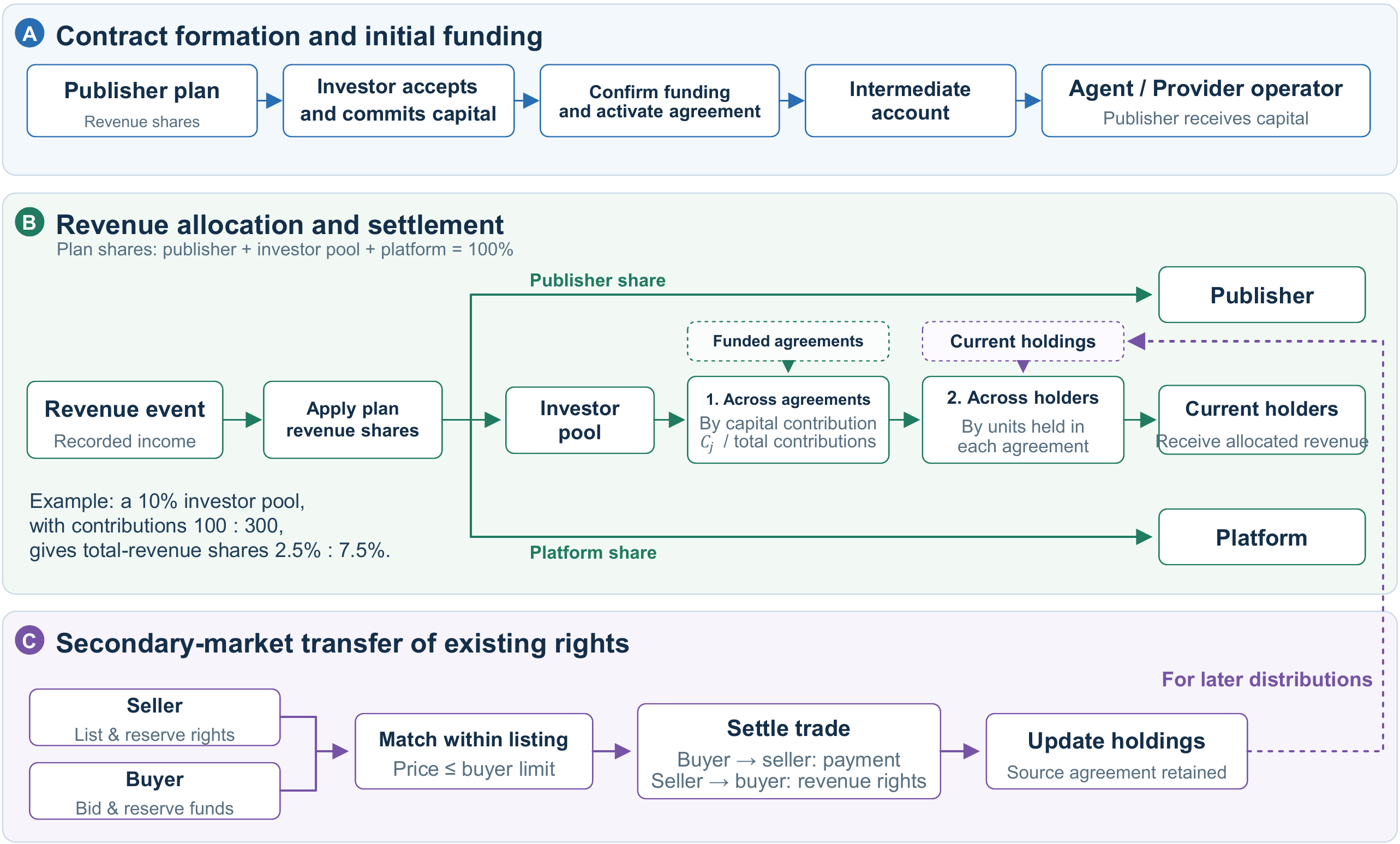}
    \caption{
    Revenue-right financing and transfer in \system{}. Primary
    investment supplies capital to an Agent or Provider operator in
    exchange for rights to specified future revenue. Recorded service
    revenue is later divided among the publisher, investor pool, and
    platform, with investor allocations determined from funding
    agreements and current holdings. Existing rights may subsequently
    be transferred: the buyer pays the selling holder, while future
    distributions follow the updated holdings.
    }
    \label{fig:revenue-financing}
\end{figure*}

Figure~\ref{fig:revenue-financing} shows the distinction among primary
funding, subsequent revenue distribution, and secondary transfer of
existing rights. These flows involve different payments and should not
be treated as a single revenue transaction.

\subsection{Contract Formation and Funding}
\label{sec:financing-contract}
\label{sec:financing-capital}

A publisher plan identifies the financed service, accounting unit, and
revenue shares assigned to the publisher, investors, and platform.
Shares $b_O$, $b_I$, and $b_P$ are expressed in \emph{basis points},
where one basis point is $1/10{,}000$ of the amount being divided:
\begin{equation}
b_O+b_I+b_P=10{,}000.
\label{eq:financing-plan-shares}
\end{equation}
The \emph{investor pool} is the share $b_I>0$ reserved for investors
collectively; individual allocations depend on their contributions
and holdings, as defined below. These rights concern specified receipts,
not ownership of the Agent, model, service, or company. A finite duration,
cumulative payment cap, or principal-repayment schedule requires separate
terms and cannot be inferred from the revenue share alone.

An investor accepts a plan and commits an amount $C_j$ under agreement
$j$. Confirmation checks funding and activates the agreement, then
transfers capital to the publisher through an intermediate account.
The disbursed amount $C_j^{\mathrm{settled}}$ records completed capital
transfers; normal full disbursement establishes
$C_j^{\mathrm{settled}}=C_j$. This \emph{capital settlement} differs from
\emph{revenue settlement}, which distributes later recorded service
income. Investment principal is therefore recorded as funding rather
than operating revenue. Appendix~\ref{app:financing-details} specifies
the records and funding controls underlying this distinction.

\subsection{Revenue Allocation and Settlement}
\label{sec:financing-investor-allocation}
\label{sec:financing-revenue-capture}
\label{sec:financing-distribution}

Allocation proceeds in two stages: dividing the investor pool among
funding agreements, then dividing each agreement's allocation among its
current holders. At settlement time $t$, let $\mathcal{J}_{\pi}(t)$
contain the active agreements linked to plan $\pi$ with positive recorded
capital settlement. For a nonempty selected set, agreement $j$ has ideal share
\begin{equation}
b_j^{*}(t)=b_I\frac{C_j}{\sum_{k\in\mathcal{J}_{\pi}(t)}C_k}.
\label{eq:financing-investor-share}
\end{equation}
Weights use committed amounts $C_j$, which equal disbursed amounts in
the normal completed funding path. The implementation approximates these
allocations with integer basis points $b_j(t)$ while conserving $b_I$.

A revenue right can be subdivided into transferable units linked to its
source agreement. Let $\mathcal{H}_j(t)$ contain the selected holders of
agreement $j$'s rights and $u_{j,h}(t)$ holder $h$'s units. Its ideal share is
\begin{equation}
b_{j,h}^{*}(t)=b_j(t)
\frac{u_{j,h}(t)}{\sum_{k\in\mathcal{H}_j(t)}u_{j,k}(t)}.
\label{eq:financing-holder-share}
\end{equation}
A second integer allocation divides $b_j(t)$ among these holders. For
illustration, a 10\% investor pool shared by contributions of 100 and
300 units assigns ideal total-revenue shares of 2.5\% and 7.5\%.
Holding half the first agreement's units gives half its 2.5\% share,
not half the publisher's revenue. New investments change the division
within the pool while leaving its total share unchanged.

\emph{Holder resolution}, the lookup identifying these recipients,
follows source-agreement references to current positive holdings,
including supported transfers across \emph{tenants}, the organizational
scopes governing accounts and permissions. If no derived holding exists,
it uses the original investor. Recipients are updated immediately before
distribution, rather than reconstructed from ownership when income first
occurred. A pending event can therefore use a later holding configuration.
The lookup also does not determine agreement eligibility by comparing
event time with agreement start and end dates.

A \emph{revenue event} records service income submitted for distribution,
whether entered explicitly or captured from service usage. Its recorded
net amount $R_e$ is the distribution base, not necessarily profit after
operating costs. For active recipients $i$ with shares $b_i$, including
the publisher, investor holders, and platform, exact allocations are
\begin{equation}
d_{i,e}^{*}=R_e\frac{b_i}{10{,}000},\qquad
\sum_i b_i=10{,}000.
\label{eq:financing-event-entitlement}
\end{equation}
Actual payments use a minimum increment of $10^{-6}$ accounting units.
The allocation routine carries forward rounding differences and adjusts
payments so that, for event amounts at the supported precision,
\begin{equation}
d_{i,e}\geq0,\qquad \sum_i d_{i,e}=R_e.
\label{eq:financing-event-conservation}
\end{equation}
Settlement links these payments to their revenue event and recipient
accounts. Events awaiting an eligible funded investment can remain
pending. Integer-share allocation, payment rounding, and wallet transfers
are detailed in Appendix~\ref{app:financing-allocation}.

\subsection{Secondary-Market Transfer}
\label{sec:financing-right-holdings}

A \emph{secondary transfer} sells an existing revenue right to another
holder, rather than creating a new investment in the publisher.
Transferable units retain their funding-agreement and revenue-policy
references, connecting a change in ownership to the recipient lookup
above. For publisher investments, creating these units requires confirmed
investment and completed capital settlement.

A holder creates a \emph{listing}, an offer to sell a specified quantity
at a minimum unit price $p_{\ell}^{\min}$. The units are reserved against
other use, and the listing generates a sell order at that minimum price.
A buyer submits quantity $q_b$ and limit price $p_b$, the maximum price
it accepts per right unit. Its initial payment reservation is
\begin{equation}
W_b=q_bp_b.
\label{eq:secondary-buyer-reservation}
\end{equation}
Reserving both sides' commitments precedes matching; it does not itself
transfer ownership or complete payment.

Matching operates within a listing, prioritizing higher-priced buy orders
and lower-priced sell orders, with creation time resolving ties. For sell
limit $p_s$, the \emph{execution price}, the price recorded for a trade, is
\begin{align}
p_A^{*}&=p_s,
\label{eq:secondary-agent-price}\\
p_P^{*}&=\max\{p_s,p_{\ell}^{\min}\},
\label{eq:secondary-provider-price}
\end{align}
where $A$ and $P$ identify the Agent and Provider paths. Each path requires
$p_b\geq p^{*}$, ensuring that the recorded price does not exceed the
buyer's limit. If $r_b$, $r_s$, and $r_{\ell}$ are the remaining buy,
sell, and listing quantities, the matched quantity and payment are
\begin{align}
q^{*}&=\min\{r_b,r_s,r_{\ell}\},
\label{eq:secondary-match-quantity}\\
V^{*}&=q^{*}p^{*}.
\label{eq:secondary-consideration}
\end{align}

Settlement connects the buyer-to-seller payment to an equal decrease
and increase in their holdings, conserving existing units and retaining
the source references used in subsequent distributions. For changes in
buyer, seller, and publisher funds, excluding separately specified charges,
\begin{equation}
\begin{split}
\Delta C_b&=-V^{*},\qquad \Delta C_s=V^{*},\\
\Delta C_{\mathrm{publisher}}&=0.
\end{split}
\label{eq:secondary-cashflow}
\end{equation}
Thus, resale changes the holder and pays the seller without adding to
the publisher's funding. Supported cross-tenant transfers also enter
subsequent holder resolution. Prices follow submitted orders, not a
system forecast of future revenue; execution remains contingent on a
compatible counterparty. Appendix~\ref{app:financing-transfers} specifies
the reservation and ownership updates.

\subsection{Financing and Publisher Profit}
\label{sec:financing-served-demand}
\label{sec:financing-publisher-profit}
\label{sec:financing-accounting-scope}

Financing changes the resources available for service delivery, whereas
resale changes who receives an existing revenue allocation. To isolate
the first effect, let $X(B)$ be the invoice cost of requests admitted
under daily budget $B$. The replay admits the longest initial sequence
of complete requests that fits each day's budget and compares
$X_0=X(B)$ with $X_1=X(aB)$ for $a\geq1$. Requests retain their original
order and costs; additional funding can serve more of the same offered
demand but does not create new arrivals. This retrospective calculation
assumes daily budget resets and does not derive the multiplier $a$
from a particular investment. Appendix~\ref{app:financing-workload}
provides the workload definitions.

For publisher $o$, let $R_o(X)$ be revenue from served work $X$ and
$V_o(X)$ its variable cost, the modeled operating cost of that work.
With investor fraction $s_o$ and platform fraction $z_o$, define
\emph{contribution profit} as revenue remaining after these shares
and variable costs:
\begin{equation}
\Pi_o(X)=(1-s_o-z_o)R_o(X)-V_o(X).
\label{eq:financing-contribution-profit}
\end{equation}
Suppose variable costs consume a constant fraction $\nu$ of revenue,
so $g=1-\nu$ remains before sharing. Let $R_0,R_1$ be baseline and
financed revenue, $z_0,z_1$ their platform fractions, and $s$ the
financed investor fraction. The baseline has no investor share, giving
$\Pi_0=(g-z_0)R_0$ and $\Pi_1=(g-s-z_1)R_1$. For $R_0>0$ and
$g-s-z_1>0$, financing improves contribution profit exactly when
\begin{equation}
\frac{R_1}{R_0}>\frac{g-z_0}{g-s-z_1}.
\label{eq:financing-growth-threshold}
\end{equation}
This condition identifies the additional revenue required to offset
the smaller retained share. Once existing demand is fully served, a
larger budget cannot provide that growth under the model. Increasing
the sharing burden without additional revenue reduces contribution.

\paragraph{Agent and Provider accounting.}
For admitted API invoice cost $X$, suppose the Agent charges a markup
$m$ over its API expense and the Provider incurs variable cost $vX$.
With $s_A,s_P$ their investor fractions and $z_A,z_P$ their platform
fractions, the model gives
\begin{align}
R_A&=(1+m)X,\qquad R_P=X,\notag\\
\Pi_A&=(1-s_A-z_A)(1+m)X-X,\notag\\
\Pi_P&=(1-s_P-z_P)X-vX.
\label{eq:financing-agent-provider-profit}
\end{align}
Investor receipts are $Y_I=s_AR_A+s_PR_P$, and platform receipts are
$Y_{\mathrm{platform}}=z_AR_A+z_PR_P$. The Agent's API payment is the
Provider's revenue, an internal transfer rather than a second external
customer sale. Cancelling this payment and the revenue-sharing transfers
therefore yields
\begin{equation}
\Pi_A+\Pi_P+Y_I+Y_{\mathrm{platform}}=R_A-vX.
\label{eq:financing-aggregate-conservation}
\end{equation}
Sharing reallocates this contribution; additional aggregate contribution
requires more revenue-generating work or changed operating costs.
Investor receipts are not net returns without accounting for contributed
capital, payment timing, and any retained rights. Evaluation examines
these financing conditions separately from secondary-market prices,
trading activity, and resale returns, which the financing replay does
not measure.
\section{Service-Failure Protection}
\label{sec:service-failure-protection}

\system{} addresses the fourth challenge by specifying which service
failures qualify for compensation, how much is payable, and how payment
is funded. A Provider publishes a protection plan, and buyers subscribe
to its terms. An \emph{incident} records a service disruption; an
\emph{impact record} summarizes affected requests; and a \emph{claim}
requests compensation based on that evidence. Approved payments draw
on a \emph{reserve}, an account and associated budget set aside for
compensation. The following design connects these records from coverage
to payment, then separates their financial effects from technical recovery.

\begin{figure*}[t]
    \centering
    \includegraphics[width=\textwidth]
    {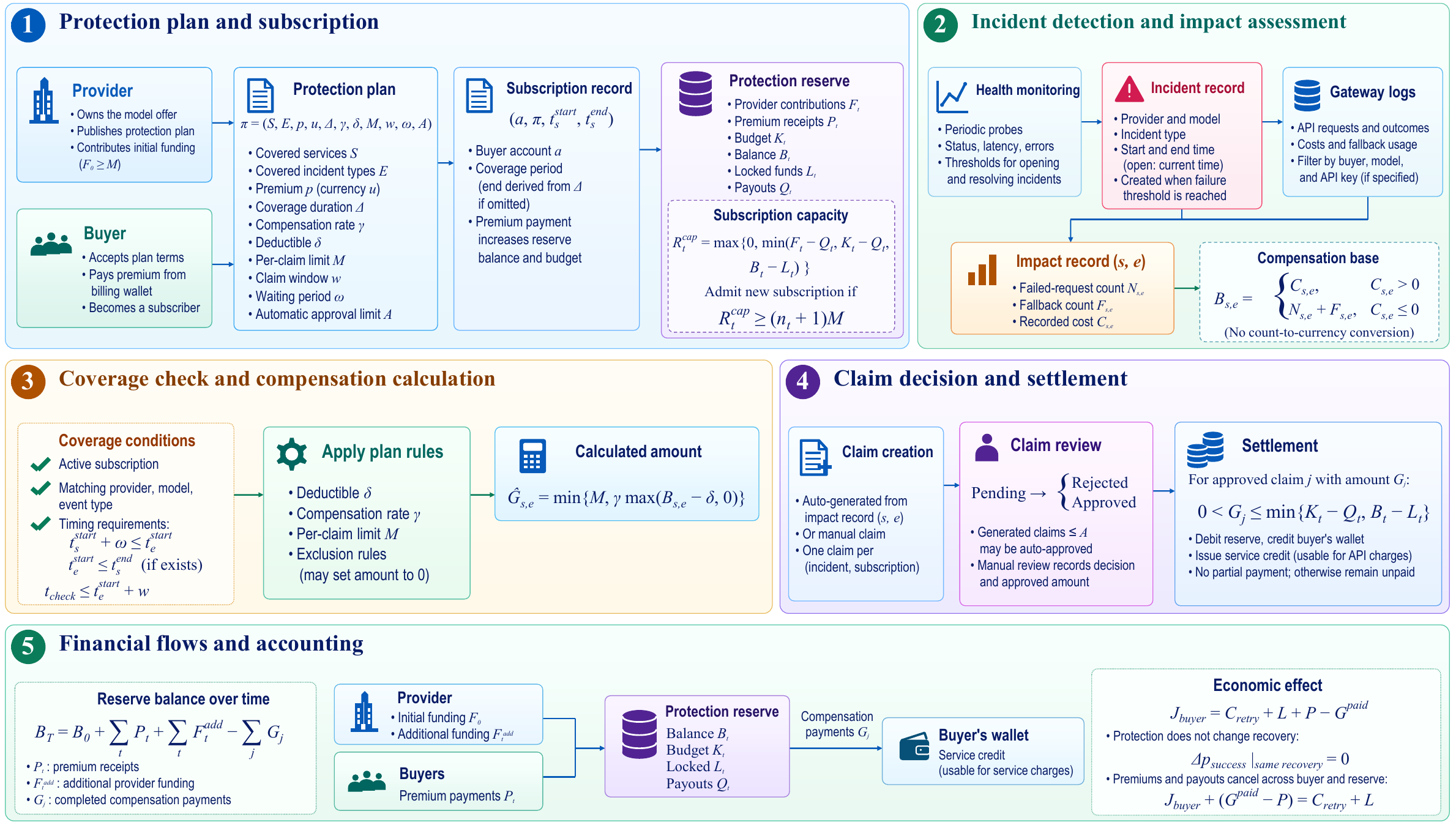}
    \caption{
    Service-failure protection in \system{}. A Provider-funded reserve
    and buyer protection fees support coverage under predefined plan
    terms. Monitoring and request records establish incident evidence,
    coverage and impact rules determine an eligible claim amount, and
    approved claims are paid as service credit subject to available
    reserve funding. Financial compensation is evaluated separately
    from retry and failover recovery.
    }
    \label{fig:service-protection}
\end{figure*}

Figure~\ref{fig:service-protection} summarizes the protection workflow
from contract formation and incident evidence to claim approval and
reserve-funded settlement. Technical recovery and financial protection
remain separate throughout the process.
\subsection{Protection Contract and Reserve}
\label{sec:protection-plan}
\label{sec:protection-capacity}

A plan identifies the protected service and specifies
\begin{equation}
\pi=(\mathcal{S},\mathcal{E},p,u,\Delta,\gamma,\delta,M,w,\omega,A),
\label{eq:protection-plan}
\end{equation}
where $\mathcal{S}$ and $\mathcal{E}$ identify covered services and incident
types. The \emph{premium} $p$ is the protection fee, $u$ the payment
currency, and $\Delta$ the default coverage duration. The compensation
rate $\gamma$ applies after a \emph{deductible} $\delta$, the amount
excluded before compensation. Each claim is capped at $M$. The window
$w$ limits claim submission time, $\omega$ specifies the waiting period
before coverage becomes effective, and $A$ limits automatic approval.

The publishing Provider must own the specified model offer, the
model-specific service being protected. A subscription associates buyer
account $a$ with plan $\pi$ and coverage times
$t_s^{\mathrm{start}}$ and $t_s^{\mathrm{end}}$; an omitted end time is
derived from $\Delta$. On application, the buyer accepts the terms and
pays the fixed premium from its billing wallet, the account used for
service payments. The payment increases the reserve's balance and budget.

Before publication, Provider funding $F_0$ must satisfy $F_0\geq M$.
Subsequent admission also accounts for existing subscriptions and payments.
At time $t$, let $F_t$ denote cumulative Provider contributions, $Q_t$
cumulative compensation paid, $K_t$ the recorded reserve budget, $B_t$
its account balance, and $L_t$ funds locked against other use. The amount
used to determine subscription capacity is
\begin{equation}
\begin{split}
R_t^{\mathrm{cap}}=\max\{0,\min(&F_t-Q_t,\;K_t-Q_t,\\
&B_t-L_t)\}.
\end{split}
\label{eq:protection-capacity-reserve}
\end{equation}
These bounds respectively account for Provider funding after payouts,
remaining budget, and currently available funds. For $M>0$, let $n_t$
count active subscriptions that have not expired, including those
without an end date. An active published plan admits another subscription
only if
\begin{equation}
R_t^{\mathrm{cap}}\geq(n_t+1)M.
\label{eq:protection-admission}
\end{equation}
Premium receipts do not increase the Provider-contribution bound
$F_t-Q_t$. This rule allocates capacity for one maximum claim per counted
subscription; because $M$ is a per-claim limit, it does not establish
sufficient funding for every sequence of repeated incidents.

\subsection{Incident Evidence and Coverage}
\label{sec:protection-monitoring}
\label{sec:protection-impact}

Periodic \emph{probes}, test requests to the protected service, record
status, latency, and errors. Configured numbers of consecutive unhealthy
observations open an incident; consecutive healthy observations resolve
it. An automatically created incident starts when the failure threshold
is reached, not at the first failed observation. Its Provider, model,
type, and time interval determine the scope of subsequent checks.

Coverage requires an active subscription and matching Provider, model,
and event type. For incident $e$, the timing conditions are
\begin{align}
t_s^{\mathrm{start}}+\omega&\leq t_e^{\mathrm{start}},\\
t_e^{\mathrm{start}}&\leq t_s^{\mathrm{end}}
\quad\text{if an end time exists},\\
t_{\mathrm{check}}&\leq t_e^{\mathrm{start}}+w,
\label{eq:protection-time-coverage}
\end{align}
where $t_{\mathrm{check}}$ is the time of the coverage check. The deadline
is measured from incident start rather than resolution.

For each eligible subscription, \emph{gateway logs}---records of API
requests, outcomes, and costs---identify failed requests or requests
using a fallback service during the incident interval. Selection matches
the buyer and Provider, with additional model and API-key restrictions
where specified. An open incident uses the current time as its interval
end. The impact record aggregates failed-request count $N_{s,e}$,
fallback count $F_{s,e}$, and recorded cost $C_{s,e}$. These define the
\emph{compensation base}, the input to the plan's amount calculation:
\begin{equation}
B_{s,e}=\begin{cases}
C_{s,e}, & C_{s,e}>0,\\
N_{s,e}+F_{s,e}, & C_{s,e}\leq0.
\end{cases}
\label{eq:protection-operational-base}
\end{equation}
The count-based branch specifies no count-to-currency conversion and
must not be interpreted as a measured monetary loss. A request can
contribute to both counts, and fallback activity can include successful
recovery. Thus, this operational base differs from the assumed loss on
unresolved requests used in Section~\ref{sec:protection-economics}.

The calculated compensation is
\begin{equation}
\widehat{G}_{s,e}=\min\{M,\;\gamma\max(B_{s,e}-\delta,0)\}.
\label{eq:protection-impact-compensation}
\end{equation}
The rule applies the deductible and rate, then enforces the per-claim
limit. A matching \emph{exclusion rule}, a configured condition that
disallows compensation, can set the amount to zero. Service-health
observations support these coverage decisions but do not independently
determine the buyer's downstream business loss.

\subsection{Claim Decision and Settlement}
\label{sec:protection-claims}
\label{sec:protection-settlement}

A positive impact amount can generate a pending claim linked to its
subscription and incident. An incident--subscription identifier allows
repeated generation to retrieve the existing claim. Manual claims enter
the same decision process after coverage checks:
\begin{equation}
\mathrm{pending}\rightarrow\begin{cases}
\mathrm{rejected},\\
\mathrm{approved}\rightarrow\mathrm{settled}.
\end{cases}
\label{eq:protection-claim-lifecycle}
\end{equation}
Review records the decision and approved amount. Ordinary manual claims
apply the deductible, rate, and cap to the reviewed base. Generated
claims already contain the calculated compensation, so review caps their
amount without applying the deductible and rate again. A generated
pending claim without an exclusion can be approved automatically when
its requested amount is positive and no greater than $A>0$.

Approval authorizes an amount but does not complete payment. For approved
claim $j$ with amount $G_j$, settlement requires the plan's active reserve,
compatible payment units, and
\begin{equation}
0<G_j\leq\min\{K_t-Q_t,\;B_t-L_t\}.
\label{eq:protection-payment-condition}
\end{equation}
Unlike subscription admission, this check tests the remaining budget and
available balance for the particular payment. Insufficient funds leave
the claim unpaid rather than producing a partial payment. Automatic
approval may trigger settlement, but it does not bypass these checks.

Successful settlement debits the reserve and credits the buyer's active,
currency-matched billing wallet. It issues \emph{service credit}, usable
for service charges rather than cash withdrawal. Wallet amounts must
be exactly representable to two decimal places. Transactional record
updates and claim-specific settlement identifiers coordinate concurrent
operations and prevent supported retries from posting the payment twice.
Settlement records retain links to the claim, incident, reserve, and
accounting transaction.

Over an interval ending at $T$, reserve accounting gives
\begin{equation}
B_T=B_0+\sum_t P_t+\sum_t F_t^{\mathrm{add}}-\sum_j G_j,
\label{eq:protection-reserve-evolution}
\end{equation}
where $P_t$ denotes premium receipts, $F_t^{\mathrm{add}}$ additional
Provider funding, and the final sum includes completed payments only.
Opening reserves and additional contributions supply capital; they are
not premium revenue. This distinction separates the ability to pay a
claim from whether collected premiums cover compensation.

\subsection{Separating Recovery from Protection}
\label{sec:protection-recovery}

The economic comparison holds routing and retry decisions fixed while
varying financial protection. Let $p_1$ be the initial failure probability.
Conditional on failure, $\rho\in[0,1]$ is the assumed fraction of cases
in which an alternative Provider cannot recover the request. In the
remaining cases, its failure probability is $p_2$. For one alternate
attempt, the expected unresolved fraction is
\begin{equation}
p_{\mathrm{unresolved}}=p_1[\rho+(1-\rho)p_2].
\label{eq:protection-unresolved}
\end{equation}
The recovered fraction, measured relative to all initial requests, is
$p_1(1-\rho)(1-p_2)$. Here, $\rho$ is a scenario assumption, not a measured
statistical correlation, and these probabilities do not define the
implementation's monitoring counters.

Because compensation does not alter this recovery policy,
\begin{equation}
\Delta p_{\mathrm{success}}\bigm|_{\mathrm{same\ recovery}}=0.
\label{eq:protection-no-recovery-change}
\end{equation}
This equality states the model's separation of execution and payment,
not an independently observed improvement in reliability.

\subsection{Participant Economics}
\label{sec:protection-economics}

To evaluate financial outcomes, let $U_r$ be one if request $r$ remains
unresolved and zero otherwise, and let $\ell_r\geq0$ be its assumed
monetary loss.
The \emph{residual loss}, the loss remaining after recovery, is
\begin{equation}
L=\sum_r\ell_rU_r.
\label{eq:protection-residual-loss}
\end{equation}
The analytical protection rule uses this assumed loss rather than the
operational base in Equation~\eqref{eq:protection-operational-base}:
\begin{equation}
G^{\mathrm{elig}}=\chi\min\{M_{\mathrm{eval}},
\;\gamma\max(L-\delta,0)\}.
\label{eq:protection-residual-claim}
\end{equation}
Here, $\chi$ indicates modeled eligibility, and $\gamma$ and $\delta$
are the rate and deductible specified for the scenario. The cap
$M_{\mathrm{eval}}$ applies to the evaluated workload, not to each
implementation claim as $M$ does. Actual credited compensation
$G^{\mathrm{paid}}$ additionally depends on approval and funding.
These quantities are modeled outcomes, not observed compensation payments.

Let $C_{\mathrm{retry}}$ be recovery expenditure and $P$ total protection
fees. Valuing usable credit at its recorded amount, the buyer's
failure-related cost is
\begin{equation}
J_{\mathrm{buyer}}=C_{\mathrm{retry}}+L+P-G^{\mathrm{paid}}.
\label{eq:protection-buyer-burden}
\end{equation}
This is not the complete API bill. Relative to the same recovery policy
without protection, the change is $P-G^{\mathrm{paid}}$. The reserve's
net outgoing payment is the opposite amount, giving
\begin{equation}
J_{\mathrm{buyer}}+(G^{\mathrm{paid}}-P)=C_{\mathrm{retry}}+L.
\label{eq:protection-loss-conservation}
\end{equation}
Premiums and compensation therefore cancel across the buyer and reserve
under this accounting convention. Payouts exceeding fees require reserve
capital but do not alone establish an inability to pay. Conversely,
funding a payment from initial capital does not establish that premiums
cover the protection activity's costs.

This comparison assumes that the buyer can use the entire credit for
subsequent service purchases; otherwise, its economic benefit is lower.
Together, the coverage, payment, and accounting rules distinguish eligible
compensation from completed transfers and changes in loss allocation
from changes in request recovery. Evaluation specifies the pricing,
loss, and credit-use assumptions separately from these contract workflows.
\section{Evaluation}
\label{sec:evaluation}
We evaluate \system{} through contract execution, request-level replay,
and real-model Agent studies. The evaluation examines four questions:
whether contract operations preserve payment and ownership rules; how
API-price forwards affect expenditure and its variability; when additional
capital supports service delivery and operator contribution; and how
recovery and protection jointly determine completed services and loss
allocation. Matched controls separate each mechanism's financial effects
from changes in execution, funding, and recovery.

The principal findings quantify these relationships. Under the assumed
rising-price path, zero-discount forwards reduce mean expenditure by
USD~304.88. In the matched low-capital study, revenue-right financing
increases mean operator contribution by 1.0635 experimental USD and
reduces mean admission stops from 12.0 to 1.8. Under twelve controlled
local faults, recovery completes nine services, and adequately funded
protection settles the approved residual compensation. The following
analyses relate these results to price exposure, capital availability,
revenue allocation, and reserve funding.

\subsection{Workloads, Controls, and Measurement}
\label{sec:eval-workloads}
\label{sec:eval-setup}
\paragraph{Tasks, requests, and successful services.}
The controlled workload contains twelve constructed tasks: four each for
structured extraction, order lookup, and constrained product selection.
A service succeeds when both the business operation and final output meet
the predefined completion rule. A \emph{case} is one scheduled execution
of a task under one experimental setting; it may contain several model
calls. Every scheduled case enters its study denominator, including
funding-admission stops, injected faults, and resource-limit outcomes.
Table~\ref{tab:eval-cohorts} reports task, request, response, and success
counts separately for each study.

\begin{table}[!htbp]
\centering\small
\caption{Real-model study populations. Cases count scheduled task executions;
calls include all sent requests. Responses and successful services follow
their respective measurement rules. Each study retains its own denominator.}
\label{tab:eval-cohorts}
\begin{tabular}{@{}lrrrr@{}}\toprule
Study & Cases & Calls & Responses & Successes\\\midrule
Repeated capital study & 540 & 1,082 & 1,081 & 456\\
Controlled repeats and parameter tests & 624 & 1,247 & 1,246 & 487\\
Initial controlled study & 84 & 150 & 150 & 51\\
Retail contract study & 70 & 930 & 925 & 1\\
Retail baseline & 10 & 368 & 360 & 4\\
\bottomrule\end{tabular}
\end{table}

All model calls use Qwen/Qwen3-8B through Nexus and SiliconFlow. Controlled
tasks use constructed local records, callable tools, and explicit completion
checks. Two passes over nine calibration tasks produce seven and nine
successes, respectively. The final interface enumerates category codes and
orders catalog lookup before quote creation; the model selects products
and generates its own responses. This configuration is fixed before the
twelve evaluation tasks are executed. Repetitions characterize financial
and operational effects within the same calibrated task families.

The retail studies use the official retail text environment of
$\tau^2$-Bench~\cite{barres2025tau2}, including its model-driven customer
and predefined ALL completion criterion. Task selection, execution limits,
calibration passes, and protocol amendments appear in
Appendix~\ref{app:eval-protocols}; the controlled design is specified in
Appendix~\ref{app:controlled-protocol}.

\paragraph{Observed quantities and economic assumptions.}
\label{sec:eval-measurement-scope}
We measure returned token usage, tool results, request identifiers, account
changes, and contract settlements. Available usage is reconciled against
Nexus, and payments execute through authenticated business APIs on isolated
PostgreSQL research accounts. The configured Nexus charge is zero and
supplier invoices are unavailable; monetary outcomes therefore use stated
experimental tariffs and success-triggered receipts. Initial capital,
future price indices, and local fault schedules are controlled inputs.
Observed settlements and scenario-based economic valuations are reported
at their respective measurement scopes.

Table~\ref{tab:eval-parameters} defines the controlled-study parameters.
$B$ is baseline own capital, $A$ additional funding, and $r$ revenue per
successful service. Input and output tariffs are 100 and 500 experimental
USD per million tokens, selected before execution to represent small-task
losses at the protection mechanism's one-cent precision. The retail study
uses tariffs of 1 and 5 per million tokens, with
$B=\mu=0.184298$, $A=0.370000$, and $r=0.368596$ experimental USD.
Each study retains its own tariff, rounding, and coverage conventions.

\begin{table}[!htbp]
\centering\small
\caption{Controlled-task parameters fixed from the nine calibration tasks.
Amounts are experimental USD. The repeated capital study changes the
starting own balance to $0.5B$, $B$, or $1.5B$ and keeps added funds $A$ fixed.}
\label{tab:eval-parameters}
\begin{tabularx}{\linewidth}{@{}lXr@{}}\toprule
Symbol & Definition & Value\\\midrule
$\mu$ & Mean known expense per scheduled calibration task & 0.127689\\
$H_{\mathrm{cal}}$ & Largest amount set aside before a calibration request & 0.943300\\
$B$ & Baseline own capital, $\mu+H_{\mathrm{cal}}$ & 1.070989\\
$A$ & Added capital, $\lceil4\mu\rceil_{0.01}$ (rounded up to a cent) & 0.520000\\
$r$ & Assumed revenue per successful service, $2\mu$ & 0.255378\\
$d$ & Scheduled tasks between releases of earned revenue & 4\\
\midrule
\multicolumn{2}{@{}l}{Revenue-right shares: operator / investor / platform} & 89\% / 10\% / 1\%\\
\multicolumn{2}{@{}l}{Control shares: operator / platform} & 99\% / 1\%\\
\bottomrule\end{tabularx}
\end{table}

For real-model studies, \emph{operating contribution} is the operator's
retained task revenue after sharing, less inference expense at the
experimental tariff. Initial funding, fixed operating costs, and the value
of outstanding rights are excluded. Investor wallet transfers and
\emph{service credits}, which can offset later service charges, are
reported separately. Credit utilization is a scenario parameter; the
observation window covers issuance rather than subsequent redemption or
lifetime investment returns.

\paragraph{Request-level replay.}
We replay the first two months of BurstGPT~\cite{burstgpt2025}, retaining
recorded timing and usage while applying specified prices and contracts.
The dataset revision is recorded in the supplementary artifact. Removing
conversation records and the incomplete final day leaves 1,261,858 API
requests: 362,417 on days 0--17 for calibration and 899,441 on days 18--59
for evaluation. Timestamps, model labels, token counts, and request order
are preserved. At assumed input/output tariffs of USD~3/15 per million
tokens, successful first attempts cost USD~3,395.80116.

The replay uses 16,137 zero-output records as failure indicators; 16,131
also lack input lengths. Compared settings share the same assumptions for
unobserved first-attempt charges and calibration-based alternate-attempt
lengths. Five price scenarios start from index one: constant prices,
a linear rise to 1.3, a linear fall to 0.7, a seven-day rise to 1.8, and
a load-dependent increase. Forward controls match execution and
cache/batch settings; financing-cost controls match total funding;
protection controls match recovery. The replay's financing model assigns
10\% of revenue to investors and 2\% to the platform, while real-model
studies assign 1\% to the platform. These conventions are specified in
Appendices~\ref{app:eval-protocols}--\ref{app:eval-financing}.

\paragraph{Continuous-balance execution and matched controls.}
Each financing condition is executed separately so that available funds
can affect request admission. Own capital starts with $B_0$; equal capital
and revenue-right financing both start with $B_0+A$. The equal-capital
control matches available funds while omitting the investor share,
isolating the revenue-allocation component of financing.

A controlled \emph{portfolio} schedules twelve tasks against a continuous
balance. Revenue is released after each group of four scheduled tasks,
including tasks denied admission. Before a request, funds are reserved
using input bytes and maximum output length; returned usage determines
the charge and releases the unused reservation. The matched capital study
crosses three initial balances ($0.5B$, $B$, $1.5B$), three financing
conditions, and five repetitions, yielding
$5\times3\times3\times12=540$ cases. Task order and within-task arm order
are shared across capital levels within a repetition, while capital-level
order is randomized separately. Capital levels are informed by the earlier
boundary study, and the replication protocol is fixed before its own
observations. The separate 624-case study includes five repeated
comparisons, seven one-factor financing conditions, and three
fault-frequency runs.

\paragraph{Missing-usage bounds and resampling.}
Requests without returned usage retain their reservations. Let
$C_{\mathrm{known}}$ be known expense, $Y_{\mathrm{ret}}$ retained
revenue, and $H_u$ the unresolved reservation used as the declared upper
bound on missing expense. Contribution satisfies
\begin{equation}
\Pi\in[Y_{\mathrm{ret}}-C_{\mathrm{known}}-H_u,
        \;Y_{\mathrm{ret}}-C_{\mathrm{known}}].
\label{eq:eval-usage-bound}
\end{equation}
Treatment-minus-control bounds pair the lower treatment endpoint with
the upper control endpoint, and conversely. Separately, 10,000 draws
resample whole matched repetitions, preserving within-repetition pairing.
The 2.5th percentile of lower mean bounds and the 97.5th percentile of
upper mean bounds define the exploratory 95\% ranges. Portfolios are
the sampling units because tasks share operating funds. These ranges
describe observed fixed-task repetitions; broader population coverage
and multiplicity-adjusted significance are outside this design.
Appendix~\ref{app:eval-protocols} records the timing of the missing-usage
accounting protocol.
\FloatBarrier

\subsection{Contract Correctness and Accounting Controls}
\label{sec:eval-implementation}
Business-API tests validate payment amounts, recipients, funding checks,
and idempotent execution. Revenue-right transfers are checked against
subsequent recipient allocation, while repeated payment requests are
checked for a single committed transfer
(Table~\ref{tab:extension-evidence}).

All 206 backend regression tests pass. Independent decimal-arithmetic
replay verifies 165 account updates across 47 accounts with zero arithmetic
discrepancies. Direct ledger association covers 103 entries; five
intermediate wallet-funding updates are documented through their
business-operation records. These are complementary record-linkage scopes.
Overlapping regression and targeted suites are counted once.

\begin{table}[!htbp]
\centering\small
\caption{Contract validation through business APIs on isolated PostgreSQL accounts.
Synthetic funds exercise payment, ownership, funding, and idempotency rules;
economic outcomes are evaluated in the subsequent product studies.}
\label{tab:extension-evidence}
\begin{tabularx}{\linewidth}{@{}p{.26\linewidth}X@{}}\toprule
Rule tested & Result\\\midrule
Forward payment and retry & Quantity 10 and an index change from 1.0 to 1.3 produce one 3.000000 transfer.\\
Funding and restart & A 0.060000 transfer leaves 0.040000; retry changes nothing and a further unfundable transfer is rejected.\\
Payment after a right is sold & Subsequent revenue follows the current holder rather than the former investor.\\
Resale to multiple holders & Resale to two holders across organizations preserves revenue allocation.\\
Covered service & A claim for a model service not covered by the contract is rejected.\\
Payment in full & One 60.000000 payment leaves 40.000000; another 60.000000 claim is not partially paid.\\
Concurrent claims & Exactly one competing claim is paid; the compensation funds are not overdrawn.\\
\bottomrule\end{tabularx}
\end{table}

\paragraph{Revenue-allocation precision.}
A fixed sequence of twelve receipts and twelve duplicate submissions
validates revenue allocation at the supported monetary precision. The
check associated with the 624-case study records fifteen exact matches
between reference calculations and account changes; the capital
replication adds fifteen exact matches. These checks establish the tested payment and duplicate-request
behavior independently of model-task outcomes.

\paragraph{Concurrent transaction execution.}
At concurrency eight, 760 operations across 19 scenarios preserve the
checked accounting rules, with no unexpected operation failures, ledger
discrepancies, or deadlocks. The highest scenario-level p95 latency is
2,419.98\,ms. Appendix~\ref{app:implementation-evidence} reports the full
latency distribution, reference budget, shared-host configuration, and
recovery tests alongside the accounting checks.
\FloatBarrier

\subsection{API-Price Forwards}
\label{sec:eval-forwards}
Forwards link a separate settlement payment to the difference between the
agreed and reference prices. We evaluate settlement execution and the
resulting expenditure using matched model-call workloads
(Equations~\eqref{eq:forward-hedged-cost}
and~\eqref{eq:forward-cost-decomposition}). All 84 zero-discount contracts
in the initial controlled study reconcile in amount, direction, fees,
and single settlement; 84 retail contracts provide a separate execution
check. Notionals are fixed from calibration before task execution, and
settlement applies the declared final indices to measured demand.
Appendix~\ref{app:eval-forwards} includes funded and unfunded payment tests.

\paragraph{Price-path and exposure design.}
The principal resampling study uses strike one, zero discount, and a 1.2\%
fee to isolate price-linked settlement from negotiated discounts. Daily
bills and prices are jointly resampled in seven-day blocks; six blocks
form each 42-day path, and 1,000 paths characterize the supplied trace
under each assumed price pattern. Hedge ratios
$h\in\{0,.25,.5,.75,1\}$ scale forecast expenditure. Forecast multipliers
$\{.5,1,1.5\}$, matching or constant reference indices, and all-or-none
funding of positive daily payments yield 300 settings across five price
patterns. The daily payment stress is an analytical funding assumption.

We report mean expenditure, standard deviation, mean expenditure in the
50 most expensive paths, and the fraction of paths exceeding the fixed
budget of 42 calibration-mean daily bills. These measures characterize
cost, variability, tail expenditure, and budget adherence within the
resampled paths.

\begin{table}[!htb]
\centering\small
\caption{Spending with no assumed price discount, a matching reference index,
the unchanged calibration forecast, and sufficient funds for payments.
$h$ is the fraction of forecast spending covered. Negative mean $\Delta C$
means lower spending than no hedge. Results come from resampled 42-day paths.}
\label{tab:zero-discount-risk}
\begin{tabular}{@{}lrrrr@{}}\toprule
Scenario & $h$ & Std. dev. (USD) & Mean $\Delta C$ (USD) & Over budget (\%)\\\midrule
flat & 0 & 1144.71 & +0.00 & 62.2\\
flat & 1 & 1144.71 & +26.49 & 63.3\\
rise & 0 & 1152.45 & +0.00 & 72.1\\
rise & 1 & 1190.82 & -304.88 & 61.4\\
fall & 0 & 1139.84 & +0.00 & 54.1\\
fall & 1 & 1100.08 & +357.85 & 65.5\\
spike & 0 & 1107.94 & +0.00 & 72.7\\
spike & 1 & 1164.28 & -304.10 & 60.7\\
lagged-load & 0 & 1150.50 & +0.00 & 63.0\\
lagged-load & 1 & 1150.32 & +25.08 & 63.8\\
\bottomrule\end{tabular}
\end{table}

\paragraph{Expenditure reduction and budget adherence.}
Under rising prices, full forecast coverage reduces mean expenditure by
USD~304.88 and the fraction of over-budget paths from 72.1\% to 61.4\%.
The corresponding standard deviation is USD~1,190.82, compared with
USD~1,152.45 without coverage. Under a price spike, mean expenditure falls
by USD~304.10 and over-budget paths decline from 72.7\% to 60.7\%.
Table~\ref{tab:zero-discount-risk} also reports the constant, falling,
and load-dependent paths, separating each expenditure and variability
measure for the same contract configuration.

\begin{figure}[!htb]
\centering\includegraphics[width=\linewidth]{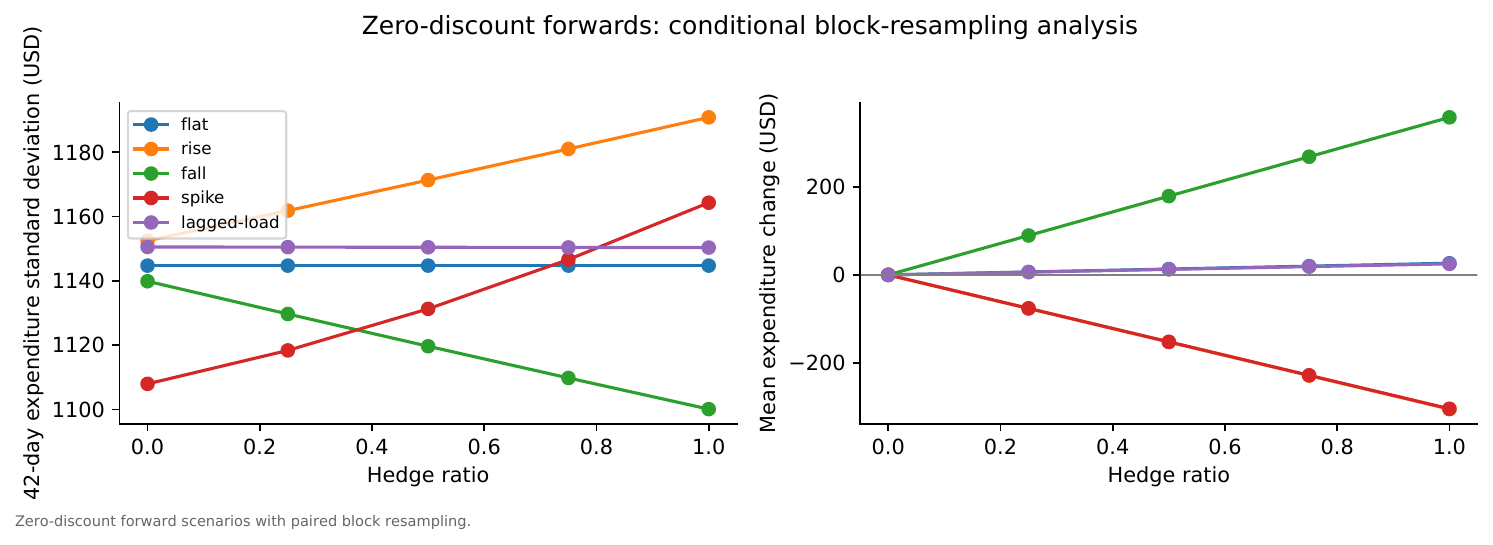}
\caption{Average spending and risk measures without an assumed price discount.
Bills and prices are jointly resampled in seven-day blocks. This plot uses
the unchanged-forecast, funded-payment settings; the full analysis
covers all 300 settings.}
\label{fig:zero-discount-risk}
\end{figure}
\FloatBarrier

\paragraph{Mechanism of expenditure variability.}
Let $C_0$ be unhedged spending and $Z=V-f$ the signed settlement net of
all forward fees. The change in expenditure variance is
\begin{equation}
\operatorname{Var}(C_0-Z)-\operatorname{Var}(C_0)
=\operatorname{Var}(Z)-2\operatorname{Cov}(C_0,Z).
\label{eq:eval-forward-variance}
\end{equation}
Variance decreases when
$2\operatorname{Cov}(C_0,Z)>\operatorname{Var}(Z)$: larger bills must be
accompanied by sufficiently offsetting net payments. In the rising-price
paths, $2\operatorname{Cov}(C_0,Z)=-85{,}383.2$ and
$\operatorname{Var}(Z)=4{,}522.5$ USD$^2$, yielding a variance change of
+89,905.7 USD$^2$. The decomposition explains how the measured reduction
in mean expenditure coexists with the observed variance response.
The complete post-hoc moment analysis appears in
Appendix~\ref{app:eval-forwards}.

\paragraph{Contract alignment and execution optimization.}
Price and quantity alignment connect the contract to actual purchases
through Equation~\eqref{eq:forward-cost-decomposition}. Request-weighted
purchase prices can differ from daily reference averages, and realized
usage can differ from the calibration notional. The sensitivity study
therefore varies these quantities explicitly.

In the separate constant-price replay, an assumed 10\% contract discount
reduces expenditure by USD~155.39 (4.58\%); the zero-discount control has
a 0.62\% expenditure increase from charges. With identical cache/batch
assumptions in both arms, adding the discounted forward reduces the
optimized unhedged expense by a further USD~113.04 (4.67\%). A conventional
forward with identical terms produces the same settlement, providing an
arithmetic control. Together, these comparisons identify the financial
contribution of the stated contract terms alongside execution optimization.
\FloatBarrier

\subsection{Revenue-Right Financing}
\label{sec:eval-financing}
Revenue-right financing exchanges current funding for a share of future
service receipts. Equation~\eqref{eq:financing-growth-threshold} identifies
the additional revenue needed to cover that share and execution expense.
We assess this relationship with five freshly executed portfolios for
each of three financing conditions at three starting balances. Funding
and settlement use the implemented contracts. Agent-issued rights are
evaluated with real-model execution; the supporting request replay
also analyzes Provider financing.

\begin{table}[!htbp]
\centering\small\setlength{\tabcolsep}{4pt}
\caption{Five separate runs per setting. Completed services are means out of
twelve. Own, Equal, and Right denote own capital, equal capital without
investor sharing, and revenue-right financing. Contribution differences are
experimental USD, excluding initial funding and credits. Brackets show the
range due to missing usage, not variation across repetitions.}
\label{tab:finrep-capital}
\begin{tabular}{@{}lrrrrr@{}}\toprule
& \multicolumn{3}{c}{Completed services} & \multicolumn{2}{c}{Contribution difference}\\
\cmidrule(lr){2-4}\cmidrule(l){5-6}
Capital & Own & Equal & Right & Right--own & Right--equal\\\midrule
$0.5B$ & 0.0 & 10.4 & 10.2 & 1.0635 & -0.3049\\
$B$ & 10.6 & 12.0 & 12.0 & -0.1406 & -0.3174\\
$1.5B$ & 12.0 & 12.0 & 12.0 & -0.2962 & [-0.2742, -0.0975]\\
\bottomrule\end{tabular}
\end{table}

\paragraph{Capital-constrained service delivery.}
At $0.5B$, revenue-right financing increases mean completion from 0.0 to
10.2 services and mean operating contribution by 1.0635 experimental USD
relative to own capital. Mean admission stops decrease from 12.0 to 1.8,
an 85\% reduction. All five low-capital contribution contrasts are positive.
At $B$, mean completion rises from 10.6 to 12.0, with a contribution
difference of -0.1406. At $1.5B$, all conditions complete twelve tasks and
the contribution difference is -0.2962. The five baseline contrasts
contain two positive and three negative values; all five ample-capital
contrasts are negative. These matched conditions locate the benefit of
financing where available capital constrains service delivery.

\paragraph{Revenue growth and allocation.}
Let $s_R,s_O$ count successful services and $C_R,C_O$ denote inference
expense under revenue-right and own-capital operation. Each success earns
$r$; both conditions pay a 1\% platform share, and financing assigns 10\%
to the investor. The contribution difference decomposes as
\begin{equation}
\begin{aligned}
\Delta\Pi_{R-O}
 &=0.89rs_R-0.99rs_O-(C_R-C_O)\\
 &=\underbrace{0.99r(s_R-s_O)-(C_R-C_O)}_{\text{change before investor share}}
   -\underbrace{0.10rs_R}_{\text{investor share}}.
\end{aligned}
\label{eq:eval-financing-decomposition}
\end{equation}
The first term captures additional retained revenue and execution expense;
the second records the investor allocation across financed receipts.
This decomposition uses the system's payment-rounding convention and the
observed model executions. At ample capital, equal completion isolates
the allocation cost of financing; at low capital, newly admitted work
provides the revenue supporting its positive contribution.

The equal-capital control matches initial funds without investor sharing.
Its mean contribution exceeds the revenue-right condition by 0.3049 at
low capital and 0.3174 at baseline. The ample-capital Right--equal range
is [-0.2742, -0.0975]. This experimental control separates capital
availability from the sharing obligation. Full missing-cost bounds and
portfolio-resampling ranges appear in
Appendix~\ref{app:eval-financing}.

\begin{figure}[!htbp]
\centering\includegraphics[width=\linewidth]{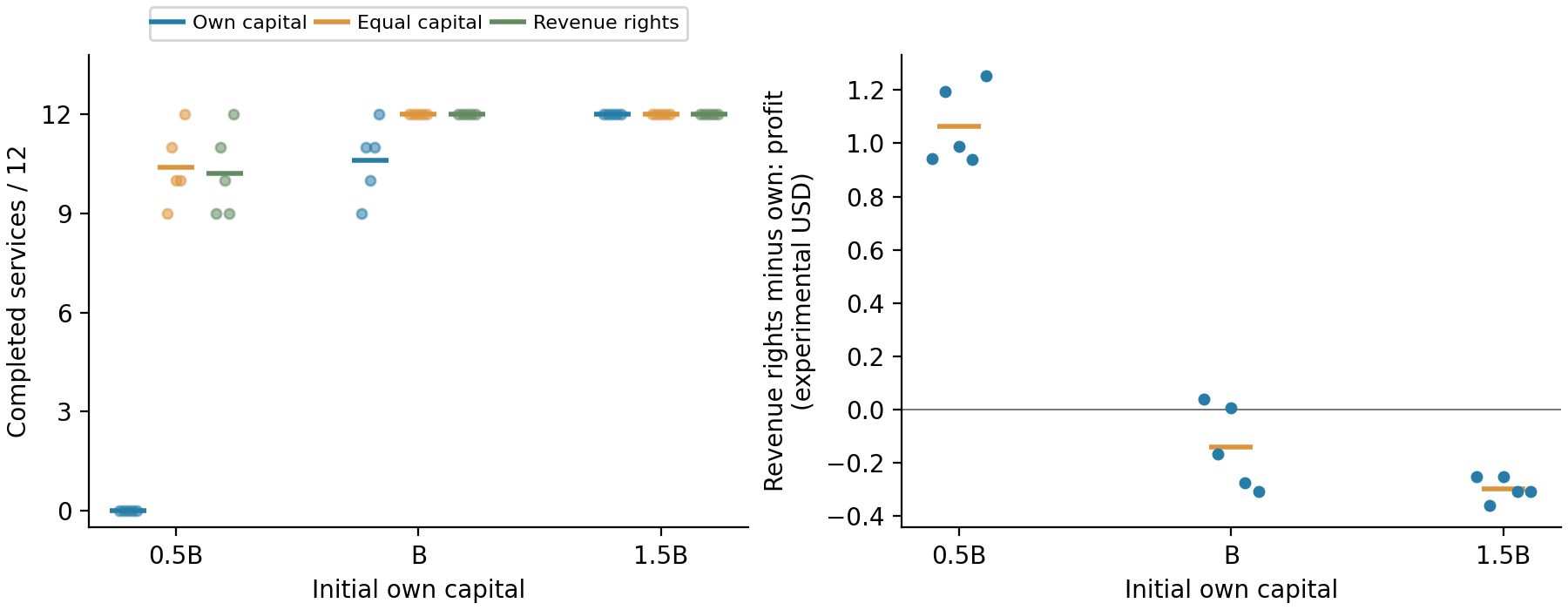}
\caption{Five runs at each starting capital level. Left: completed services
under the three financing settings, with mean bars. Right: contribution
with revenue-right financing minus contribution with own capital. Any
nonzero whiskers bound missing charges; they are not resampling ranges.
Tasks within a run share funds and are not independent repetitions.}
\label{fig:finrep-capital}
\end{figure}

\paragraph{Advance reservations and admission.}
The median reservation-to-measured-charge ratio is 12.0153 in the 624-case
study and 12.4382 among returned-usage requests in the capital replication.
The latter records have a minimum reservation of 0.545800, compared with
the low-capital opening balance of 0.535495. These observations link the
financing result to the tested advance-reservation rule as well as to
inference expense. The diagnostic describes returned-usage records;
rejected-request reservations and alternative reservation policies are
outside its observed population.

\paragraph{Supporting cohorts and operating conditions.}
The initial controlled study completes three additional services with
revenue-right financing and records a contribution increase of 0.1463
experimental USD. The earlier five-portfolio study, including its
incident-affected run, has a mean Right--own bound of
[-0.5295, -0.3411]. In the one-factor sensitivity study, the low-capital
condition is the one of eight displayed conditions with a positive lower
contribution bound. These separate cohorts characterize the dependence
on starting capital, task receipts, receipt timing, and realized execution.

The retail study records one successful case among 70, with zero successes
in its three financing arms. Equal capital increases scoreable completions
from one to four; the revenue-right arm has no scored completion and
retains 0.172938 in unresolved reservations. Consequently, this cohort
characterizes admission and expenditure without success-triggered revenue.
Its complete results remain in Appendix~\ref{app:eval-financing}.

\paragraph{Investor cash flows and funding-charge controls.}
In the capital replication, the investor contributes 0.5200 and receives
mean wallet transfers of 0.2540 at low capital and 0.3000 at baseline
and ample capital. Outstanding entitlements and rounding remainders are
tracked separately under the observation-window convention in
Section~\ref{sec:eval-measurement-scope}.

A separate replay fixes aggregate funding at USD~1,710.86. Assumed charges
are USD~16.87 for own-capital opportunity cost, USD~33.98 for fixed credit,
USD~11.25 for Provider credit, and USD~526.48 for revenue sharing. This
comparison isolates charging rules at matched funded volume; repayment,
collateral, and default-risk equivalence are outside that comparison.
An equal-funded loan is also included in the synthetic accounting tests
(Appendix~\ref{app:implementation-evidence}). Together, the growth and
charge controls quantify the service revenue required by specified
financing terms.
\FloatBarrier

\subsection{Service-Failure Protection}
\label{sec:eval-protection}
Recovery and protection address complementary outcomes: completed service
and allocation of covered residual loss. We vary local recovery through
fresh task execution and apply alternative protection plans to each
recorded trajectory, keeping execution fixed within a protection comparison.

The fault study schedules zero, six, or twelve local faults across twelve
tasks. Before each run, twelve alternative plans cross per-claim limits
of 0.03/0.13, premiums of 0.01/0.10/0.60, and Provider-funded reserves of
0.13/1.56. Each plan has separate accounts and whole-claim payment checks.
Policy results are reported separately. Premiums and service credits are
accounted for outside the balance used for task admission.

\paragraph{Service completion and eligible residual loss.}
With twelve scheduled faults, local recovery increases completion from
zero to nine services; with six faults, it increases completion from six
to ten (Table~\ref{tab:robust-protection}). Recovery uses an alternate local
route to the same upstream Provider; common-route outages remain terminal.
Compensation applies to covered terminal injected failures with known
expense. Semantic task errors and missing usage follow separate accounting
categories. Appendix~\ref{app:eval-protection} reports the initial
controlled and retail interventions using these definitions.

For eligible residual loss $L_e$, premium $P$, and issued credit $G$, a
buyer able to use fraction $v$ of the credit bears $L_e+P-vG$. Relative
to the same recovery policy without protection, its benefit is
\begin{equation}
B_{\mathrm{protection}}=vG-P.
\label{eq:eval-protection-value}
\end{equation}
We evaluate assumed utilization fractions $v=0$, $0.5$, and $1$.
$L_e$ includes losses within the coverage rules. At full utilization,
the benefit $G-P$ is the reserve-funded transfer net of premiums,
with execution expenditure held fixed.

\begin{table}[!htbp]
\centering\small\setlength{\tabcolsep}{4pt}
\caption{Fault tests with a per-claim limit of 0.13, premium of 0.01, and initial
reserve of 1.56. Loss and credit amounts are experimental USD. Paid credit
is issued through the experimental contracts; benefit assumes either full
use ($v=1$) or no use ($v=0$) of that credit.}
\label{tab:robust-protection}
\begin{tabular}{@{}lrrrrr@{}}\toprule
Faults / recovery & Success & Eligible loss & Paid credit & Benefit $v=1$ & Benefit $v=0$\\\midrule
0 / no & 12 & 0.0000 & 0.0000 & -0.0100 & -0.0100\\
0 / yes & 12 & 0.0000 & 0.0000 & -0.0100 & -0.0100\\
6 / no & 6 & 0.2900 & 0.2900 & 0.2800 & -0.0100\\
6 / yes & 10 & 0.0900 & 0.0900 & 0.0800 & -0.0100\\
12 / no & 0 & 0.5500 & 0.5500 & 0.5400 & -0.0100\\
12 / yes & 9 & 0.1400 & 0.1400 & 0.1300 & -0.0100\\
\bottomrule\end{tabular}
\end{table}

\begin{table}[!htbp]
\centering\small
\caption{Claim limits and available funds with twelve scheduled faults and a
premium of 0.01. Approved, paid, and unpaid columns report total amounts
in experimental USD, not numbers of claims. Initial funds exclude the
premium. Loss above the claim limit is not counted as an unpaid approved claim.}
\label{tab:robust-reserves}
\begin{tabular}{@{}lrrrrr@{}}\toprule
Recovery & Claim limit & Initial funds & Approved & Paid & Unpaid\\\midrule
no & 0.1300 & 1.5600 & 0.5500 & 0.5500 & 0.0000\\
no & 0.1300 & 0.1300 & 0.5500 & 0.1300 & 0.4200\\
no & 0.0300 & 0.1300 & 0.3600 & 0.1200 & 0.2400\\
no & 0.0300 & 1.5600 & 0.3600 & 0.3600 & 0.0000\\
yes & 0.1300 & 1.5600 & 0.1400 & 0.1400 & 0.0000\\
yes & 0.1300 & 0.1300 & 0.1400 & 0.1400 & 0.0000\\
yes & 0.0300 & 0.1300 & 0.0900 & 0.0900 & 0.0000\\
yes & 0.0300 & 1.5600 & 0.0900 & 0.0900 & 0.0000\\
\bottomrule\end{tabular}
\end{table}

\paragraph{Reserve capitalization and settlement coverage.}
With twelve faults, no recovery, and a 0.13 per-claim limit, an initial
reserve of 1.56 funds all 0.5500 in approved compensation. With recovery,
both tested reserve levels fund all 0.1400 in approved claims
(Table~\ref{tab:robust-reserves}). Reducing the initial reserve to 0.13
without recovery yields 0.1300 paid and 0.4200 unpaid; reducing the
claim limit to 0.03 yields 0.3600 approved. These controls distinguish
the effect of coverage limits on obligations from the effect of reserves
on settlement.

Premiums increase both reserve balance and budget. For example, 0.13 in
initial funds plus a 0.01 premium makes 0.14 available. Settlement checks
the full approved amount for each submitted claim and reports unpaid
amounts separately. The experiment applies per-claim full-payment checks;
the analytical replay below specifies its own carry-forward schedule.

\begin{figure}[!htbp]
\centering\includegraphics[width=\linewidth]{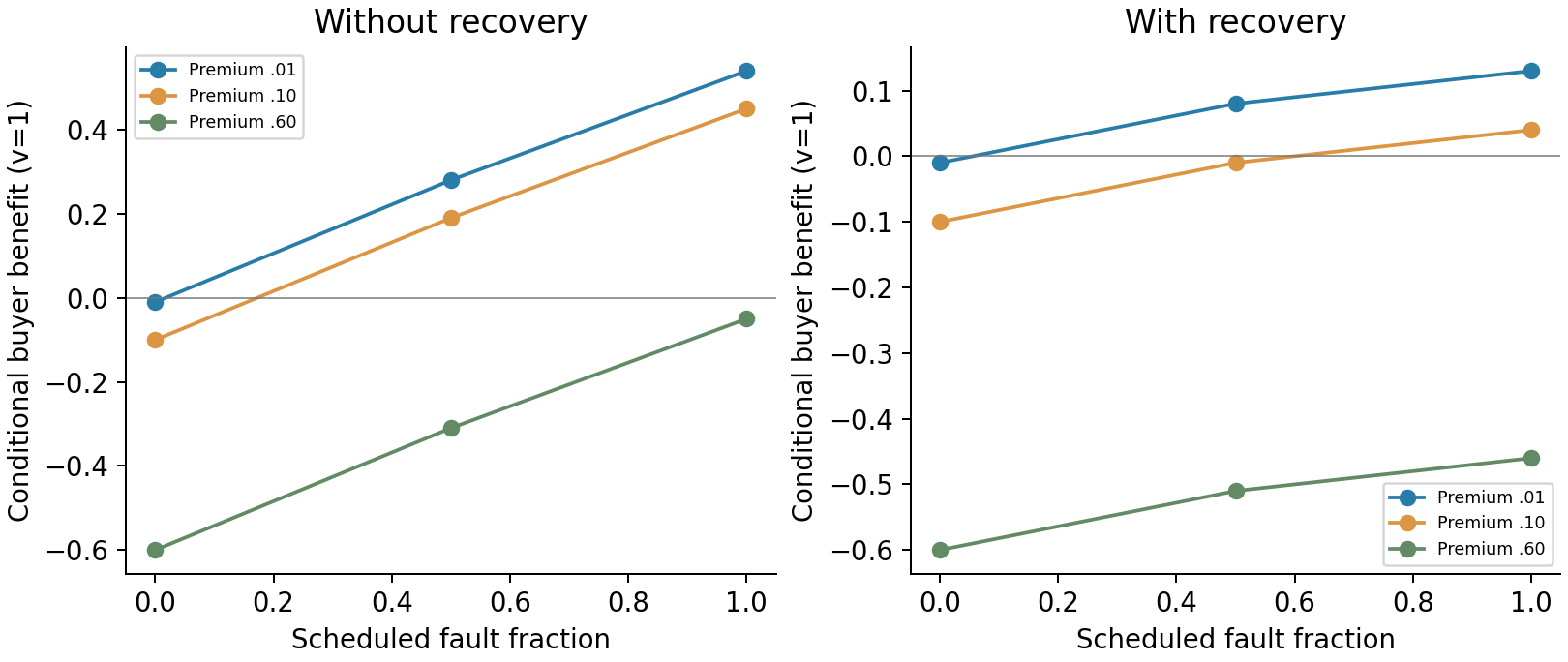}
\caption{Premium sensitivity of buyer value under full service-credit utilization,
a 0.13 per-claim limit, and a 1.56 initial reserve. Net value follows
Equation~\eqref{eq:eval-protection-value}.}
\label{fig:robust-protection}
\end{figure}

\paragraph{Premium and credit-utilization sensitivity.}
At full utilization, the ample-reserve fault scenarios yield net buyer
benefits of 0.0800--0.5400 with the 0.01 premium. Break-even occurs at
$P=vG$; raising premiums or lowering utilization changes net value
according to Equation~\eqref{eq:eval-protection-value}. With no eligible
fault or with $v=0$, the net value equals the premium outflow.
Figure~\ref{fig:robust-protection} reports the premium sensitivity under
full utilization, complementing the utilization cases in
Table~\ref{tab:robust-protection}.

\paragraph{Portfolio-level reserve sensitivity.}
The supporting request replay analyzes chronological liabilities under
a portfolio-level payment cap and explicit carry-forward of unpaid
amounts. These analytical conventions complement the per-claim
settlement experiments above. At a shared-failure fraction $\rho=0.75$,
the replay collects USD~34.22 in premiums and pays USD~220.73, reducing
modeled buyer expense by USD~186.51 under full credit utilization.
Recovery is held fixed for this financial comparison.
Appendix~\ref{app:eval-protection} specifies claim delays, reserves,
and liability limits for the analytical study.

\subsection{Combined Financial Effects}
\label{sec:eval-combined}
We evaluate the eight combinations of forwards ($D$), financing ($F$),
and protection ($S$) through a common accounting decomposition. Financing
conditions use separate executions because funds affect admission.
Forward and protection payments are applied to matched recorded
trajectories through separate accounts, holding admission fixed within
those comparisons. Contribution includes signed forward payments and
fees; service credits are reported separately. This design evaluates
combined financial accounting on measured execution.

\begin{table}[!htbp]
\centering\small
\caption{Eight combinations for recovery-enabled runs with ample funds, at an
assumed price index of 1.5 and hedge ratio of 0.5. $D$, $F$, and $S$ switch
forwards, financing, and protection on or off. Contribution is experimental
USD, excluding initial funding and service credits. Issued credits are
reported separately.}
\label{tab:controlled-combined}
\begin{tabular}{@{}rrrrrr@{}}\toprule
$D$ & $F$ & $S$ & Success & Contribution & Service credit\\\midrule
0 & 0 & 0 & 9 & 0.6034 & 0.0000\\
0 & 0 & 1 & 9 & 0.5934 & 0.1400\\
1 & 0 & 0 & 9 & 0.9772 & 0.0000\\
1 & 0 & 1 & 9 & 0.9672 & 0.1400\\
0 & 1 & 0 & 9 & 0.3735 & 0.0000\\
0 & 1 & 1 & 9 & 0.3635 & 0.1400\\
1 & 1 & 0 & 9 & 0.7474 & 0.0000\\
1 & 1 & 1 & 9 & 0.7374 & 0.1400\\
\bottomrule\end{tabular}
\end{table}

All controlled combinations complete nine tasks. At the assumed index
of 1.5 and hedge ratio of 0.5, adding the forward raises own-capital
contribution from 0.6034 to 0.9772. Protection issues 0.1400 in service
credit for a 0.01 premium. The all-mechanism configuration records
0.7374 in contribution and 0.1400 in credits. With ample funds and
matched completion, the financing comparison measures the share of
receipts allocated to investors. These results separate contract-specific
payments within the combined outcome.

The retail combinations retain their own execution trajectories, known
expenses, and missing-usage reservations
(Appendix~\ref{app:eval-combined}). Their eight known-cost contributions
are negative; within each fixed financing trajectory, the forward and
protection contrasts identify their respective settlement and fee effects.
Between-financing differences also incorporate the separate model executions.

\paragraph{Fixed-volume economic decomposition.}
The request replay crosses eight combinations, five price paths, and two
forward discounts while holding offered work and recovery fixed. With
$F=0$, funding has the assumed opportunity cost of own capital; with
$F=1$, the stated revenue-sharing charge replaces that cost. Modeled
\emph{economic expense} is
\begin{equation}
\begin{split}
C={}&C_{\mathrm{API}}+C_{\mathrm{retry}}+C_{D,\mathrm{fee}}-V_D\\
   &+C_{\mathrm{fund}}+P_S-A_S,
\end{split}
\label{eq:eval-economic-cost}
\end{equation}
Here $C_{\mathrm{API}}$ and $C_{\mathrm{retry}}$ are primary and alternate
request expenses, $C_{D,\mathrm{fee}}$ the forward fee, $V_D$ signed paid
settlement, $C_{\mathrm{fund}}$ funding cost, $P_S$ the protection premium,
and $A_S$ paid compensation. This metric includes opportunity cost and
values paid compensation at face value; initial capital is excluded.

Under constant prices, the assumed 10\%-discount forward reduces economic
expense from USD~3,438.59 to USD~3,282.46. The all-mechanism configuration
has expense USD~3,605.09, or 4.84\% above baseline, because the fixed
workload provides no additional admitted service to offset financing and
protection charges. The decomposition attributes this difference to
contract-specific costs and payments under a common workload. Together
with the capital-constrained and fault studies, it identifies the operating
conditions that give each financial mechanism a productive role.
\FloatBarrier

\section{Conclusion}
This paper presented \system{}, a unified contractual and accounting
infrastructure for AI services, integrating service commitments,
API-price forwards, revenue-right financing, and service-failure
protection.

Evaluation combines request-level replay over 362,417 calibration and
899,441 evaluation requests with real-model Agent execution and
PostgreSQL-backed contract experiments. Under the rising-price scenario,
forwards reduce mean API expenditure by USD~304.88. In the low-capital
setting, revenue-right financing increases mean contribution profit by
1.0635 experimental USD and reduces admission stops from 12.0 to 1.8.
Under controlled faults, recovery completes 9 of 12 tasks in each
recovery-enabled arm, while adequately funded protection settles all
approved compensation.

\section*{Authors and Contributors} \textbf{Cary Chang} served as the contributor to this work, leading the conceptualization,  system design, and manuscript preparation. \textbf{Jialin Zhou} contributed to project support. We also thank members of the broader research and developer community for discussions, feedback, and ideas that helped refine the research questions, system design, and evaluation presented in this report.
\begingroup\small
\setlength{\bibsep}{3pt}
\bibliographystyle{plainnat}
\bibliography{references}
\endgroup
\clearpage
\appendix
\section{Contract Records and Detailed Validation}
\label{app:claim-details}

This appendix formalizes the record structure, fulfillment conditions, and
transaction constraints underlying Section~\ref{sec:representation}.
The definitions distinguish contractual entitlements, current ownership,
and completed payments while preserving the product-specific rules for
consumption, forwards, revenue rights, and protection.

\subsection{Record Model}
\label{app:claim-record-model}

Let
\begin{equation}
\mathcal{T}=\{\mathrm{consumption},\mathrm{forward},
\mathrm{revenue},\mathrm{protection}\}.
\end{equation}
A claim's complete record description is
\begin{equation}
c=(\iota,\eta,\tau,\mathcal{P},\mathcal{R},\mathcal{U},
\Theta_{\tau},x_{\tau},\mathcal{L}),
\label{eq:claim-tuple}
\end{equation}
where $\iota$ identifies the record or linked record group, $\eta$ identifies
its tenant context (the account and permission scope), and $\tau\in\mathcal{T}$
is the economic type. The participant mapping $\mathcal{P}$ associates roles
with accounts and entities. References $\mathcal{R}$ identify the service,
price index, revenue policy, or protection plan; $\mathcal{U}$ specifies
quantity and payment units. Terms $\Theta_{\tau}$ describe the obligation,
state $x_{\tau}$ its status and operational quantities, and $\mathcal{L}$
links source, fulfillment, transfer, and settlement records.

The claim space is
\begin{equation}
\mathcal{C}=\bigsqcup_{\tau\in\mathcal{T}}\mathcal{C}_{\tau},
\label{eq:claim-space}
\end{equation}
where $\mathcal{C}_{\tau}$ contains claims of type $\tau$. The disjoint union
$\bigsqcup$ preserves the type distinction even when fields or units coincide.
Separate models, references, and validation routines realize this notation.
Product-specific validation determines how descriptive service fields enter
processing. Quality attributes specify the service terms associated with a
claim.

An amount $v=(a,u)$ contains numerical value $a$ and unit $u$. Supported
units include service credits, usage units, USD, and CNY. Ordinary transfers
require compatible units, while wallet and cross-organization operations
use explicit accounting links. Each amount retains its unit throughout
these operations. Time terms include prepaid
validity and optional expiry, forward dates, revenue-policy effective times,
optional funding-agreement start and end dates, and protection coverage,
waiting periods, and claim windows. Any finite duration or repayment cap is specified separately from the
revenue-share fraction.

\subsection{Product-Specific Fulfillment Rules}
\label{app:claim-fulfillment-rules}

\paragraph{Prepaid consumption and delivery.}
For prepaid lot $\ell$, let $r_{\ell}$ be the remaining amount,
$k_{\ell}$ its locked amount, $s_{\ell}$ the validity start, and
$e_{\ell}$ the expiry, with $e_{\ell}=\varnothing$ denoting no expiry.
For account $a$ and time $t$, eligible lots are
\begin{equation}
\begin{split}
\mathcal{E}_a(t)=\{\ell:\;&\operatorname{account}(\ell)=a,\;
\operatorname{status}(\ell)=\mathrm{active},\\
&s_{\ell}\leq t,\;
(e_{\ell}=\varnothing\lor t<e_{\ell}),\;r_{\ell}>0\}.
\end{split}
\end{equation}
Here, $\lor$ permits either no expiry or an expiry after the current time.
The available amount is
\begin{equation}
A_a(t)=\sum_{\ell\in\mathcal{E}_a(t)}\max(r_{\ell}-k_{\ell},0).
\label{eq:prepaid-available}
\end{equation}
A request consuming $q>0$ units requires allocations $z_{\ell}$ satisfying
\begin{equation}
\begin{split}
0\leq z_{\ell}&\leq\max(r_{\ell}-k_{\ell},0),\\
\sum_{\ell\in\mathcal{E}_a(t)}z_{\ell}&=q.
\end{split}
\label{eq:lot-allocation}
\end{equation}
Lots are selected by ascending expiry, with non-expiring lots last and
creation time breaking ties. Insufficient balance rejects consumption;
successful consumption reduces the selected balances and records the
allocations and accounting transaction.

For a preorder, purchased quantity $Q$, delivered quantity $D$, and additional
delivery $d>0$ satisfy $D+d\leq Q$ and $D'=D+d$. An Agent preorder becomes
delivered at $D'=Q$. These quantities establish recorded service delivery.
A prepaid transfer retains the source lot's expiry and refund eligibility.

\paragraph{Forward payment.}
For reference price $P$, fixed price $K$, and notional $N$ (the amount scaling
the price difference), the positions have values
\begin{equation}
V_{\mathrm{long}}=N(P-K),\qquad
V_{\mathrm{short}}=-V_{\mathrm{long}}.
\label{eq:claim-forward}
\end{equation}
The long account receives a positive amount and pays when that value is
negative; the short account has the opposite obligation. Price and notional
units must correspond as described in the forward analysis. Product checks
permit forwards, constrain $N$, and require future expiry within the maximum
term at creation. Fees and margin, funds held to support payment, are
recorded separately. Settlement uses the latest available index observation
and checks funding; insufficient funds produce a default record rather
than establishing a completed payment.

\paragraph{Revenue allocation.}
For revenue event $e$, let $R_e$ be its recorded net amount and $b_i$ the
share assigned to recipient $i$ among $m$ recipients. Shares use basis points,
each equal to $1/10{,}000$ of the amount. The policy requires
\begin{equation}
b_i>0,\qquad\sum_{i=1}^{m}b_i=10{,}000.
\label{eq:revenue-bps}
\end{equation}
The exact allocation is
\begin{equation}
d_{i,e}^{*}=\frac{b_i}{10{,}000}R_e.
\label{eq:revenue-entitlement}
\end{equation}
Both Agent and Provider settlement use the recorded net amount. At the
supported decimal precision, actual distributions $d_{i,e}$ are adjusted
to preserve the event amount:
\begin{equation}
d_{i,e}\geq0,\qquad\sum_i d_{i,e}=R_e.
\label{eq:revenue-conservation}
\end{equation}
For successive events and a fixed participant set with fixed shares, the
rounding difference carried to the next allocation is
\begin{equation}
\epsilon_{i,e+1}=\epsilon_{i,e}
+\frac{b_i}{10{,}000}R_e-d_{i,e},
\end{equation}
where $\epsilon_{i,e}$ is the previously accumulated difference. This records
the difference between exact entitlements and finite-precision payments.
Investor participants are updated before allocation using assets linked
to the funding agreement; if no matching assets exist, the original investor
is used. The recipient set is determined from the holdings selected at
settlement time.

\paragraph{Protection coverage and payment.}
For protection contract $c$ and incident/submission records $e$, coverage is
\begin{equation}
\begin{split}
\chi(c,e)={}&\chi_{\mathrm{active}}\land\chi_{\mathrm{period}}
\land\chi_{\mathrm{waiting}}\\
&\land\chi_{\mathrm{provider/model}}\land\chi_{\mathrm{event}}
\land\chi_{\mathrm{deadline}}.
\end{split}
\label{eq:coverage-predicate}
\end{equation}
Each term denotes a passed check: active subscription, covered incident
period, any waiting-period requirement, Provider or model scope, covered
event type, and claim deadline. The conjunction $\land$ requires every
applicable condition to hold.

For a cost-denominated compensation base $B$, let $\delta$ be the deductible
(the excluded amount), $\gamma$ the compensation rate, and $M$ the plan's
maximum amount.
Incident-impact calculations and ordinary manual claims use
\begin{equation}
\widehat{G}=\min\{M,\;\gamma\max(B-\delta,0)\}.
\label{eq:protection-amount}
\end{equation}
Impact records retain request costs together with failure and fallback
counts. The monetary interpretation above uses a cost-denominated $B$;
activity counts retain their event-count units.
Automatically generated claims already contain the calculated amount;
review caps that amount without applying the deductible and rate again.

For approved amount $G$, let $R^{\mathrm{budget}}$ be the reserve budget,
$R^{\mathrm{paid}}$ cumulative payments, $B_r$ its account balance, and
$L_r$ locked balance. Settlement requires
\begin{equation}
0<G\leq\min\{R^{\mathrm{budget}}-R^{\mathrm{paid}},\;B_r-L_r\}.
\label{eq:reserve-admissibility}
\end{equation}
Successful settlement credits the buyer's billing wallet for service charges.
An approved claim and an executed payment remain distinct records.

\subsection{Operation and Transaction Controls}
\label{app:claim-operation-controls}

For operation $\alpha$ on type $\tau$, the conditions summarized in
Equation~\eqref{eq:claim-transition} are
\begin{equation}
\begin{split}
G_{\tau,\alpha}={}&G_{\mathrm{authority}}\land G_{\mathrm{policy}}
\land G_{\mathrm{state}}\\
&\land G_{\mathrm{terms}}\land G_{\mathrm{resources}}.
\end{split}
\label{eq:claim-guards}
\end{equation}
These check permission, whether an operation is enabled or needs approval,
its current status, contract conditions, and resources such as balances,
margin, or reserves. The applicable checks differ by operation. For example,
an Agent preorder progresses through approval and recorded delivery, a
forward can settle or default, a revenue event settles after distribution,
and protection approval precedes a separate payment step.

Product records retain terms, participants, status, holdings, and source
links. Validation checks the required fields and conditions; transaction
routines lock affected records and invoke account or wallet operations.
Delivery, valuation, distribution, claim, transfer, and accounting records
retain links between each operation and its outcome. Market transfers
release the relevant payment reservation, post payment, decrease the seller's
total and locked units, increase the buyer's total and available units,
and record the transfer. Receiving holdings preserve selected agreement
and policy references for later recipient selection.

For account $a$, let $B_a$ and $L_a$ be its balance and locked balance,
and $C_a^{\mathrm{used}}$ and $C_a^{\mathrm{limit}}$ its used and permitted
credit amounts. Account constraints require
\begin{equation}
0\leq L_a\leq B_a,\qquad
0\leq C_a^{\mathrm{used}}\leq C_a^{\mathrm{limit}}.
\end{equation}
Product checks additionally enforce available prepaid amounts, remaining
delivery quantities, distributable revenue, and reserve funding. The
balanced-posting relation appears in Equation~\eqref{eq:ledger-balance};
cross-organization processing may require checking a linked settlement
group rather than an isolated account balance.

An \emph{idempotency key} associates a supported operation's retries with
its recorded result. The common transfer routine rechecks this key after
acquiring account locks. Alongside product-state checks, this supports at
most one committed financial posting for each scoped operation key.

\section{Forward Valuation, Funding, and Settlement}
\label{app:forward-details}

This appendix specifies the records and funding rules supporting the
forward mechanism. It distinguishes contractual valuation from completed
payment and derives the condition under which a funded forward reduces
expenditure variance.

\subsection{Reference, Formation, and Valuation Records}

\paragraph{Contract record.}
A forward contract is represented by
\begin{equation}
\begin{split}
f_{\mathrm{ctr}}=(
&a_L,a_S,\mathcal{I},N,K,t_0,T,\\
&m_0,m_1,\phi_L,\phi_S).
\end{split}
\label{eq:forward-contract}
\end{equation}
Here $a_L$ and $a_S$ are the long and short accounts, $\mathcal{I}$
identifies the reference-price index, $N$ is the submitted multiplier,
$K$ the fixed price, $t_0$ activation time, and $T$ recorded expiry.
The coefficients $m_0$ and $m_1$ govern initial and subsequent margin,
while $\phi_L$ and $\phi_S$ govern service fees. These numerical fields
use the product's quotation and payment units. The normalized notional
used in the expenditure analysis is related to the service quantity
by $N=qP_0$. This transformation defines the analytical units; submitted
contract fields use the product's quotation and payment units.

Reusable product records specify the reference index, allowed notional
range, maximum term, and fee and margin parameters. Individual contracts
retain the selected accounts, accepted terms, funding amounts, state,
and latest valuation. The benchmark-creation path associates an index
with a Provider marketplace offer having a positive price in USD per
thousand tokens. Observations retain their timestamps and sources.

Quote confirmation checks authority, quote state and deadline, and record
versions when supplied. It accepts the selected quotation and withdraws
competing offered quotations. A requester choosing the buy side becomes
the long account; otherwise, it becomes the short account. Funding
previews estimate requirements, but activation calculates them from the
accepted notional and product parameters. The coefficients used for fees
and margins are stored in basis points, each equal to $1/10{,}000$, and
converted to fractions under the product's unit convention.

Valuation selects the latest stored observation for the index within
the contract's tenant, its account and permission scope. When no observation
exists, the implementation can initialize one from configured source
information. The selected observation and its recorded source provide the
price evidence used by the workflow.

Valuation uses decimal arithmetic rounded to six decimal places and
retains the selected observation, price, amount, and margin requirement.
The valuation is the contract-defined price-difference amount. Funding
charges and scenario-level economic outcomes are recorded separately.

\subsection{Margin Requirements and Funding Checks}
\label{sec:forward-margin}

\paragraph{Initial funding.}
Activation requires sufficient funds for both margin and service fees:
\begin{align}
M_L^0&=m_0N, & M_S^0&=m_0N,
\label{eq:forward-initial-margin}\\
F_L&=\phi_LN, & F_S&=\phi_SN.
\label{eq:forward-service-fees}
\end{align}
The superscript $0$ denotes activation. Fees go to the platform account,
while margin remains reserved against other use. Product coefficients
must be interpreted with the relevant notional and payment units.

\paragraph{Subsequent requirements.}
For participant $j\in\{L,S\}$, let $B_{j,t}^{M}$ be its funded margin
at time $t$. Given valuation $V_t$, maintenance coefficient $m_1$, and
notional $N$, the contract-level requirement is
\begin{equation}
R_t=|V_t|+m_1N.
\label{eq:forward-contract-margin}
\end{equation}
The recorded account requirement and additional amount due are
\begin{align}
R_{j,t}&=\min\{R_t,\;B_{j,t}^{M}+|V_t|\},
\label{eq:forward-account-margin}\\
D_{j,t}&=\max\{R_{j,t}-B_{j,t}^{M},0\}.
\label{eq:forward-margin-shortfall}
\end{align}
Both participants' calculations use $|V_t|$ under the same margin policy.
The margin coefficients are configured contract parameters.

A positive $D_{j,t}$ can create or update a \emph{margin call}, a request
for additional funding that records the amount, deadline, and status.
Enabled automatic funding attempts draw on linked billing wallets after
currency, availability, and balance checks. A separate default check
records an unmet obligation when funded margin is insufficient. Payment
remains subject to the current obligation and available funding.

\subsection{Settlement Funding and Idempotency}

Settlement accepts active or defaulted contracts and recalculates the
obligation using the latest stored observation at settlement. Let $B_a$
denote the payer's balance,
$L_a$ its total locked balance, and $M_a$ its funded margin for this
contract. After releasing the contract's reservation, available funds are
\begin{equation}
A_a^{\mathrm{available}}=B_a-\max\{L_a-M_a,0\}.
\label{eq:forward-settlement-available}
\end{equation}
This expression releases the contract's funds while retaining other
reservations; margin is not added again to the account balance.

Let $A_t=|V_t|$ be the payment due. The funding preview reports
\begin{align}
M_a^{\mathrm{applied}}&=\min\{M_a,A_t\},\\
C_a^{\mathrm{additional}}&=\max\{A_t-M_a^{\mathrm{applied}},0\},\\
S_a&=\max\{A_t-A_a^{\mathrm{available}},0\}.
\label{eq:forward-settlement-shortfall}
\end{align}
These are the portion covered by margin, the portion beyond margin,
and the shortage after all available funds are considered. The second
amount need not imply a shortage: existing unlocked funds can cover it.

Settlement records link the payment and funding preview to the contract
and accounting transaction. Contract-specific identifiers associate
repeated requests with an existing settlement, and state checks prevent
settling an already completed contract again. Database transactions and
record locks coordinate the associated updates. Unused wallet-originated
funding can return through the wallet path, subject to its currency
precision. Insufficient funding produces no proportional partial payment.

\subsection{Expenditure Variability}
\label{sec:forward-variability}

With full payment, fixed $N$ and $K$, scenario-independent charges
$F+H$, and finite variances, the expenditure model gives
\begin{equation}
\begin{split}
\operatorname{Var}(C_H)
={}&\operatorname{Var}(C_0)+N^2\operatorname{Var}(P^{*})\\
&-2N\operatorname{Cov}(C_0,P^{*}).
\end{split}
\label{eq:forward-variance}
\end{equation}
Variance measures fluctuation across the modeled scenarios; covariance
measures how expenditure and the settlement reference price move together.
The contract reduces variance only when the covariance term offsets
the variation added by its price-dependent payment. This relationship
explains why price alignment matters independently of the agreed fixed
price. The identity provides the analytical basis for comparing reference
alignment and hedge ratios across scenarios.

\section{Revenue-Right Financing: Supporting Definitions}
\label{app:financing-details}

This appendix specifies the funding, revenue-allocation, and ownership
rules supporting Section~\ref{sec:revenue-right-financing}. It also defines
the workload-accounting models used to distinguish additional funded
service from the cost of sharing revenue.

\subsection{Contract Records and Capital Disbursement}

A publisher plan is represented as
\begin{equation}
\pi=(o,\mathcal{S},u,b_O,b_I,b_P,\mathcal{D}),
\label{eq:publisher-plan}
\end{equation}
where $o$ is the publisher, $\mathcal{S}$ its Agent or Provider service,
and $u$ the accounting unit. Shares $b_O,b_I,b_P$ are defined in
Section~\ref{sec:financing-contract}; $\mathcal{D}$ contains plan descriptions
and source-record references. The corresponding fractions are
$\alpha_O=b_O/10{,}000$, $\alpha_I=b_I/10{,}000$, and
$\alpha_P=b_P/10{,}000$. Agent plans require positive shares for all three
groups. Provider plans can omit an investor share, although financing
uses $b_I>0$. Provider service references identify the partner, the
runtime executing its calls, and the selected model offer.

An investment agreement records
\begin{equation}
\begin{split}
g_j=(&\iota_j,\pi,a_j,C_j,C_j^{\mathrm{settled}},\\
&\sigma_j,\mathcal{L}_j),
\end{split}
\label{eq:investment-agreement}
\end{equation}
where $\iota_j$ identifies the agreement and $a_j$ the investor account.
The amounts $C_j$ and $C_j^{\mathrm{settled}}$ denote committed and disbursed
capital, $\sigma_j$ is the current status, and $\mathcal{L}_j$ links funding,
disbursement, and derived revenue-right records. Separate Agent and
Provider revenue policies maintain the corresponding participant records.
Funding agreements identify publisher investments by their participants
and revenue policy. Publisher investments use immediate capital settlement
at confirmation.

Investment requests retain the accepted policy, publisher and investor
references, and billing-wallet links. Wallet-funded amounts must fit the
available balance and be exactly representable to two decimal places.
Confirmation checks authority, product policy, and the supplied record
version when present, under the relevant tenant contexts. A wallet bridge,
the routine connecting wallet operations to financial accounts, funds
the investor-side account and transfers the contribution to an intermediate
settlement account called escrow. The agreement then becomes active.

For an active Agent agreement with funded escrow, capital settlement
transfers $C_j^{\mathrm{due}}=C_j-C_j^{\mathrm{settled}}$ to the publisher's
funding account. Completion records $C_j^{\mathrm{settled}}=C_j$, the
transaction, and settlement time. The Provider path similarly transfers
capital and records the publisher-side receipt when funding crosses
tenants. These linked records connect the investor's outflow to the
publisher's capital receipt.

Escrow connects the investor-side funding operation to the publisher's
capital receipt. Disbursement is performed for each confirmed investment.
An Agent plan's funding goal is retained as plan information.

\subsection{Share Allocation and Revenue Processing}
\label{app:financing-allocation}

\paragraph{Integer agreement shares.}
Equation~\eqref{eq:financing-investor-share} defines ideal allocations.
For $n$ selected agreements in deterministic order, let
$A_{j-1}=\sum_{k=1}^{j-1}b_k$, with $A_0=0$, be the basis points already
assigned. For $j<n$, the stored allocation is
\begin{equation}
\begin{split}
b_j=\min\{&\max\{1,\lfloor b_j^{*}\rfloor\},\\
&b_I-A_{j-1}-(n-j)\},
\end{split}
\label{eq:financing-integer-share}
\end{equation}
where $\lfloor\cdot\rfloor$ rounds down. The second bound retains at least
one basis point for every remaining agreement. The final agreement receives
$b_n=b_I-A_{n-1}$. Positive allocations require $n\leq b_I$; the Provider
routine explicitly rejects more confirmed investments than the available
pool can accommodate at this precision. Holder shares use a second integer
allocation with remainder assignment. An agreement with more selected
holders than assigned basis points is rejected.

For example, adding an active 100-unit investment to the main text's
100- and 300-unit example changes the ideal total-revenue shares from
2.5\% and 7.5\% to 2\%, 6\%, and 2\%. The pool remains 10\%; the individual
percentages are recalculated rather than fixed at investment creation.

\paragraph{Revenue records and readiness.}
A revenue event identifies its policy, source and settlement accounts,
amounts, accounting unit, and supporting service references. Agent usage
capture retains request and usage identifiers together with consumer and
publisher tenant references. Provider capture also retains runtime and
model-offer information. The recorded billing amount is transferred to
a revenue source account, and an event identifier associates repeated
creation requests with that event.

Let $G_e$ be gross recorded revenue and $R_e$ the recorded net amount
used for distribution. Agent creation sets $R_e=G_e$. Provider creation
can accept a separate net amount, but the evaluated usage-capture procedures
supply the same captured amount as both gross and net. This captured income
is the distribution base; operating costs enter the separate contribution-
profit calculation.

Automatic settlement of a publisher plan with an investor allocation
requires a linked active investment with positive disbursed capital.
Otherwise, an event can remain pending, and confirmation of an investment
can trigger later settlement. Before distribution, recipients are refreshed
from the currently active funded investments and linked positive holdings.
The allocation convention uses the recipient set selected at settlement.

\paragraph{Payment precision and rounding carry.}
For participant $i$, let $b_{i,e}$ be the share used for event $e$ and
$\varepsilon=10^{-6}$ the payment increment. A \emph{rounding carry}
$\kappa_{i,e}$ retains the signed difference between previously calculated
shares and payments. The routine computes
\begin{align}
y_{i,e}&=R_e\frac{b_{i,e}}{10{,}000}+\kappa_{i,e},\\
d_{i,e}^{(0)}&=\varepsilon\left\lfloor
\frac{\max(y_{i,e},0)}{\varepsilon}\right\rfloor.
\end{align}
It then adjusts payments in increments of $\varepsilon$ to match $R_e$.
If their sum is too low, increments go to the largest unpaid remainders;
if too high, increments are removed from positive payments using the
corresponding remainder ordering. Deterministic participant order resolves
ties. For event amounts at the supported precision, payments satisfy
Equation~\eqref{eq:financing-event-conservation}, and
\begin{equation}
\kappa_{i,e+1}=y_{i,e}-d_{i,e}.
\label{eq:financing-carry}
\end{equation}
For an unchanged participant identity across $E$ events,
\begin{equation}
\begin{split}
\sum_{e=1}^{E}d_{i,e}
={}&\sum_{e=1}^{E}R_e\frac{b_{i,e}}{10{,}000}\\
&+\kappa_{i,1}-\kappa_{i,E+1}.
\end{split}
\end{equation}
The retained difference accounts for cumulative proportional amounts and
actual payments for the unchanged participant identity specified above.

\paragraph{Recording and wallet transfers.}
Settlement batches select pending events, refresh recipients, allocate
payments, and post transfers. Distribution records link the event,
participant, account, transaction, applied share, and carry; the event
then becomes settled. Event and account locks coordinate concurrent
updates. Batch and event--participant \emph{idempotency keys}, identifiers
associating repeated requests with existing operations, prevent duplicate
financial postings through the supported paths.

Investor distributions can be transferred to linked billing wallets,
whose amounts have two-decimal precision. For incoming distribution $d$
and retained remainder $\omega$, the bridge credits
\begin{equation}
w=0.01\left\lfloor\frac{\omega+d}{0.01}\right\rfloor,
\qquad \omega'=\omega+d-w.
\end{equation}
The remainder is retained for later transfers. For cross-tenant
investments, clearing and corresponding distribution records link the
publisher-side allocation to the investor-side receipt.

\subsection{Transfer Records and Settlement}
\label{app:financing-transfers}

A revenue-right asset represents a holding
\begin{equation}
a=(\iota_a,\eta_a,g,\pi,h,U,A,L),
\label{eq:secondary-revenue-asset}
\end{equation}
where $\iota_a$ identifies the asset and $\eta_a$ its tenant context.
The references $g$ and $\pi$ identify the source agreement and revenue
policy; $h$ is the holder account. Here $g$ is an agreement reference,
not the contribution fraction used in the economic analysis. Quantities
$U,A,L$ are total, available, and locked units. The last category records
units reserved against other use.

Creation requires a confirmed investment and, for publisher investments,
completed capital disbursement. Conversion creates a positive number of
units no greater than the source investment position. Source-agreement
and policy references and selected service attributes are retained.
Agent investment confirmation invokes asset creation; the Provider path
supplies a corresponding conversion routine. Provider listing requires
a Provider-financing source-agreement reference even though the broader
asset model contains additional right types. A numerical relationship
between units and investment amount establishes neither a resale value
nor a redemption price.

\paragraph{Reservation and order entry.}
Listing $q_{\ell}$ units requires authority and $0<q_{\ell}\leq A$.
The holding changes to
\begin{equation}
A'=A-q_{\ell},\qquad L'=L+q_{\ell},\qquad U'=U,
\label{eq:secondary-listing-reservation}
\end{equation}
where primes indicate the updated state. Both creation paths open active
listings and create or ensure the associated sell order at the listed
quantity and minimum price. Approval follows the applicable listing
workflow. Remaining listing and order quantities allow partial execution.

A buy order requires an active listing, quantity no greater than its
remaining units, and applicable access checks. Any supplied quote-version
identifier is checked. Buyer funds are reserved according to
Equation~\eqref{eq:secondary-buyer-reservation}, subject to decimal
precision and available balance. The Provider path rejects purchases by
the user owning the listing, and both matching paths reject identical
buyer and seller accounts. The checks apply to the account and ownership
identifiers associated with the submitted transaction.

The crossing checks are $p_b\geq p_s$ for the Agent path and
$p_b\geq\max\{p_s,p_{\ell}^{\min}\}$ for the Provider path. Together
with Equations~\eqref{eq:secondary-agent-price}--\eqref{eq:secondary-provider-price},
these preserve $p^{*}\leq p_b$. The listing-generated sell order has
$p_s=p_{\ell}^{\min}$, so both paths use the same minimum price in that
case. A positive match records the trade and reduces the remaining buy,
sell, and listing quantities; holdings and payment change at settlement.

\paragraph{Ownership and payment updates.}
For a trade backed by sufficient reserved seller units, settlement updates
\begin{align}
U_s'&=U_s-q^{*}, & L_s'&=L_s-q^{*},\\
U_b'&=U_b+q^{*}, & A_b'&=A_b+q^{*}.
\label{eq:secondary-holding-update}
\end{align}
The subscripts denote seller and buyer, giving
\begin{equation}
U_s'+U_b'=U_s+U_b.
\label{eq:secondary-unit-conservation}
\end{equation}
Settlement locks the relevant records, releases the needed buyer payment
reservation, and posts the buyer-to-seller transfer. It creates or locates
the buyer's holding using the source right type and identifiers. The
receiving holding preserves agreement and policy references. A transfer
record identifies both holdings and the quantity, while the settlement
record links the trade to its accounting transaction. A database
transaction connects these updates within \system{}.

The Agent path blocks disputed trades and releases surplus payment
reservations when a buy order is completed. The Provider path has its
own settlement and reservation-release logic. State checks and settlement
identifiers prevent supported retries from posting completed trades twice.

\paragraph{Recipients after transfer.}
The recipient lookup follows source-agreement references to current
positive holdings, including supported cross-tenant transfers. If no
derived holding exists, the original investor is used with units
corresponding to the agreement's budget. Generated participant records
retain the recipient account, tenant context, agreement, and asset
identifier. Source links connect the receiving holding to the original
investment without treating the resale payment as new publisher funding.

For agreement $j$ at settlement time $t_s$, its ideal holder fraction of
total plan revenue, after the agreement-level integer allocation but
before holder-level rounding, is
\begin{equation}
\alpha_{j,h}(t_s)=\frac{b_j(t_s)}{10{,}000}
\frac{u_{j,h}(t_s)}{\sum_{k\in\mathcal{H}_j(t_s)}u_{j,k}(t_s)}.
\label{eq:secondary-holder-entitlement}
\end{equation}
The ideal payment is $d_{j,h,e}^{*}=\alpha_{j,h}(t_s)R_e$. Holder shares
are converted to integer basis points before monetary allocation.
The recipients are the holders selected at $t_s$. Secondary-market
experiments measuring order arrivals,
execution rates, resale prices, and investment exit returns are outside
the financing evaluation.

\subsection{Workload Accounting}
\label{app:financing-workload}

Let $B_t$ be the operating budget and $D_t$ the cost of serving all
available demand in period $t$. A divisible-work approximation admits
\begin{equation}
X_t(B_t)=\min\{D_t,B_t\}.
\label{eq:financing-fluid-service}
\end{equation}
Here, served work is measured by its cost. The approximation can use
fractional requests; greater funding increases service only while
available demand exceeds the budget.

The complete-request replay instead retains the original request order.
For costs $c_{t,1},\ldots,c_{t,n_t}$ in period $t$, define
\begin{equation}
\begin{split}
k_t(B)=\max\biggl\{k\in\{0,\ldots,n_t\}:\\
\sum_{r=1}^{k}c_{t,r}\leq B\biggr\}.
\end{split}
\end{equation}
The case $k=0$ admits no requests. This selects the longest initial
sequence fitting the daily budget and excludes later requests once
the next one would exceed it. Across days,
\begin{equation}
X(B)=\sum_t\sum_{r=1}^{k_t(B)}c_{t,r}.
\label{eq:financing-prefix-service}
\end{equation}
Baseline and financed scenarios use $X_0=X(B)$ and $X_1=X(aB)$ for
$a\geq1$. Costs come from completed trace records, so the calculation
is retrospective rather than an online predictor of output cost.
Budgets reset daily by assumption, and $a$ is a specified budget multiplier.
These definitions evaluate the amount of existing demand that can be served
under each funding scenario.

\section{Implementation Validation and Synthetic Accounting Tests}
\label{app:implementation-evidence}

This appendix evaluates whether \system{} executes the contract and
accounting rules described in the main text. Database experiments test
transaction correctness and concurrent execution, while deterministic
synthetic tasks test cash-flow conservation under controlled funding and
failure conditions. Their contract-level assertions complement the
service and economic outcomes measured by the model-driven studies.

\subsection{Transaction Correctness and Ledger Reconciliation}
The backend regression suite passes all 206 tests. A focused suite passes
20 gateway and source-routing checks, and a contract-specific suite passes
seven experiments. These suites overlap with the regression tests and
with previously exercised contract lifecycles; their counts are therefore
not added as independent observations.

The contract-specific experiments execute 80 business API calls against an
isolated PostgreSQL database and record seven final accounting states.
Participant identities and initial funds are controlled experimental inputs;
subsequent economic operations use the implemented contract APIs.
Validation links account-balance changes to wallet records, ledger entries,
and cross-organization transactions. Monetary comparisons use exact
six-decimal arithmetic.

An independent decimal-arithmetic replay verifies 165 balance changes across
47 accounts, with no arithmetic discrepancies. Direct ledger association
covers 103 entries. For credit-funded movements, the verified quantity is
net position, defined as cash balance minus credit used. The five intermediate
wallet-to-account funding changes are documented through their supporting
wallet records. The validation therefore reports balance arithmetic and
record association at their respective observation scopes.
Table~\ref{tab:extension-evidence} summarizes the contract rules and assertions
covered by these experiments.

\begin{figure}[tb]\centering
\includegraphics[width=\linewidth]{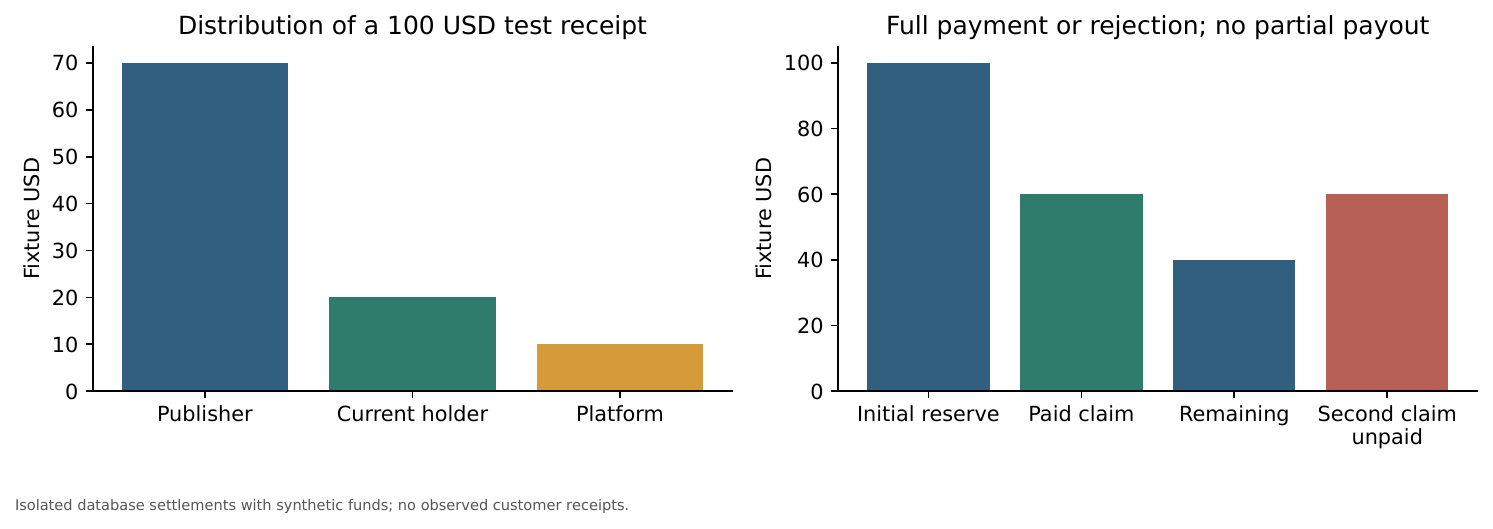}
\caption{Contract settlement in isolated database experiments: revenue
allocation after a rights transfer (left) and dedicated-reserve payments
(right). Controlled initial funds and synthetic revenue events exercise
the corresponding ownership and funding rules.}
\label{fig:contract-cashflows}\end{figure}

\subsection{Recovery Semantics and Concurrent Execution}
Recovery tests cover bounded retries, session renewal, and a simulated
30-day idle interval. After an inconclusive health probe, a previously
verified service remains degraded but eligible for routing. Services that
were never verified, explicitly stopped, disabled, or definitively failed
retain their respective routing restrictions. The simulated interval
exercises elapsed-time state transitions under these routing rules.

The concurrency experiment executes 760 operations across 19 scenarios
at concurrency eight. No unexpected operation failures, ledger discrepancies,
invariant violations, or deadlocks are observed. Eleven scenarios have
95th-percentile latency above the prespecified 1,500\,ms budget; the largest
measured value is 2,419.98\,ms. Accounting assertions and latency are reported
as separate measurements. The experiment uses one local Windows/Docker host
while other experiments are active, so the measurements characterize this
shared-host configuration.

\begin{figure}[tb]\centering
\includegraphics[width=\linewidth]{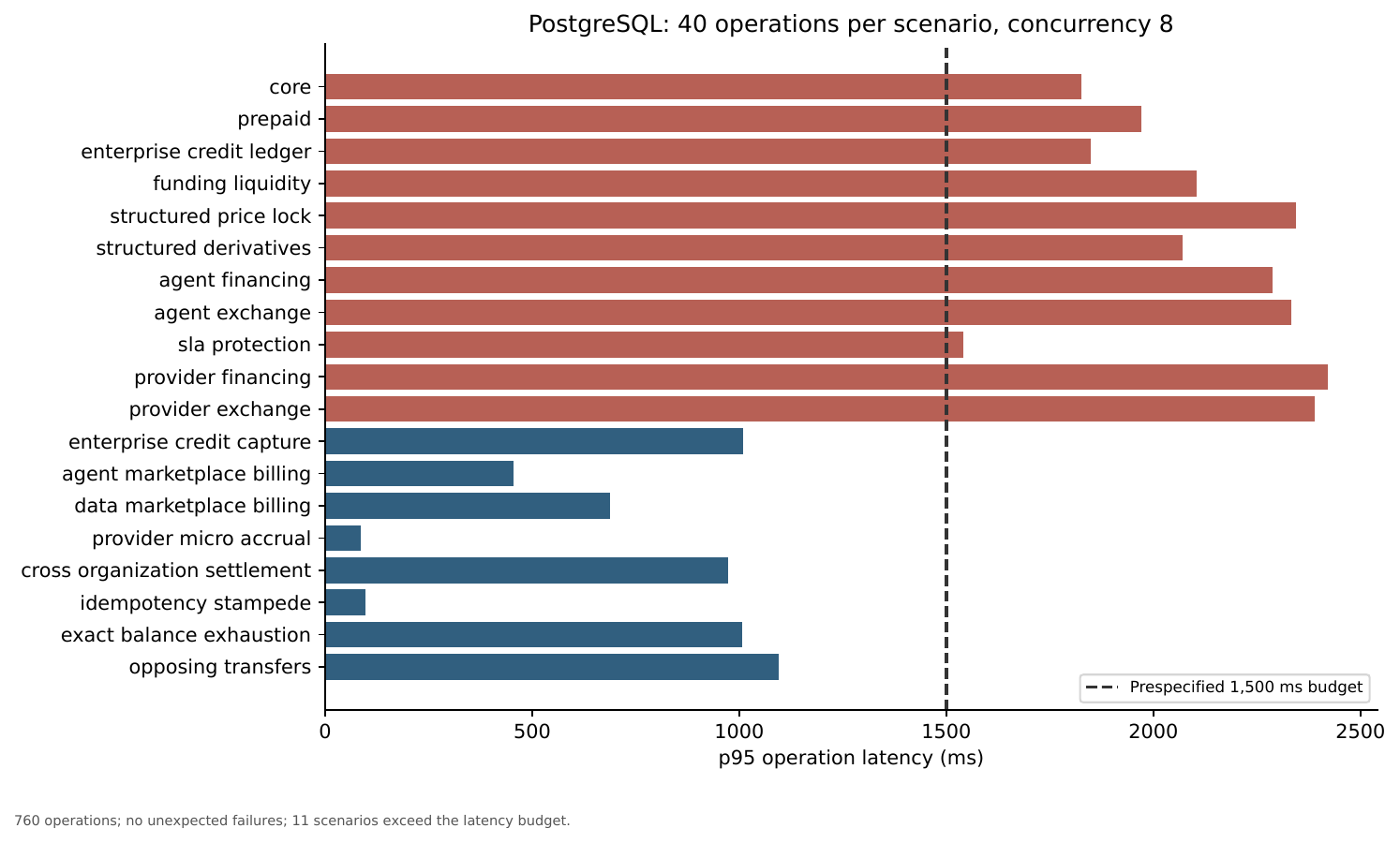}
\caption{Concurrent transaction latency across the 19 tested scenarios.
The reference line marks the prespecified 1,500\,ms budget; red bars indicate
scenarios above that budget.}
\label{fig:current-concurrency}\end{figure}

\subsection{Continuous-Balance Financing Tests}
A separate synthetic workload tests financing without the daily budget
resets used in the request-level replay. It contains 24 scripted tasks,
each with three nominal calls costing USD~0.04, 0.06, and 0.05.
The costs and task outcomes are specified inputs rather than public-benchmark
measurements. The operator starts with USD~0.60, and financing supplies an
additional USD~0.60 from a separately tracked investor.

Only successful synthetic tasks generate receipts, set to 1, 1.5, 2, or
3 times the USD~0.15 reference task cost. A revenue right allocates 10\%
of gross receipts to the investor. An equal-funded loan control instead
requires repayment of principal plus 1\% at the end of the sequence.
These terms are experimental assumptions, not market quotations.
Each condition re-executes the task sequence under its available-cash
constraint. Capital injections are excluded from contribution profit;
investor receipts and outstanding obligations are recorded separately.
The trajectories validate continuous cash-flow accounting under the
specified funding and revenue-allocation terms.

\begin{figure}[tb]\centering
\includegraphics[width=\linewidth]{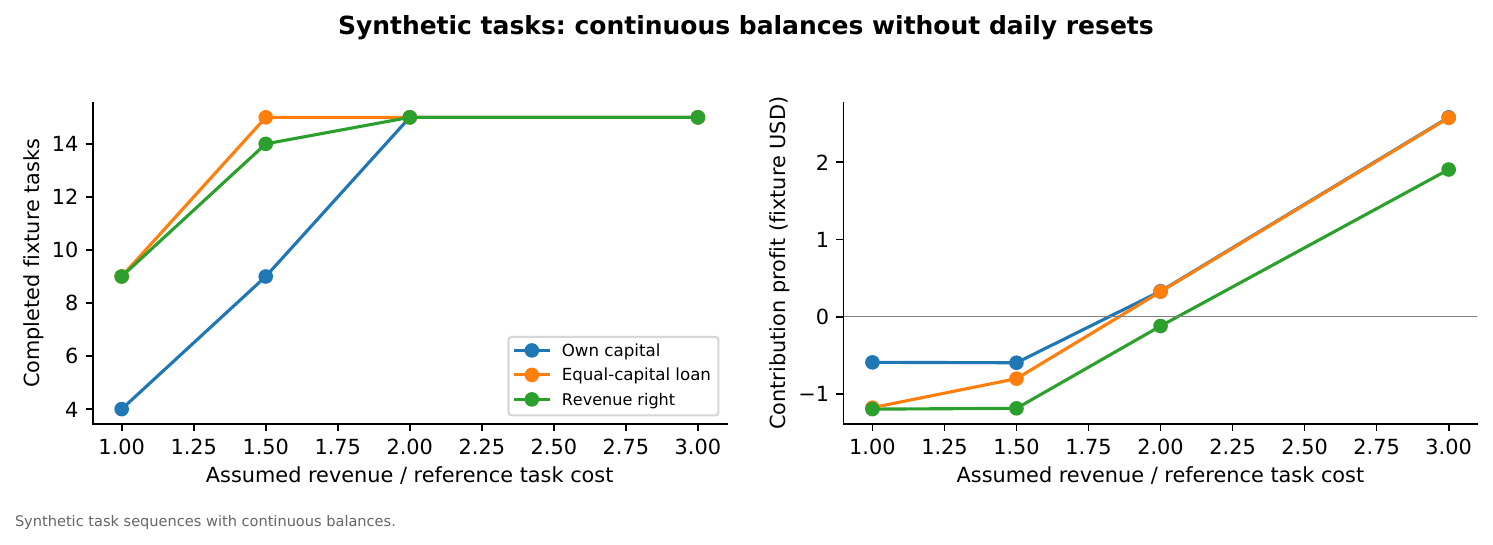}
\caption{Financing outcomes for synthetic task sequences with continuous
balances. Scripted task counts and contribution quantify the accounting
effects of the specified funding and receipt assumptions.}
\label{fig:finance-fixture}\end{figure}

\subsection{Per-Claim Protection Tests}
The protection tests exercise per-claim payment and reserve accounting.
The parameters are a USD~0.10 cap per claim, a USD~0.20 initial reserve,
and a USD~0.01 premium per covered synthetic task. A claim is paid in full up to its cap or rejected when
available reserve funds are insufficient; partial payment is not assumed.

Scripted incidents include timeout, rate limiting, primary-path outage,
and common-path outage. Tool errors and incorrect answers are excluded
from coverage. Recovery-enabled conditions permit one alternate attempt.
A residual covered failure requests USD~0.15 before application of the
claim cap. These eligibility assignments are controlled inputs, not
coverage labels inferred from BurstGPT. The trajectories test payment
and reserve accounting; actual API claim validation is assessed separately
by the PostgreSQL experiments above.

\begin{figure}[tb]\centering
\includegraphics[width=\linewidth]{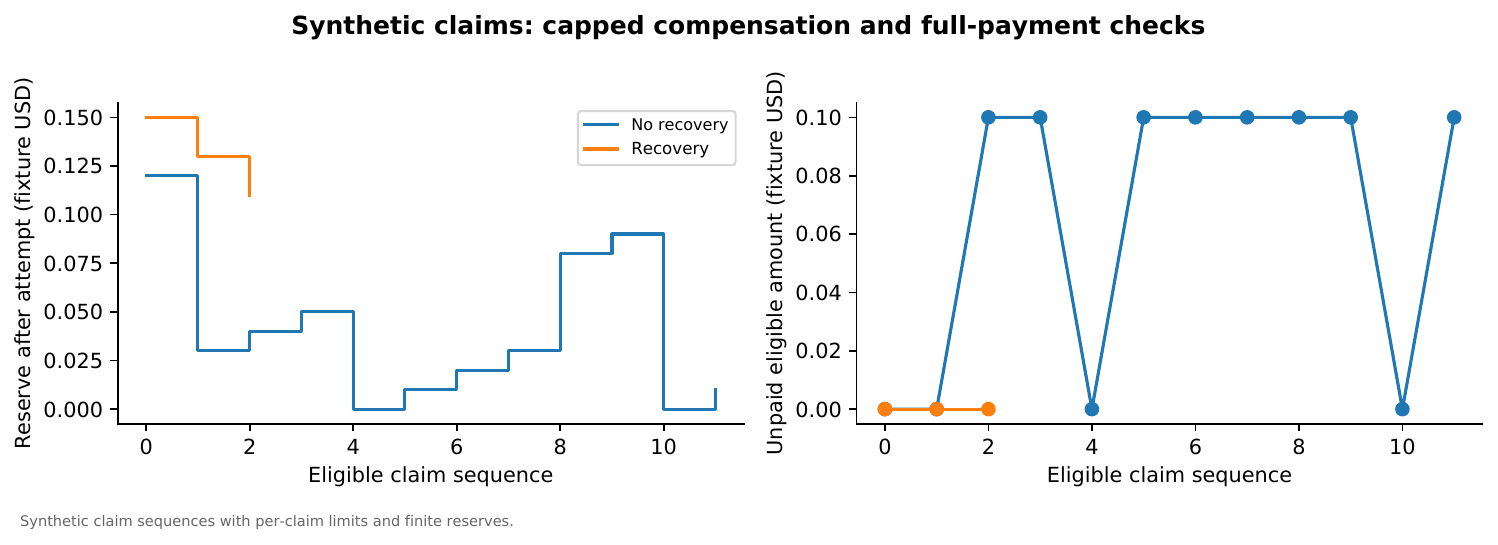}
\caption{Reserve balances and payment outcomes in deterministic protection tests.
Each submitted claim is checked against its cap and available funds.
The horizontal axis indexes eligible claims separately for each condition:
12 without recovery and 3 with recovery.}
\label{fig:protection-fixture}\end{figure}

\subsection{Factorial Accounting Checks}
The synthetic workload also crosses the eight combinations of forwards
($D$), financing ($F$), and protection ($S$) with recovery enabled or disabled
and four receipt multipliers, yielding 64 conditions. Eight equal-funded
loan controls bring the total to 72. Event-level cash flows are recorded
for every condition, and conservation is checked across the operator,
investor, reserve, service-receipt accounts, and assumed customer payer.
These checks validate cash-flow conservation across combined financial
operations, using a synthetic population separate from the model-driven
task studies.
\FloatBarrier

\section{Controlled Agent Evaluation: Experimental Design}
\label{app:controlled-protocol}

The controlled study isolates the effects of financing, recovery, and
protection within a tractable service workflow. It uses real model
inference and executable business tools, while fixing customer requests,
local business records, and economic assumptions. The task split,
completion criteria, parameter-selection rules, and treatment structure
were specified before evaluation. The tool interface was finalized during
calibration, as described below. Completed outcomes are reported in
Appendix~\ref{app:eval-supplement}, with controlled and retail outcomes
reported under their respective study denominators.

\subsection{Workload and Completion Criteria}
Each task combines a fixed customer request, constructed local business
records, Qwen/Qwen3-8B inference through Nexus and SiliconFlow, and callable
business tools. Neither a model-driven customer nor a model-based judge
is used. A task generates a receipt only when the requested operation
and its final output both satisfy the completion rule.

\begin{table}[htbp]\centering\small
\caption{Controlled task families and completion criteria. Each family
contains three calibration tasks and four evaluation tasks.}
\label{tab:controlled-workload}
\begin{tabular}{@{}p{.16\linewidth}p{.34\linewidth}p{.42\linewidth}@{}}
\toprule Family & Model-directed operation & Completion criterion\\\midrule
Extraction & Extract a reference, quantity, and city; submit the record &
Submitted fields and final JSON match the request.\\
Order lookup & Invoke the order-lookup tool &
The requested order is retrieved and final JSON matches its state.\\
Product selection & Query the catalog; select and quote an item &
The quoted item is the cheapest eligible in-stock item; quantity, total,
and final JSON are correct.\\\bottomrule
\end{tabular}\end{table}

The workload contains 21 tasks: nine for calibration and 12 for evaluation,
with distinct identifiers and business records. Initial ordering is
determined by SHA-256 hashes of task identifiers. The shared-template
design targets these task families. Every scheduled case remains in its
study denominator, including errors and funding-admission stops.

The model configuration uses temperature zero, disabled thinking, a
512-token output limit, and per-task limits of six logical requests,
18 physical requests, and 600 seconds. A logical request is an intended
model call; its retries count as additional physical requests. The first
response must invoke a tool; subsequent responses may invoke tools or
finish the task. Separate executions retain the variability of remote
inference at this configuration.

\subsection{Calibration and Workflow Configuration}
Before economic parameters are fixed, all nine calibration tasks must reach a terminal
outcome, at least two of the three tasks in each family must succeed,
and returned usage must reconcile with Nexus records. This criterion
establishes workload readiness before the economic parameters are fixed.

Two calibration passes use the same nine task identifiers, business records,
split, and initial ordering. The final interface enumerates valid category
codes and exposes dependent operations in the order catalog lookup, quote
creation, and final output. This configuration is selected using calibration
tasks; the model selects the product and supplies its own responses.
Evaluation tasks are first executed after calibration is complete.
The resulting study evaluates a calibration-informed, constrained service
workflow. Appendix~\ref{app:eval-protocols} reports both calibration passes
and their measurements.

\subsection{Measurement and Operating-Capital Model}
Returned input and output token counts and request latency are measured.
Under the economic-valuation conventions in
Section~\ref{sec:eval-measurement-scope}, each request is assigned the
following laboratory expense at the supported precision:
\begin{equation}
 c_i=\operatorname{round}_{10^{-6}}
 \left(\frac{100 n_i^{\mathrm{in}}+500 n_i^{\mathrm{out}}}{10^6}\right).
\end{equation}
Here $n_i^{\mathrm{in}}$ and $n_i^{\mathrm{out}}$ are returned token counts.
Amounts are in experimental USD. The prespecified scale makes small-task
losses representable at cent precision; fees, rounding, and coverage are
interpreted at this fixed scale. Requests with unknown usage retain their
reservations under the stated cost-bound convention.

Let $\mu$ be the mean known calibration expense per scheduled task and
$H_{\mathrm{cal}}$ the largest conservative single-request reservation
observed during calibration. Reservations are derived from the serialized
UTF-8 payload, a framing allowance, and the maximum output allowance.
The operating parameters are
\begin{equation}
 B=\mu+H_{\mathrm{cal}},\qquad
 A=\lceil4\mu\rceil_{0.01},\qquad r=2\mu,
\end{equation}
where $B$ is initial operator capital, $A$ is additional funding, $r$ is
the receipt per successful task, and $\lceil\cdot\rceil_{0.01}$ rounds up
to cents. These symbols follow Table~\ref{tab:eval-parameters}.
Receipts are paid after every four scheduled tasks; denied tasks still
advance this schedule. Balances evolve continuously and never reset
between tasks.

The revenue-right condition allocates receipts as 89\% to the operator,
10\% to the investor, and 1\% to the platform through the implemented
accounting procedures. Funding principal is excluded from contribution;
interim investor transfers and outstanding rights are tracked separately. Changing receipts on an already observed trajectory
provides a conditional accounting sensitivity. An operational comparison
requires fresh execution whenever the change can affect task admission.

\subsection{Treatments and Outcome Measures}
\paragraph{Financing.}
Own-capital, equal-capital, and revenue-right portfolios each execute the
12 evaluation tasks. The equal-capital control separates the benefit of
additional available funds from the cost of sharing revenue. Outcomes
include completed services, admission stops, continuous balances, operator
contribution, and investor cash flow. Known operator contribution is
retained customer receipts minus known inference expense; injected capital
is excluded, and unknown expense and outstanding receivables are reported
separately. Productive growth requires both additional completed services
and positive incremental contribution, rather than contract approval alone.

\paragraph{Recovery and protection.}
Four additional portfolios cross own-capital or revenue-right financing
with recovery disabled or enabled. Their ample operating capital,
$100\mu+H_{\mathrm{cal}}$, separates recovery from a binding funding
constraint. A fixed schedule injects timeout, rate limiting, primary-path
outage, or common-path outage before the second logical Agent request.
Recovery uses an alternative local path while holding the model and supplier
fixed. The measured recovery scope is therefore local-path continuation. Incident reachability,
subsequent responses, and completed services are distinct outcomes.

Protection is applied to the same execution trajectory without changing
task completion or operating capital. Only covered terminal injected
failures generate claims; incorrect answers do not. Eligible loss is known
incurred laboratory expense rounded down to cents. The claim cap is
$L=\max(0.01,\lceil\mu\rceil_{0.01})$, the premium is
$\max(0.01,\lceil0.01\mu\rceil_{0.01})$, and the funded and scarce reserves
are $12L$ and $L$. The contract API determines approval, rejection, and
settlement, including funding checks. Payments are service credits, not
cash or replenishments of operating capital. Requested, approved, paid,
and unpaid amounts and the remaining reserve are reported separately.

\paragraph{Forwards.}
Contracts are approved before evaluation, with strike one, zero discount,
and calibration-fixed notional $N_h=12\mu h$ for
$h\in\{0.25,0.50,0.75,1\}$. The no-contract control uses $h=0$.
Terminal indices $I\in\{0.5,1,1.5\}$ are assumed price scenarios rather
than market observations. The contract fee rate is 0.012. Fees, both
counterparties' settlements, and unpaid amounts are reconciled to API
records. Conditional expenditure applies scenario prices to realized
demand, subtracts signed forward payments, and adds fees. Realized demand
may differ from the calibration notional; applying prices after execution
does not change the observed admission decisions.

\paragraph{Combined comparisons and pairing.}
The seven execution arms produce 84 scheduled task--arm cases. A fixed
pseudorandom schedule varies arm order within each task. Combinations of
forwards ($D$), financing ($F$), and protection ($S$) reuse a trajectory
only when the financial operation cannot alter available operating funds
or recovery. Financing and recovery changes therefore require distinct
executions. Protection-adjusted service value is distinguished from cash
contribution. One chronological portfolio per arm defines the unit
for descriptive comparison.

\subsection{Verification and Scope of Inference}
Request identifiers link outcomes and token usage to contract operations
and ledger changes. Contract operations use authenticated business APIs
on a dedicated PostgreSQL database. Independent decimal accounting checks
amounts and balances exactly; effectiveness results require successful
reconciliation. Restart handling preserves completed operations and
settled claims, while incomplete interruptions remain visible. Seven
offline checks assess the workload implementation separately from the
model-driven effectiveness measurements.

\begin{table}[!ht]\centering\small
\caption{Measurement scope of the controlled evaluation.}
\begin{tabular}{@{}p{.30\linewidth}p{.61\linewidth}@{}}
\toprule Component & Measurement and interpretation\\\midrule
Model and tool execution & Responses, returned usage, and completed
operations are observed on the controlled task set.\\
Business records and faults & Constructed records and local fault schedules
define the controlled execution conditions.\\
Prices and receipts & Experimental tariffs, success-triggered receipts,
and terminal price indices define the scenario valuation.\\
Contract settlement & Account changes and service-credit payments are
observed in isolated database experiments and reconciled independently.\\
Study population & Fixed task families define the repeated-study population;
calibration, controlled evaluation, and retail cohorts retain separate denominators.\\\bottomrule
\end{tabular}\end{table}
\FloatBarrier

\section{Supplementary Evaluation Methods and Results}
\label{app:eval-supplement}

This appendix provides the workload definitions, cohort-specific results,
and sensitivity analyses supporting the main evaluation. The organization
follows the three financial mechanisms and their combined accounting.
Request replay characterizes expenditure under specified scenarios,
model-driven execution measures service and capital outcomes, and synthetic
contract tests validate accounting rules. Each study retains its own
population and measurement scope, as defined in
Section~\ref{sec:eval-measurement-scope}.

\subsection{Workloads, Calibration, and Measurement}
\label{app:eval-protocols}
Table~\ref{tab:eval-cohorts} summarizes the evaluation cohorts. The protocols
below specify their selection, execution, and accounting rules. Calibration
and development attempts are reported separately from evaluation outcomes.

\subsubsection{Request-Level Replay}
\paragraph{Data and pricing.}
We use the first two months of BurstGPT~\cite{burstgpt2025}, with the dataset
version recorded in the supplementary artifact. Removing conversation
records and the incomplete final day leaves 1,261,858 API requests.
Days 0--17 provide 362,417 calibration requests, and days 18--59 provide
899,441 evaluation requests. Replay preserves timestamps, model labels,
token counts, and original request order. All requests remain in the
offered workload, although budget-constrained conditions may admit only
a subset. No arrivals or duplicate requests are added.

The reference tariff is \$3 per million input tokens and \$15 per million
output tokens, yielding \$3,395.80116 for successful first-attempt requests.
This amount is a reference-tariff valuation. Of 16,137 zero-output
records, 16,131 also lack input lengths. Zero output supplies the failure
indicator; unobserved first-attempt charges use the same assumptions in
paired conditions, and alternate-attempt lengths are estimated from
calibration data. The trace supplies request timing and usage; task
correctness and contractual coverage are assessed in the separate
model-driven and contract studies.

\paragraph{Matched conditions.}
The principal combined comparison fixes the offered requests, provides
sufficient funding access, and models one alternate attempt after a
primary failure. Only financial contracts vary. Separate experiments
examine budget-limited service growth and changes in technical recovery.
Forward comparisons hold requests, prices, and cache/batch settings equal;
financing-cost comparisons hold the funded amount equal; protection
comparisons hold the recovery policy equal. This matching isolates
financial-contract effects from model selection, caching, and failover.

Prices are scenario assumptions expressed relative to the reference
tariff. The paths include a constant value of 1, linear changes from 1
to 1.3 or 0.7 over 42 days, a seven-day increase to 1.8, and an increase
triggered by the preceding five-minute arrival count. Calibration defines
bursts as 1,299 requests per five-minute window and long inputs by a
2,527-token threshold. Each paired baseline faces the same path. Financing
and shared-failure stresses are examined separately; combined stresses
describe joint outcomes rather than isolated mechanism effects.

\subsubsection{Controlled Agent Cohorts}
\paragraph{Calibration and initial study.}
The controlled workload contains 12 evaluation tasks: four each for
structured extraction, order lookup, and constrained product selection.
It uses Qwen/Qwen3-8B through Nexus and SiliconFlow, constructed business
records, and controlled local faults, without a customer-simulation or
judge model. Appendix~\ref{app:controlled-protocol} defines its completion
criteria and treatment structure.

The first calibration pass succeeds on 3/3 extraction tasks, 3/3 lookup tasks,
and 1/3 selection tasks; all 23 responses match Nexus records. The second pass
uses the explicit category codes and dependency-ordered interface described
in Appendix~\ref{app:controlled-protocol}. All nine tasks succeed, and all
21 responses reconcile exactly. Task identifiers, records, split, and
initial ordering are unchanged. The final workflow is selected using these
calibration observations before any evaluation-task model calls.
Both passes retain their measurements and separate calibration status.

The initial study completes all 84 scheduled task--arm cases, with
51 successful services and 150 responses from 150 physical requests.
Usage is available for every returned response, totaling 49,086 input
and 5,323 output tokens. Median client request latency is 4.53 seconds,
including network effects. Monetary valuation follows the prespecified
laboratory tariff of 100/500 experimental USD per million input/output
tokens and the assumed business receipts defined in
Section~\ref{sec:eval-measurement-scope}. Calibration fixes $\mu=0.127689$,
$H_{\mathrm{cal}}=0.943300$, $B=1.070989$, $A=0.52$, and $r=0.255378$.
Receipts are delayed until each block of four scheduled tasks, and balances
never reset between tasks. Capital effects therefore depend on the
request-reservation policy as well as the laboratory tariff. Each arm
provides one chronological portfolio for descriptive comparison.

\paragraph{Repeated portfolios and parameter sensitivity.}
A protocol fixed before inference specifies five separately executed
portfolios with randomized orders of the same 12 controlled tasks. Each
compares own capital, equal capital, and revenue rights, together with
recovery enabled or disabled under ample own capital. Seven additional
portfolios vary one financing factor at a time, and three vary fault
frequency. Accounts are isolated, model responses are newly obtained,
and completion rules and calibration remain fixed. The repeated-task
design measures financial outcomes under controlled operating conditions.

All 624 scheduled cases reach a terminal outcome. The 1,247 physical
requests yield 1,246 responses and 487 successful services. There is one
request with unknown returned usage, 39 funding-admission stops, no
infrastructure interruptions, and 98 terminal injected-fault failures.
Available response usage matches Nexus records, totaling 436,701 input
and 43,028 output tokens. Unknown usage is handled by the reservation
bounds below. Admission uses input-byte and maximum-output reservations,
with unused amounts released after measured usage is known. The median
reservation-to-measured-charge ratio of 12.0153 characterizes the
advance-funding requirement in this study.

\paragraph{Missing-usage accounting.}
A request without returned usage retains its reservation, making those
funds unavailable to later tasks. Let $H_u$ be the unresolved reservation,
$C_{\rm known}$ known expense, and $Y_{\rm ret}$ retained portfolio
receipts. Under the reservation bound, feasible contribution lies in
\begin{equation}
[Y_{\rm ret}-C_{\rm known}-H_u,\;
 Y_{\rm ret}-C_{\rm known}].
\end{equation}
Treatment-minus-control bounds subtract the upper control endpoint from
the lower treatment endpoint, and conversely. Known-expense contribution
and the interval covering unresolved usage are reported separately.

For the earlier repeated-portfolio cohort, these bounds are a retrospective
analysis of the unresolved request in the second portfolio. The execution
protocol, task list, and recorded outcomes are unchanged, and all affected
portfolios remain in that cohort. The same accounting rule is prespecified
for the subsequent capital replication.

\paragraph{Matched capital replication.}
A separate five-block protocol was fixed after inspection of the earlier
single-portfolio boundary results but before collection of its own
observations. Each block freshly executes own-capital, equal-capital,
and revenue-right arms at initial capital $0.5B$, $B$, and $1.5B$, giving
$5\times3\times3\times12=540$ scheduled cases. Each block shares task order
and randomized within-task arm order across capital conditions; capital-
condition order is randomized separately. Calibration, model, tariff,
receipts, delay, and reservation policy remain unchanged. The fixed
randomization seeds are recorded in the supplementary artifact.

The replication records 1,082 physical requests, 1,081 responses,
456 successful services, 84 admission stops, one request with unknown
usage, and no infrastructure interruptions. Every scheduled case has
an outcome. Contribution excludes principal and bounds unresolved usage
rather than assigning zero cost. Exploratory 95\% percentile envelopes
jointly resample the five matched blocks across conditions using 10,000
draws. Repeated economic portfolios are the sampling units. The earlier
incident-affected study retains its separate observations and denominator.

\subsubsection{Retail Task Studies}
\paragraph{Environment and task selection.}
The retail studies use a fixed version of the official text environment
of $\tau^2$-Bench~\cite{barres2025tau2}; its revision is recorded in the
supplementary artifact. SHA-256 ordering partitions 114 tasks into
22 calibration and 92 evaluation tasks. The baseline takes the first
five evaluation tasks, each twice, with selection fixed independently of
pilot rewards. These identifiers also appear in preliminary integration
trials. The baseline is a repeated evaluation of this custom subset, with its
prior integration exposure recorded separately from task selection.

The official instructions, tools, and ALL completion criterion are used.
All model roles employ Qwen/Qwen3-8B through Nexus and SiliconFlow.
The customer is model-driven, with no scripted responses substituted.
Applicable model-based evaluator checks use the same model, rather than
the upstream default judge.

\paragraph{Execution limits.}
Execution is serial, with temperature zero, thinking disabled, a
2,048-token output cap per request, 30 requests per role per task,
a 600-second between-call deadline, 60 orchestrator steps, and a limit
of 10 consecutive tool errors. Transport timeouts are bounded by the
remaining task time. Transient transport, routing, rate-limit, and server
failures permit at most three physical requests per logical call, with
two- and four-second backoff. Each physical request consumes the request
limit and has a distinct identifier; uncertain prior usage remains unknown.
Tools execute only after a complete response and are not replayed.
Authentication is renewed before requests at 20-minute intervals.

\paragraph{Baseline outcomes and reconciliation.}
The baseline records ten attempted trials: nine receive an official score,
four succeed, and one is unscored. Across all attempts, 368 requests yield
360 responses; eight have no usage in a completed client response.
Seven logical calls require retries, all of which eventually return a
complete model response, and three authentication renewals are recorded.
These are naturally observed incidents rather than randomized treatment
effects; usage of preceding failed attempts remains unknown.

For operational success across all attempted trials, 10,000 bootstrap
draws over task means give a descriptive 95\% percentile interval of
[10.0\%, 70.0\%]. Repetitions remain within task clusters. Operational
success counts completed successful tasks; unscored trials retain missing
official rewards, and tool outcomes remain in the recorded trajectories.

All 360 returned responses link to successful gateway records, with exact
agreement in token counts, selected source, and configured wallet charges.
Conditional financial analyses apply the assumed 1/5 input/output tariff
per million tokens to returned Agent usage under the measurement conventions
in Section~\ref{sec:eval-measurement-scope}. Simulated-customer and evaluator
usage enter the experimental budget separately from operator expense;
unresolved usage retains its accounting bounds.

\begin{table}[tb]
\centering\small
\caption{Real fixed evaluation model usage by role. Simulated-user and evaluator costs
belong to the experimental budget rather than Agent operating expenditure.
Latency covers returned responses; task duration also includes failed attempts
and retry delays.}
\label{tab:live-usage}
\begin{tabular}{lrrrr}\toprule
Role & Responses & Input tokens & Output tokens & Mean latency (s)\\\midrule
Agent & 237 & 1,658,417 & 13,796 & 9.87\\
Simulated user & 121 & 157,622 & 7,009 & 9.49\\
Evaluator & 2 & 9,604 & 180 & 4.19\\
\bottomrule\end{tabular}
\end{table}

\begin{figure}[tb]\centering
\includegraphics[width=\linewidth]{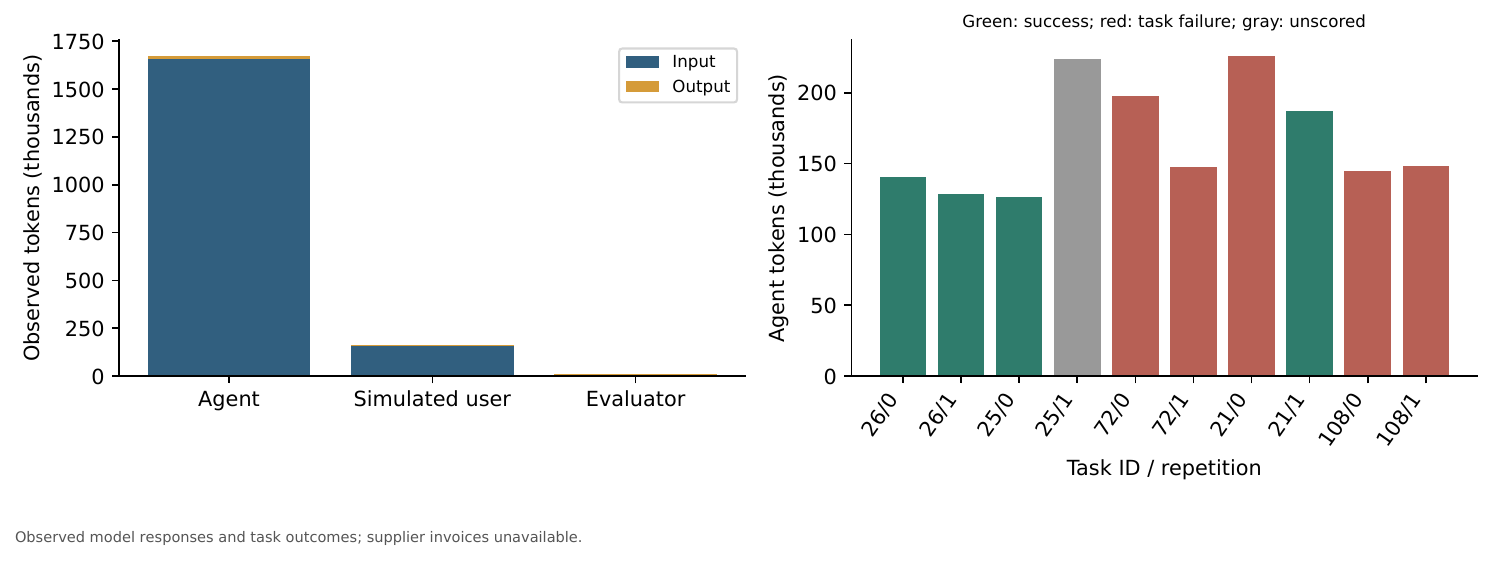}
\caption{Retail baseline token usage and task outcomes. Repeated attempts are
paired by task identifier; task success follows the official completion rule.}\label{fig:live-agent}\end{figure}

\begin{table}[tb]\centering\small
\caption{Outcomes of all prespecified task attempts under the official completion
rule. Tool-error counts and total evaluation time are reported separately.}
\label{tab:live-tasks}
\begin{tabular}{llrrr}\toprule
Task/repetition & Outcome & Tools (errors) & Agent tokens & Seconds\\\midrule
26/0 & Success & 10 (2) & 140,455 & 373.7\\
26/1 & Success & 10 (2) & 128,610 & 247.7\\
25/0 & Success & 10 (4) & 126,529 & 385.0\\
25/1 & Unscored & 17 (6) & 223,946 & 600.0\\
72/0 & Failure & 10 (9) & 197,961 & 515.1\\
72/1 & Failure & 14 (6) & 147,962 & 314.7\\
21/0 & Failure & 18 (9) & 225,862 & 518.2\\
21/1 & Success & 15 (5) & 187,367 & 436.2\\
108/0 & Failure & 12 (10) & 145,018 & 245.5\\
108/1 & Failure & 9 (3) & 148,503 & 328.1\\
\bottomrule\end{tabular}\end{table}

\paragraph{Retail contract interventions.}
The contract study schedules 70 cases: ten previously unexecuted retail
tasks, each in seven execution arms. Independent calibration uses five
tasks twice each. Within-task arm order follows a fixed pseudorandom
schedule. Every scheduled outcome is retained, and each arm constitutes
one chronological economic portfolio.

At the assumed 1/5 input/output tariff, known calibration Agent cost per
attempt is $\mu=0.184298$. The initial balance is $B=\mu$, additional
capital is $A=\lceil2\mu\rceil_{0.01}=0.370000$, and the success receipt
is $2\mu=0.368596$ experimental USD. Balances evolve continuously.
Before each physical Agent request, a conservative reservation covers
the UTF-8 prompt bound and output allowance. Returned usage determines
the charge and release of unused funds; unknown usage retains its hold.
Simulator and judge expenses are separate, and principal is not revenue.

Contract approvals, funding checks, and transfers use authenticated
business APIs on isolated PostgreSQL accounts with disclosed initial
funds. Actual inference passes through Nexus and SiliconFlow, and returned
usage is reconciled to gateway requests. A fixed task-specific schedule
injects local faults before the third logical Agent request. The study
measures contract and execution outcomes under this controlled local-fault
schedule and the declared economic assumptions.

All 70 cases terminate, with 54 reaching an Agent request. Of 23 officially
scored cases, one succeeds. The remaining 47 are unscored: 24 encounter
funding-admission stops, 20 terminate at injected incidents, and three
reach task-resource limits. Calibration separately yields nine scored
failures and one unscored attempt without changing task selection.
The study obtains 925 complete responses from 930 physical requests,
all reconciled to gateway records. Five requests have unknown usage,
with their reservations reported separately from known expense. These
outcomes use the intervention cohort's own task population and denominator.

\begin{figure}[tb]\centering
\includegraphics[width=\linewidth]{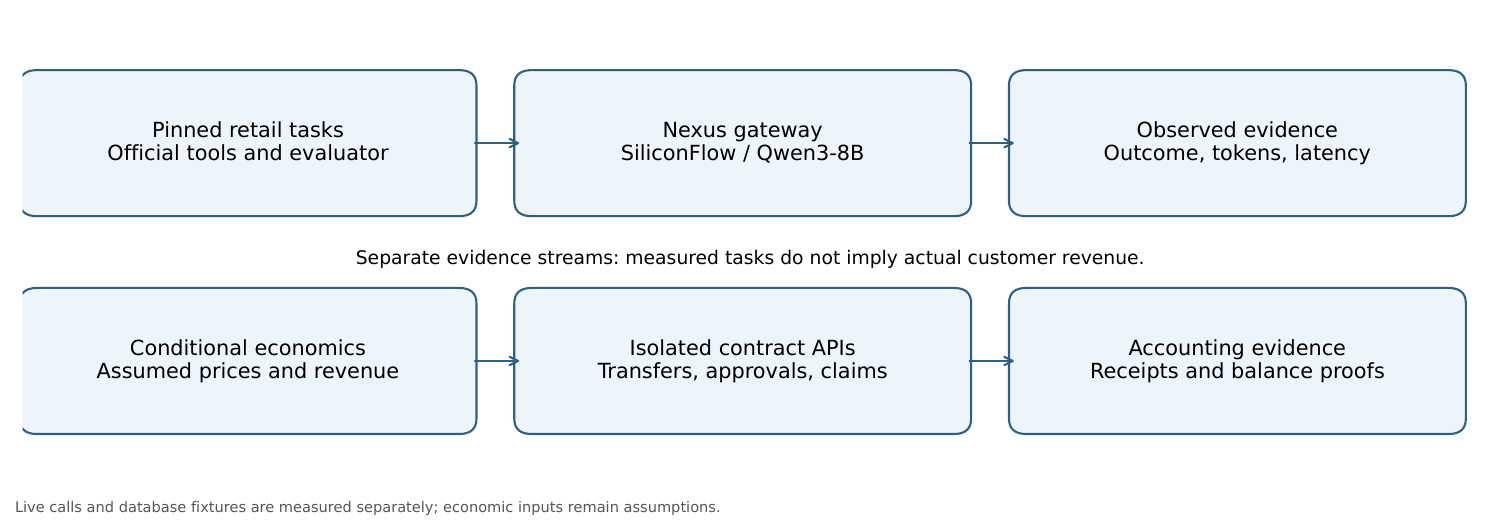}
\caption{Measurement organization connecting model-driven execution, database
contract experiments, and scenario-based economic valuation.}\label{fig:evidence-flow}
\end{figure}
\FloatBarrier

\subsection{Forward Settlement, Price Mismatch, and Risk Decomposition}
\label{app:eval-forwards}
The analyses support Section~\ref{sec:eval-forwards} by reconciling
settlements on measured Agent demand and quantifying expenditure responses
to price paths, funding availability, and contract-to-bill alignment.

\subsubsection{Settlement on Measured Agent Demand}
\paragraph{Controlled workload.}
All 84 forward contracts in the initial controlled cohort reconcile in
signed payment, fee, and settlement count. Contracts are approved before
evaluation with zero discount and calibration-fixed notionals. Declared
terminal-price scenarios are applied to measured demand, with the same
trace retained in the no-contract control. This pairs actual settlement
execution with conditional expenditure analysis.

\begin{figure}[htbp]\centering
\includegraphics[width=.85\linewidth]{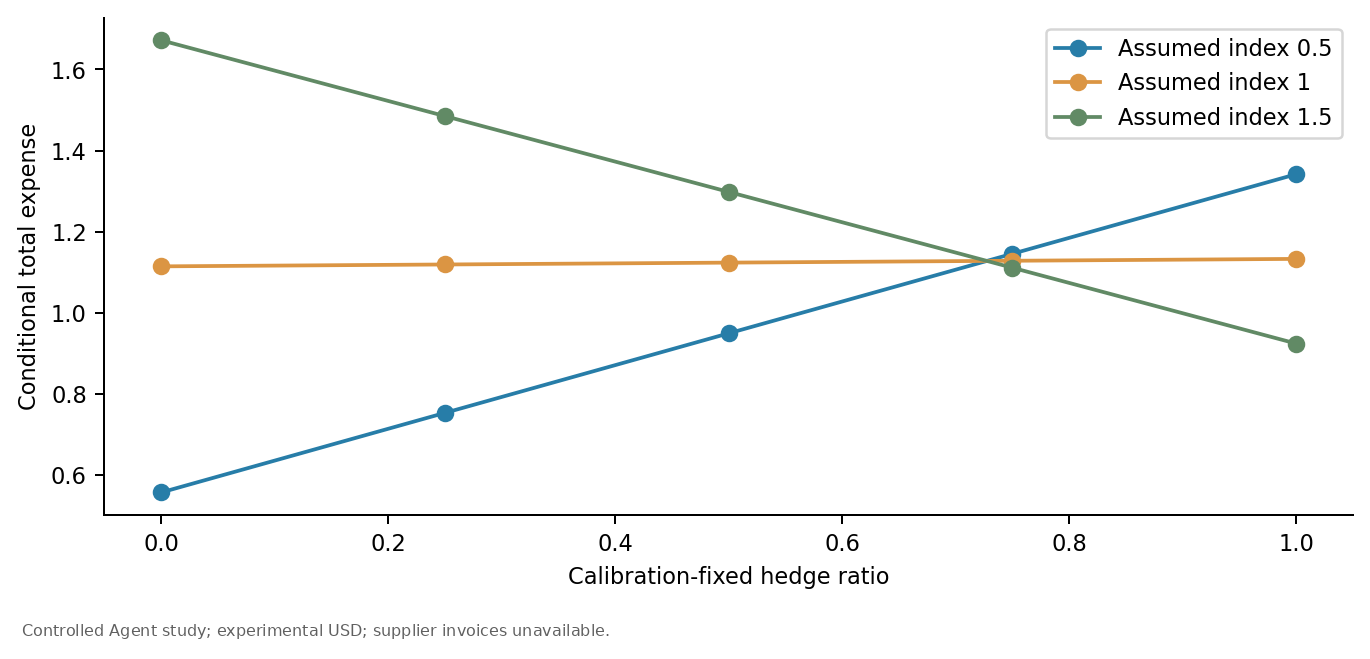}
\caption{Conditional expense versus hedge ratio on the own-capital recovery trajectory under the declared terminal-price scenarios.}\label{fig:controlled-forwards}
\end{figure}
\paragraph{Retail workload.}
Before any intervention task, we form zero-discount contracts at strike one
with notionals $h n\mu$, $h\in\{0.25,0.5,0.75,1\}$ and $n=10$.
The 84 arm--hedge--index contracts use declared final indices 0.5, 1 and 1.5;
these are scenarios, not observed provider price changes. Settlement occurs
after the workload within the ten-day validity window, leaving the observed
task-admission decisions unchanged. A separately recorded 1.2\%
experimental fee is transferred through the ledger. Both counterparties use
adequate disclosed test funds. Actual API settlements are compared exactly
with $hn\mu(I-1)$ and both account balances, one settlement per contract.
The zero-hedge and nonzero-hedge controls use measured returned demand,
allowing realized usage to vary relative to the calibration notional.
The associated block-resampled analysis characterizes the declared
price-path scenarios.
\begin{figure}[tb]\centering
\includegraphics[width=0.9\linewidth]{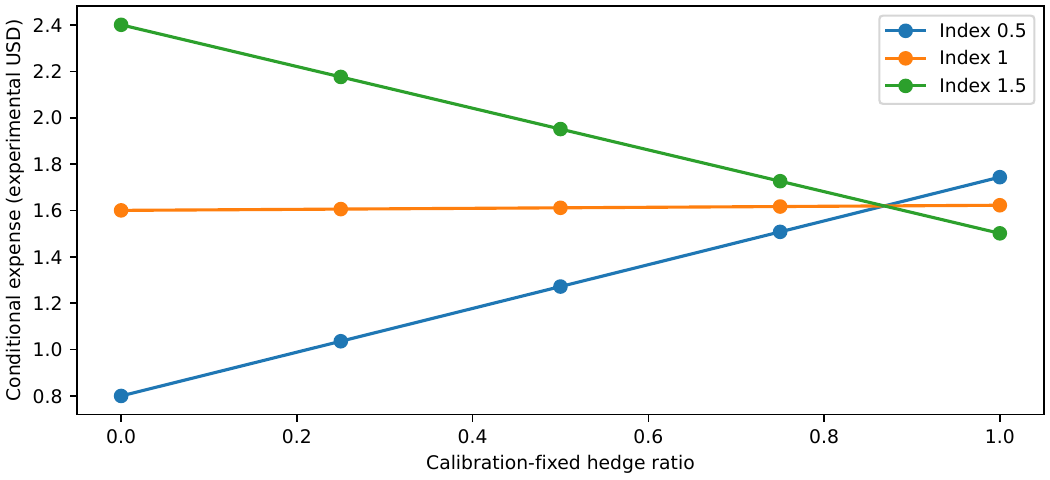}
\caption{Zero-discount forward comparison on the recovery-enabled own-capital
arm's measured returned demand. Calibration-fixed notionals and declared
price indices determine actual API settlements and fees, which enter the
conditional expenditure calculation.}
\label{fig:live-forwards}\end{figure}

\subsubsection{Expenditure Sensitivity and Exposure Mismatch}
\paragraph{Conditional expenditure on the retail baseline.}
For identical measured demand in each price scenario, let $C_s$ denote its
conditional cost, $P_s$ the signed paid settlement, and $f$ the fee. Then
$C_s^D-C_s=f-P_s$. The scenario analysis of the ten-trial retail baseline
uses zero strike discount, hedge ratios
0, 0.25, 0.5, 0.75 and 1, and a declared nominal exposure of 0.25 per scheduled
attempt, varied by factors 0.5, 1 and 1.5. This is an exposure sensitivity,
not a forecast fitted to evaluation outcomes. We compare matched price/index
movement, output-only price movement, and an unchanged index. Under full
settlement, a unit-strike forward with a 1.2\% nominal fee reduces expenditure
above an index of 1.012. This threshold identifies the price movement
needed to offset fees. Buyer and seller funding stresses report paid and
unpaid obligations separately; the BurstGPT block analysis supplies the
corresponding conditional path-risk measures.

\begin{figure}[tb]\centering
\includegraphics[width=\linewidth]{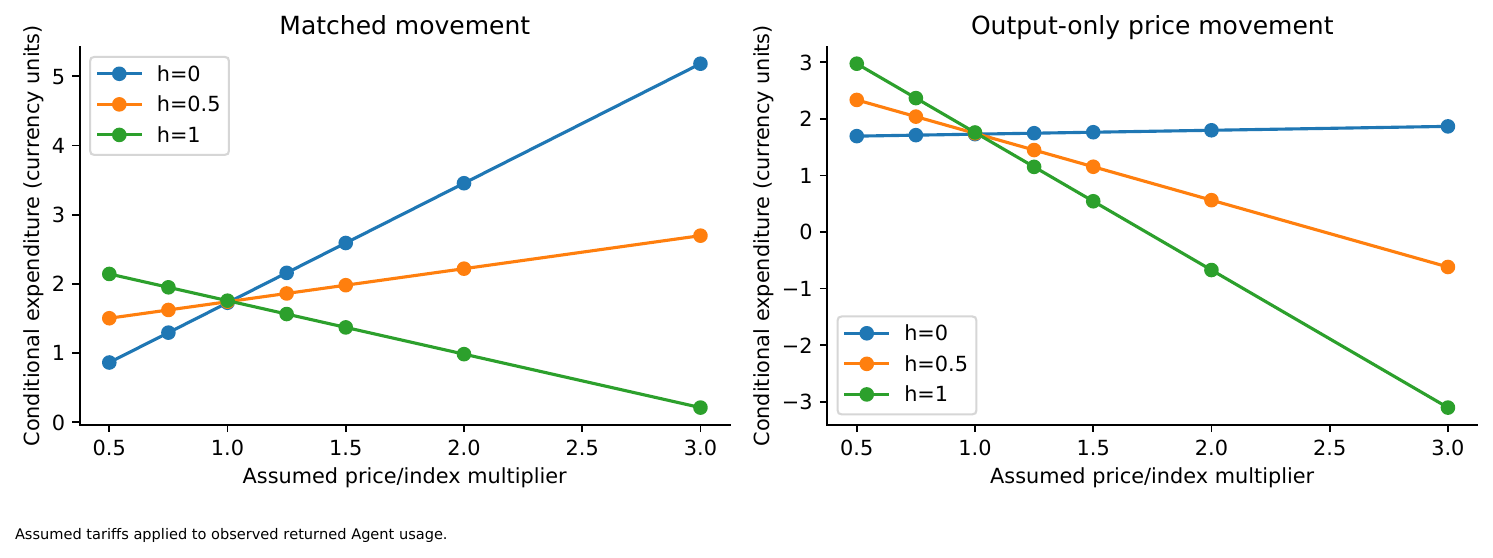}
\caption{Paired forward scenarios applied to measured returned Agent usage.
Declared price and notional scenarios compare matched and output-only
movements to characterize settlement alignment with operating expense.}
\label{fig:agent-forward-effectiveness}\end{figure}

\paragraph{Payments under matched execution.}
The reference forward covers 80\% of the calibration-mean daily expenditure,
with fees and holding-period funding charges of 1\% and 0.2\% of the
contracted amount, respectively. Contract prices assume zero or 10\% discount.
Under constant prices and the assumed 10\% discount, modeled expenditure
falls by \$155.39 (4.58\%). The contract covers about 52\% of realized base
expenditure, explaining why the reduction is smaller than the assumed
discount. The buyer receives \$176.58 from the seller and pays \$21.19
in fees and funding charges. With no discount, expenditure instead
increases by 0.62\%.

The financial effect complements execution optimization. Under identical
assumed cache and batch eligibility, ordinary optimization
reduces the unhedged invoice to \$2,418.20. Adding the discounted forward
reduces expenditure by a further \$113.04 (4.67\%). The ordinary optimization
uses 25\% batch-cost eligibility, 40\% input-cache hits, and 5\% cache writes,
with a 50\% batch discount and input-price multipliers of 0.1 for cache
hits and 1.25 for writes, applied identically in both arms. Its separate
\$977.60 saving is attributed to caching and batching. An ordinary
forward with identical terms gives the same payment, serving as an
arithmetic consistency control for the financial calculation.

\paragraph{Price and demand mismatch.}
With otherwise unchanged contract settings, gradually rising prices reduce
buyer expenditure by 11.39\%, while gradually falling prices increase it by
3.53\%. These paths differ from the constant-price sensitivity: keeping the
price 30\% below the reference throughout evaluation increases discounted
forward expenditure by 15.75\%. The cases must not be interpreted as the
same price treatment.

Actual purchases use each request's five-minute price, whereas settlement
uses the daily average of the reference prices. This difference, together
with forecast usage errors, prevents a contract from exactly offsetting
the bill. For example, the burst/long-input class has \$4.02 of
reference expense against approximately \$83.53 of committed notional.
Its stable-price net benefit of \$7.35 reflects contractual exposure
exceeding realized demand. These class allocations are analytical
portfolio breakdowns, with settlement scaled by the committed notional.

\begin{figure}[htbp]
\centering
\includegraphics[width=\linewidth]{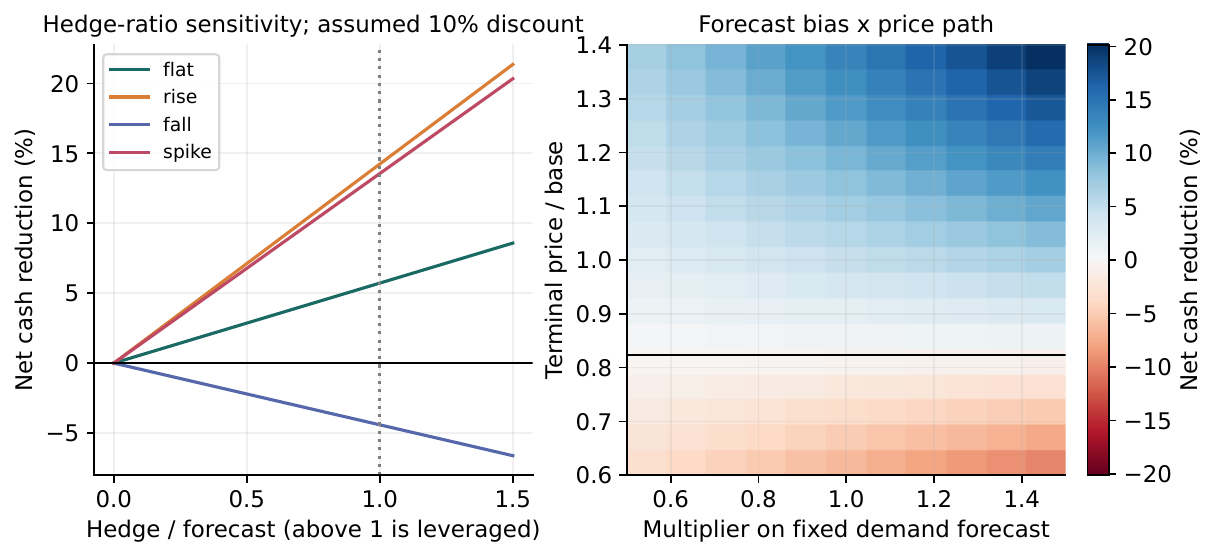}
\caption{Forward expenditure sensitivity to the covered amount, forecast
error, and assumed price path. Exposure scales the signed settlement
while the executed workload remains fixed.}
\label{fig:hedge}
\end{figure}

The forecast comparison uses only preceding days. Predictions use the fixed
calibration mean, the mean of all preceding days, the mean of the preceding
seven days, or the value seven days earlier. Their weighted absolute
percentage errors are 78.70\%, 103.77\%, 97.75\%, and 142.23\%, respectively;
this metric divides total absolute prediction error by total actual expense.
Figure~\ref{fig:hedge} varies the covered amount and forecast bias explicitly.
These measurements quantify the forecast and exposure ranges used in
the contract-to-bill sensitivity analysis.

\subsubsection{Variance Decomposition}
For $h=1$, matched reference, unscaled forecast, and funded settlement, the
1,000 paired paths yield Table~\ref{tab:hedge-variance-mechanism}. Here $Z$ is
signed settlement net of charges and $C_0$ is unhedged expenditure, as in
Equation~\eqref{eq:eval-forward-variance}. Sample moments use denominator
$n-1$. Variance and covariance columns use USD$^2$; the final column is the
paired change in mean expenditure in USD. The post-hoc decomposition
relates net-payment variation and covariance to the measured expenditure
response.

\begin{table}[tb]\centering\small
\caption{Mechanism of risk changes in conditional resampled paths. Negative
variance change is favorable; mean expenditure and variance are distinct outcomes.}
\label{tab:hedge-variance-mechanism}
\begin{tabular}{lrrrr}\toprule
Scenario & $\operatorname{Var}(Z)$ & $2\operatorname{Cov}(C_0,Z)$ & $\Delta\operatorname{Var}$ & $\Delta\mathbb{E}[C]$\\\midrule
flat & 0.0 & 0.0 & +0.0 & +26.49\\
rise & 4522.5 & -85383.2 & +89905.7 & -304.88\\
fall & 4522.5 & 93585.4 & -89063.0 & +357.85\\
spike & 47158.8 & -80862.3 & +128021.1 & -304.10\\
lagged-load & 0.2 & 413.1 & -412.8 & +25.08\\\bottomrule\end{tabular}\end{table}
\FloatBarrier

\subsection{Capital Constraints and Financing Sensitivity}
\label{app:eval-financing}
The matched capital replication is the principal financing comparison.
Supporting studies examine repeated controlled portfolios, individual
parameter changes, retail tasks, and request-level replay. Together, they
characterize financing effects across operating conditions, with the
largest matched-capital benefit observed under capital scarcity. Each
cohort retains its observations and denominator.

\subsubsection{Matched Capital Replication}
\paragraph{Admission and matched-capital outcomes.}
At $0.5B$, revenue-right financing reduces mean admission stops from
12.0 to 1.8 and raises mean contribution by 1.0635 experimental USD,
with positive contrasts in all five portfolios. Own-capital and
revenue-right admission stops are 1.4/0.0 at $B$ and 0.0/0.0 at $1.5B$.
Table~\ref{tab:finrep-uncertainty} reports all capital contrasts, missing-cost
bounds, and exploratory whole-block resampling ranges. Capital levels
were selected from the earlier boundary experiment, and the replication
protocol was fixed before its own observations. Interpretation follows
the portfolio-level sampling convention in Section~\ref{sec:eval-setup}.

\begin{table}[tb]
\centering\small\setlength{\tabcolsep}{4pt}
\caption{Contribution contrasts in the matched capital replication.
The mean column reports missing-usage bounds where non-degenerate. The envelope is the exploratory whole-block 95\%
resampling range, not another bound on a single missing request. The last
column counts definitely positive/negative blocks out of five. Units are
experimental USD; principal and credit are excluded.}
\label{tab:finrep-uncertainty}
\begin{tabular}{@{}llrrr@{}}\toprule
Capital & Contrast & Mean / cost bound & Resampling envelope & $+/-$ blocks\\\midrule
$0.5B$ & Right--own & 1.0635 & [0.9506, 1.1786] & 5/0\\
$0.5B$ & Right--equal & -0.3049 & [-0.3324, -0.2806] & 0/5\\
$0.5B$ & Equal--own & 1.3684 & [1.2642, 1.4894] & 5/0\\
$B$ & Right--own & -0.1406 & [-0.2657, -0.0155] & 2/3\\
$B$ & Right--equal & -0.3174 & [-0.3507, -0.2835] & 0/5\\
$B$ & Equal--own & 0.1769 & [0.0506, 0.3138] & 4/1\\
$1.5B$ & Right--own & -0.2962 & [-0.3294, -0.2632] & 0/5\\
$1.5B$ & Right--equal & [-0.2742, -0.0975] & [-0.2965, 0.2676] & 0/4\\
$1.5B$ & Equal--own & [-0.1987, -0.0220] & [-0.5741, 0.0000] & 0/2\\
\bottomrule\end{tabular}
\end{table}

\paragraph{Matched capital sensitivity and investor cash flows.}
The fifth matched block at $1.5B$ includes an equal-capital task that
succeeds after a retry following an upstream server error. The unresolved
usage of the earlier request retains a 0.883600 reservation. Its
contribution contrasts use the prespecified cost-bound accounting rule.

The low-minus-baseline difference in the financing-versus-own-capital contribution contrast has mean 1.2041 and exploratory matched-block envelope [0.9975, 1.4106]. The ample-minus-baseline difference in the financing-versus-own-capital contribution contrast has mean -0.1556 and exploratory matched-block envelope [-0.2587, -0.0524]. 

The investor contributes 0.5200 and receives mean wallet transfers of
0.2540 at low capital and 0.3000 at baseline and ample capital. Accrued
entitlements and rounding remainders are tracked separately. These are
observation-window cash flows with rights still outstanding, using the
scope defined in Section~\ref{sec:eval-measurement-scope}.

The separate repeated-portfolio study below retains its incident-affected
portfolio and unresolved-usage bound. The capital replication adds
matched evidence under the same tariff and reservation conventions.

\paragraph{Reservation-policy dependence.}
The 1,081 requests with returned usage have reservations ranging from
0.545800 to 0.943300 experimental USD, with a median reservation-to-known-
charge ratio of 12.4382. The low-capital opening balance is 0.535495.
These returned-usage records characterize the tested advance-funding
requirement relative to the opening balance. Their population excludes
rejected-request reservations; financing outcomes are interpreted under
this fixed reservation policy.

\subsubsection{Repeated Controlled Portfolios}
Whole continuous portfolios are the replication units because their
tasks share operating funds. All five paired differences are reported.
Exploratory 95\% envelopes use 10,000 paired-portfolio draws and the
lower 2.5th and upper 97.5th percentiles of the mean cost bounds. Without
missing usage, this reduces to percentile resampling of the observed
contrasts. Remote-model variation, provider incidents, and their effects
on later admission remain part of each realized portfolio outcome.

\begin{table}[htbp]\centering\small
\caption{Fresh portfolio contrasts. An asterisk identifies financing arms with unplanned request errors or missing usage. Brackets bound unresolved usage under the reservation limit. Recovery compares local recovery with none. Monetary values are experimental USD.}\label{tab:robust-repeats}
\begin{tabular}{@{}rrrrrr@{}}\toprule
Portfolio & Own tasks & Right tasks & $\Delta$ tasks & $\Delta$ profit & Recovery $\Delta$ tasks\\\midrule
1 & 11 & 12 & 1 & -0.1670 & 9\\
2* & 9 & 1 & -8 & [-2.1573, -1.2153] & 9\\
3 & 12 & 12 & 0 & -0.3065 & 9\\
4 & 10 & 12 & 2 & 0.0632 & 9\\
5 & 10 & 12 & 2 & -0.0799 & 9\\
\bottomrule\end{tabular}\end{table}

Revenue-right minus own-capital contribution has known-expense mean -0.3411, mean cost-bound range [-0.5295, -0.3411], and exploratory portfolio envelope [-1.3612, -0.0394].

Revenue-right minus equal-capital contribution has known-expense mean -0.5855, mean cost-bound range [-0.7739, -0.5855], and exploratory portfolio envelope [-1.7097, -0.2845].

Local recovery adds nine completed services in every one of the five portfolios, providing consistent descriptive outcomes under the fixed local-fault schedule.

\begin{figure}[htbp]\centering
\includegraphics[width=\linewidth]{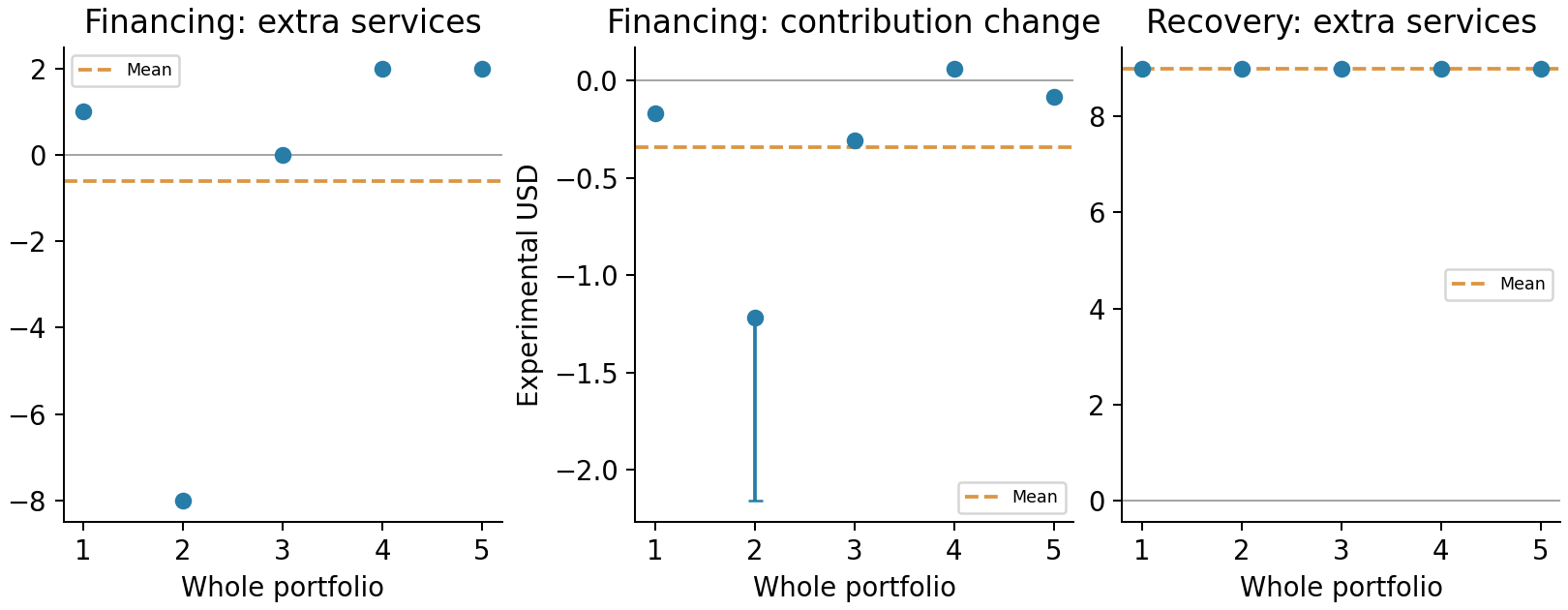}
\caption{Every observed whole-portfolio contrast. Whiskers bound unresolved usage under the reservation limit; dashed lines are known-expense means, not task-level confidence estimates.}\label{fig:robust-repeats}
\end{figure}

\clearpage
\subsubsection{Income, Capital, and Receipt-Delay Sensitivity}
Seven cells change one factor relative to the first repeated portfolio:
receipt multiples 1/1.5/3 (baseline 2), own-capital multiples 0.5/1.5
(baseline 1), or receipt delays 1/6 tasks (baseline 4). Additional investor
capital stays fixed. All tasks are executed again, so funds affect admission
and completion rather than being applied to copied successes. The same task
order is used across boundary cells. Each cell is one descriptive scenario
portfolio. Contribution excludes initial capital, and outstanding rights
are accounted for separately.
\begin{table}[!htbp]\centering\small
\caption{Fresh financing boundaries: completed service counts and contribution differences. The profit difference compares revenue rights with own capital. Equal funding isolates capital availability from the investor share.}\label{tab:robust-boundaries}
\begin{tabular}{@{}lrrrrr@{}}\toprule
Condition & Own & Equal & Right & $\Delta$ profit & Right--equal\\\midrule
Baseline & 11 & 12 & 12 & -0.1670 & -0.3065\\
Receipt multiplier 1 & 10 & 12 & 12 & -0.1826 & -0.2081\\
Receipt multiplier 1.5 & 11 & 12 & 12 & -0.2085 & -0.2298\\
Receipt multiplier 3 & 11 & 12 & 12 & -0.2487 & -0.4597\\
Capital multiplier 0.5 & 0 & 11 & 11 & 1.1919 & -0.2277\\
Capital multiplier 1.5 & 12 & 12 & 12 & -0.3065 & -0.2516\\
Delay (tasks) 1 & 12 & 12 & 12 & -0.3082 & -0.3065\\
Delay (tasks) 6 & 10 & 12 & 12 & -0.1478 & -0.3065\\
\bottomrule\end{tabular}\end{table}

At the 0.5 capital multiplier, financing completes eleven additional services and raises contribution by 1.1919 experimental USD. This is the one of eight displayed conditions with a positive lower-bound contrast. The scenario grid identifies the dependence on capital, receipts, and delay; the subsequent matched replication evaluates the low-capital condition across five portfolios.

\begin{figure}[!htbp]\centering
\includegraphics[width=\linewidth]{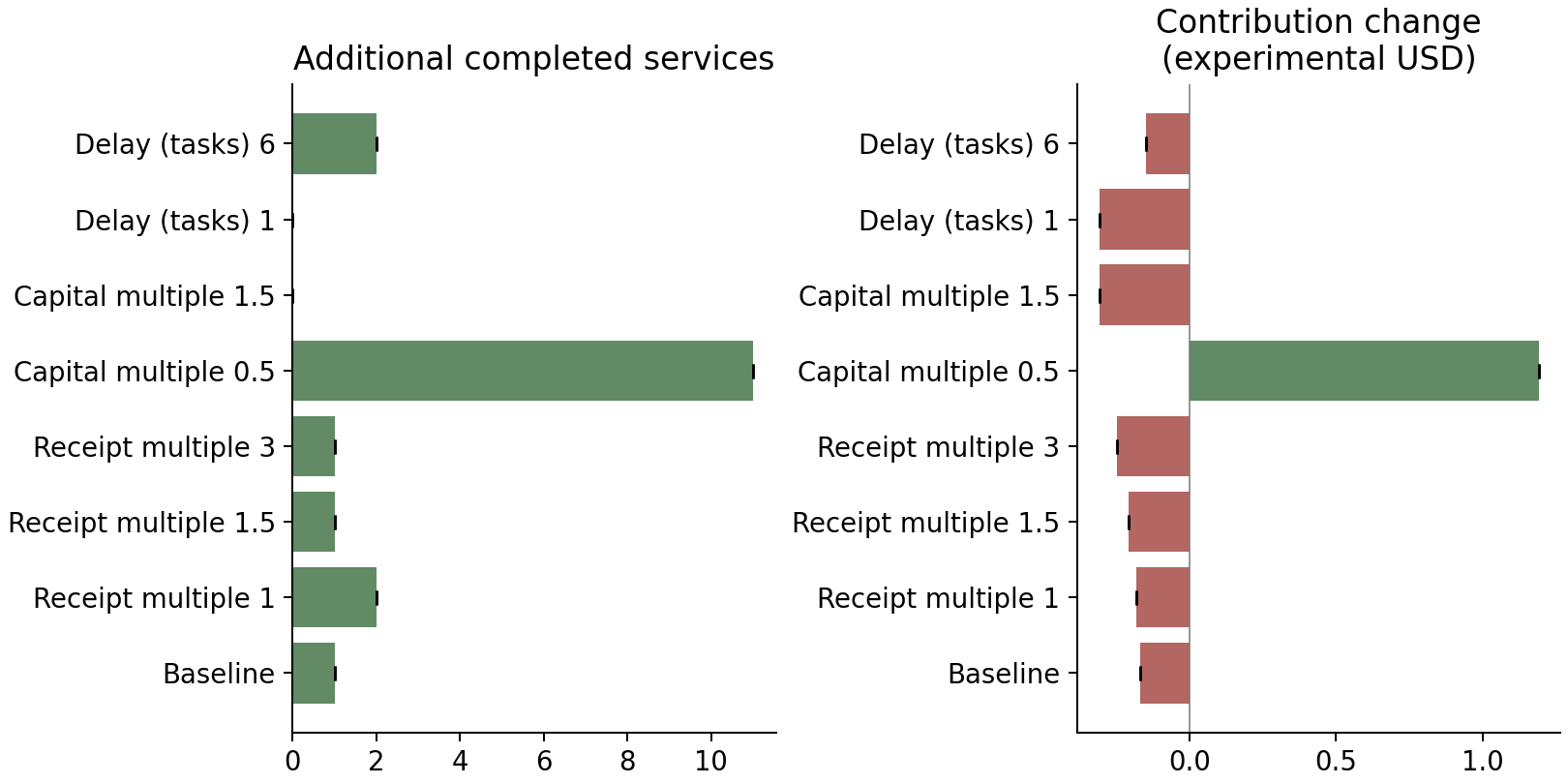}
\caption{Financing can change service completion and retained contribution differently. Each bar is a fresh scenario comparison.}\label{fig:robust-boundaries}
\end{figure}

\subsubsection{Initial Controlled Financing Study}
\begin{table}[htbp]\centering\small
\caption{Fresh financing executions; 12 scheduled tasks per arm. All monetary values are experimental USD. Profit excludes capital injection and uses known expense.}\label{tab:controlled-finance}
\begin{tabular}{@{}lrrrrr@{}}\toprule
Arm & Success & Budget stop & Expense & Retained & Profit\\\midrule
Own capital & 9 & 3 & 1.1692 & 2.2754 & 1.1062\\
Equal capital & 12 & 0 & 1.4749 & 3.0339 & 1.5590\\
Revenue right & 12 & 0 & 1.4749 & 2.7274 & 1.2525\\
\bottomrule\end{tabular}\end{table}
Relative to own capital, the revenue-right arm completes +3 additional services and changes known-cost operator contribution by 0.1463 experimental USD. The equal-capital arm identifies the effect of the same capital availability without the investor's revenue share. These realized portfolio contrasts quantify the additional service and retained contribution under the initial study's capital, tariff, and receipt schedule. The equal-capital comparison separates this capital effect from the investor allocation.

The investor pays 0.5200 and receives 0.3000 in wallet transfers, with 0.3065 accrued distribution. Rounding remainders not yet credited to wallets are tracked separately.
Cash flows use the observation-window convention, with outstanding
rights reported separately.

\begin{figure}[htbp]\centering
\includegraphics[width=\linewidth]{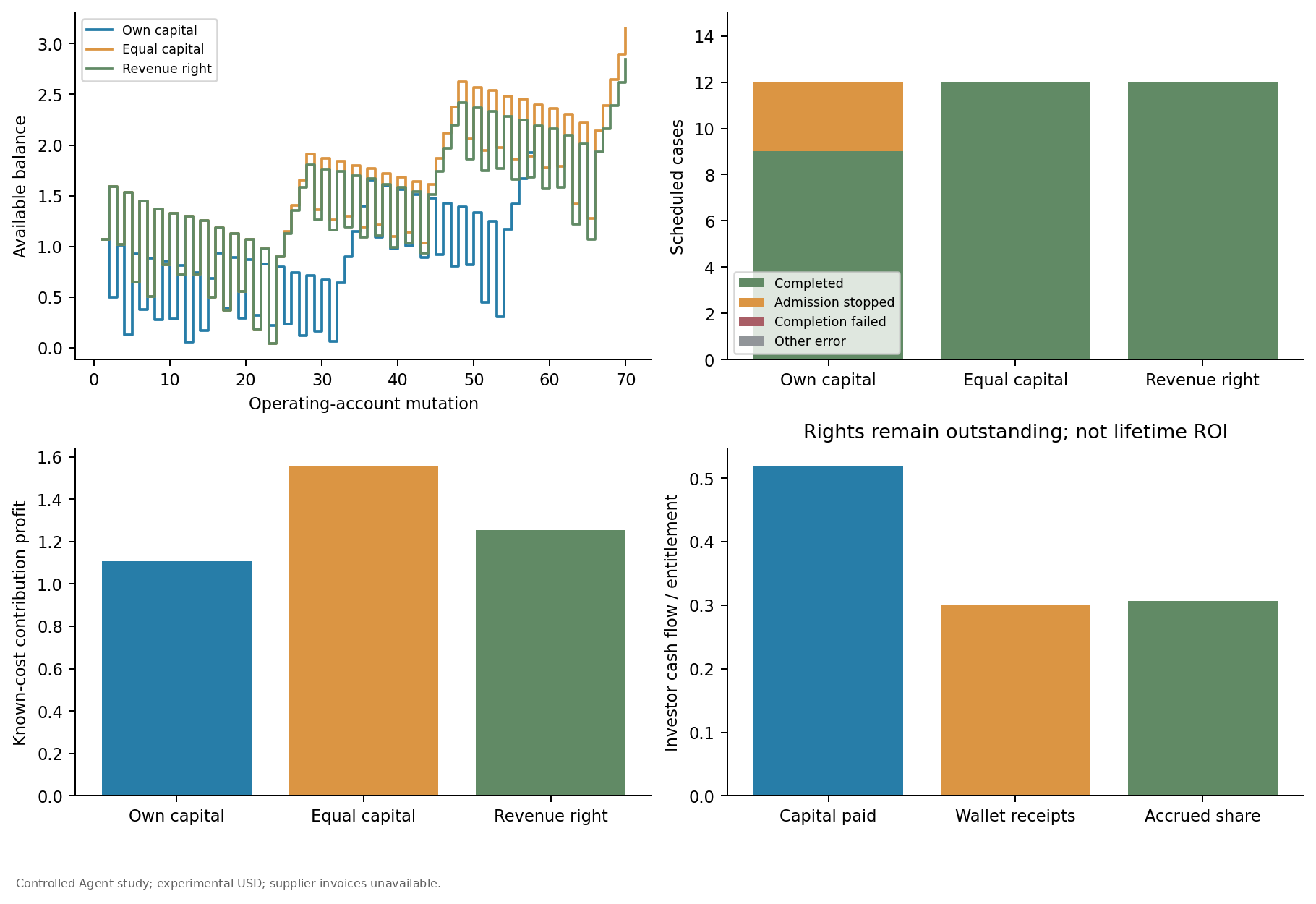}
\caption{Continuous operating funds, completed services, contribution and investor cash flow. Source: the controlled cohort's reconciled ledger and task records.}\label{fig:controlled-finance}
\end{figure}
\subsubsection{Retail Financing Interventions}
The own-capital control, equal-capital control, and revenue-right arm receive
$B$, $B+A$, and $B+A$ respectively. Only the last arm transfers $A$ from an
investor's separate research Billing Wallet through an approved investment.
Its contractual terms allocate 89\% to the operator, 10\% to investors,
and 1\% to the platform. Revenue-split execution is validated by the
separate contract tests; this retail batch records no success-triggered
revenue event. Controls pay the same 1\% platform share. Thus the equal-capital control
holds initial available funds equal when comparing revenue sharing.
Each arm executes fresh tasks because admission changes subsequent execution.
Table~\ref{tab:live-financing} reports task outcomes, known inference expense
and retained reservations. Experimental contribution profit is retained
hypothetical receipts minus known inference expense;
unknown-request usage remains outside this measured amount. Ending operating
balances also subtract retained uncertain-request reservations.

\begin{table}[tb]\centering\small
\setlength{\tabcolsep}{4pt}
\caption{Fresh financing interventions. Money is in experimental USD; receipts and tariffs are assumptions. Capital injections are excluded from contribution profit.}
\label{tab:live-financing}
\begin{tabular}{@{}lrrrrr@{}}\toprule
Arm & Scored & Successes & Budget stops & Known cost & Held funds\\\midrule
own & 1 & 0 & 9 & 0.147368 & 0.000000\\
equal & 4 & 0 & 6 & 0.515037 & 0.000000\\
right & 0 & 0 & 9 & 0.342738 & 0.172938\\
\bottomrule\end{tabular}\end{table}
\begin{figure}[tb]\centering
\includegraphics[width=\linewidth]{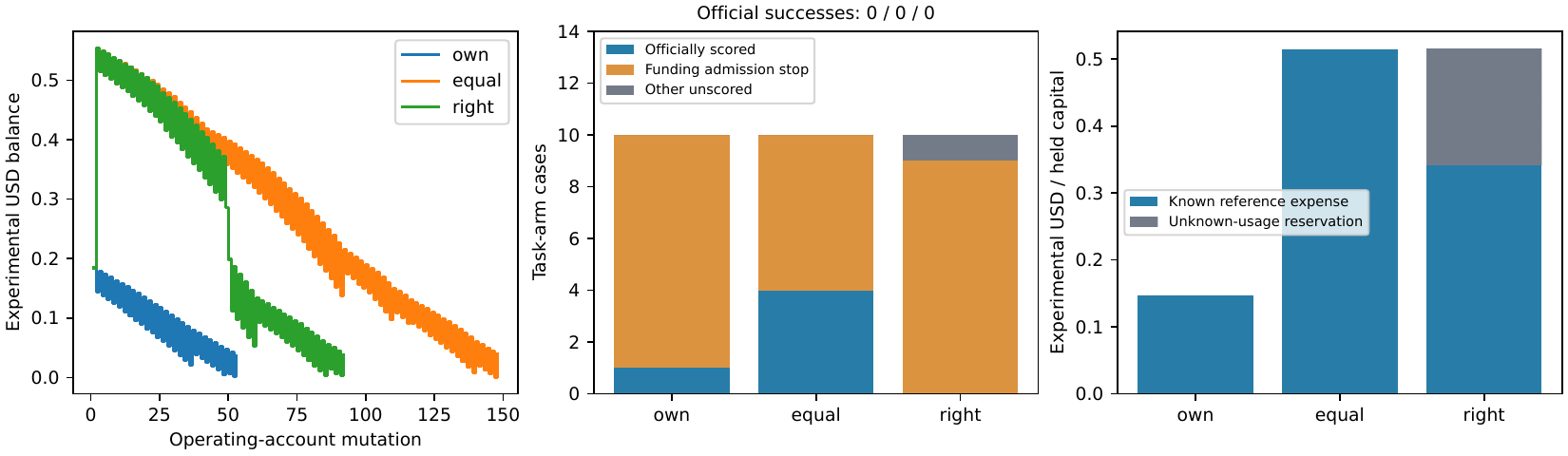}
\caption{Operating balances, admission and scoring outcomes, and measured
expenditure across financing conditions. Known expense and unresolved
request reservations are shown separately.}
\label{fig:live-financing}\end{figure}

The investor wallet transfers 0.370000 to the operator and receives
0.000000 during observation; its revenue rights remain outstanding.
The three financing arms record zero successful services in this fixed
retail sequence. Thus, the cohort characterizes admission, expenditure,
and capital retention before success-triggered revenue is realized.

\paragraph{Admission and receipt formation.}
Equal capital increases scoreable completions from one to four. The
revenue-right arm has no scored completion and retains 0.172938 after
two missing-response requests and a task-time limit. These reservations
enter later admission and are reported separately from incurred expense.
Known-cost contribution equals the negative of measured expense because
success-triggered receipts are zero. Varying assumed receipts over
1, 1.5, 2, or 3 times $\mu$ leaves this fixed trajectory's revenue
unchanged. Cross-arm differences include model variability and path
dependence, with unresolved expense bounded separately.

\subsubsection{Analytical Financing Comparisons}
\paragraph{Sharing costs and break-even growth.}
For the same completed workload, let $R$ be assumed receipts generated only by
successful tasks and $C$ its conditional operating cost. Own-capital and
equal-capital controls yield $R-C$, while a revenue share $\alpha$ yields
$(1-\alpha)R-C$ before additional fees. The allocation assigns $\alpha R$ to the investor while excluding
capital injections from contribution. The assumed receipt grid spans
0.05, 0.10, 0.25, 0.50, 1 and 2 per successful task with $\alpha=0.1$.
The post-hoc analysis gives the additional successes needed to offset
that allocation under stationary cost per success. A finite break-even
threshold requires a positive retained incremental margin. This analytical
threshold complements the freshly executed capital-constrained comparisons
against both controls.

\begin{figure}[tb]\centering
\includegraphics[width=\linewidth]{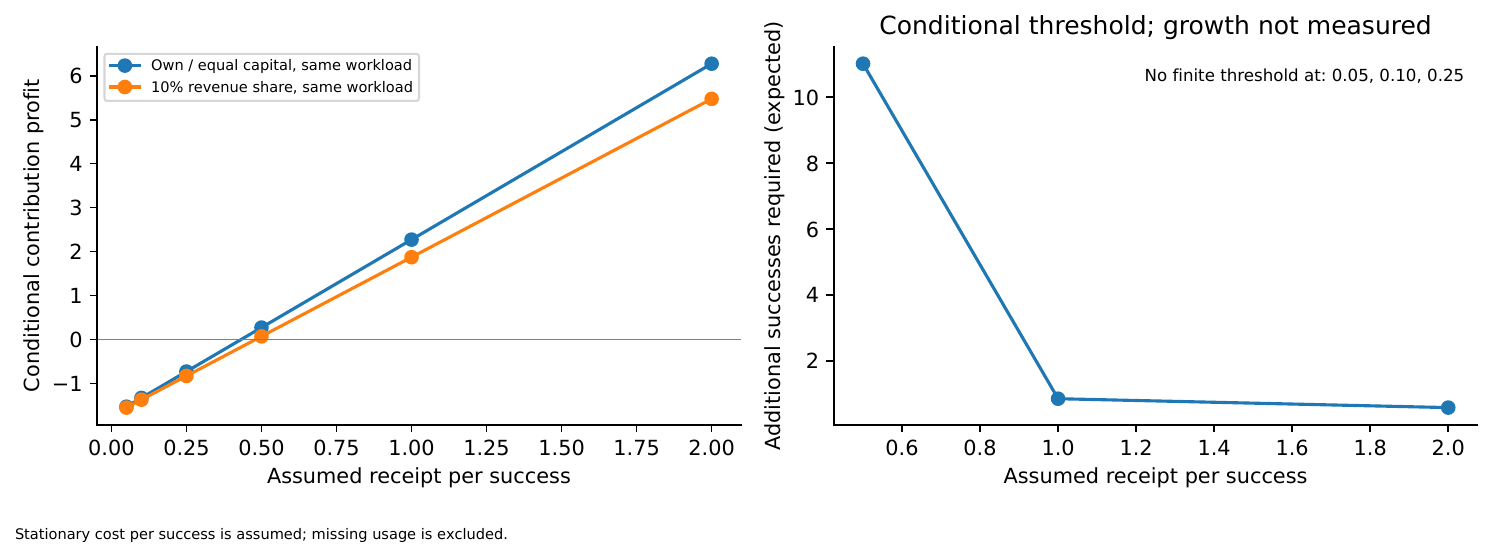}
\caption{Same-workload financing cost and conditional growth requirement.
The right panel gives the analytical break-even requirement under
stationary cost per success.}
\label{fig:agent-financing-effectiveness}\end{figure}

\paragraph{Additional funding under constrained demand.}
The financing replay admits complete requests in their original daily order,
stopping when the next request would exceed the budget. It does not select
later, cheaper requests. Initial daily budgets are the calibration invoice's
50th, 75th, and 90th percentiles: \$3.05370, \$41.65998, and \$166.25893.
Budgets reset daily; costs are known retrospectively from the trace. This
experiment evaluates the consequence of a larger budget, not an online
admission policy or a causal estimate of demand created by investment.

The Agent charges a 50\% markup over its API expense, and the Provider's
variable cost is 60\% of its API revenue. Financing assigns 10\% of each
publisher's revenue to investors and 2\% to the platform; the self-funded
baseline has no modeled revenue sharing. Contribution profit is the revenue
remaining after these shares and modeled variable costs, excluding fixed
costs and the initial capital contribution.

\begin{table}[htbp]
\centering
\small
\setlength{\tabcolsep}{4pt}
\caption{Contribution-profit change when financing doubles the daily budget.
Each row compares the same demand at the stated initial budget with and
without financing. Full requests are preserved.}
\label{tab:sharing}
\begin{tabular}{@{}lrrr@{}}
\toprule
Initial budget & Budget (\$/day) & Agent (\%) & Provider (\%)\\
\midrule
50th percentile & 3.05370 & +21.23 & +32.59\\
75th percentile & 41.65998 & -8.11 & +0.50\\
90th percentile & 166.25893 & -19.17 & -11.59\\
\bottomrule
\end{tabular}
\end{table}

Table~\ref{tab:sharing} shows that the benefit depends on the initial capital
constraint. At the lowest budget, additional service offsets revenue sharing;
at higher budgets, insufficient unserved demand remains. At the lowest
budget, the matched approximation permitting fractional requests yields
21.17\% for the Agent, compared with 21.23\% for complete requests. The
observed reversal is therefore primarily associated with available capital,
not request rounding. With no added service, the same sharing terms reduce
Agent and Provider contribution profit by 36\% and 30\%, respectively.
Under the stated margins, breaking even requires revenue growth above
56.25\% for the Agent and 42.86\% for the Provider.

\begin{figure}[htbp]
\centering
\includegraphics[width=\linewidth]{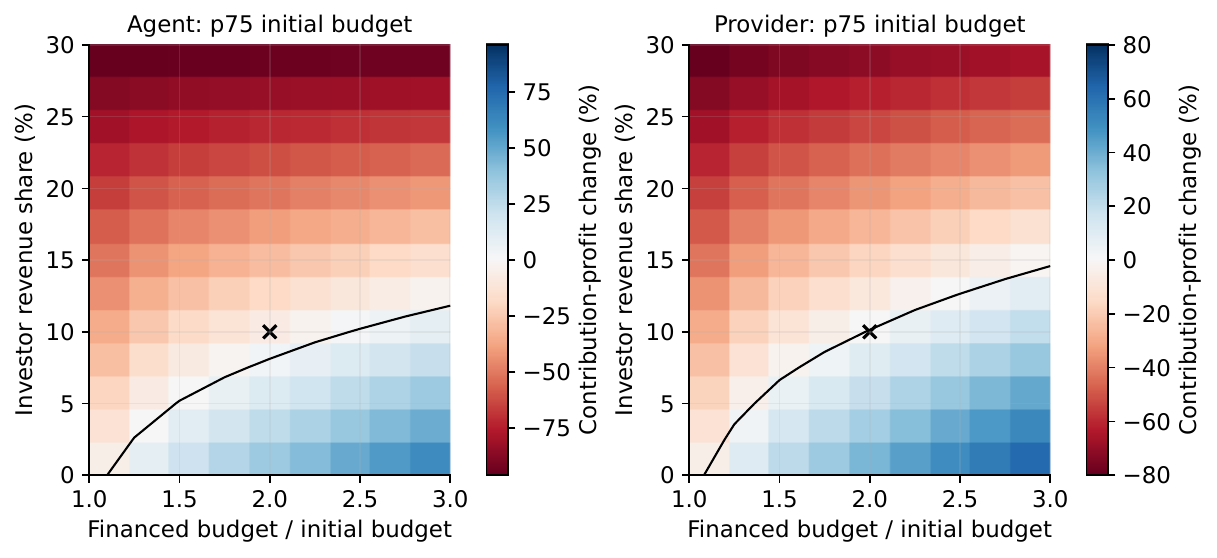}
\caption{Contribution-profit changes at the 75th-percentile initial budget,
varying additional funding and investor share. Zero-change contours show
where additional service revenue offsets sharing for each publisher.}
\label{fig:profit}
\end{figure}

\paragraph{Equal-funded-amount alternatives.}
To separate funding availability from funding charges, a second comparison
holds the total funded amount at \$1,710.86. Half of each day's positive
cash requirement is assigned 30 days of financing. Assumed charges are
12\% annual opportunity cost for own capital---the modeled cost of committing
the operator's own funds---12\% interest plus a 1\% initial fee for fixed
credit, 8\% interest for Provider credit, and 10\% of modeled operating
receipts plus a 1\% initial fee for revenue sharing. Receipts are 1.5 times
the primary API invoice. Resulting charges are \$16.87, \$33.98, \$11.25,
and \$526.48, respectively. The funded total is the sum of financing
amounts, not peak outstanding debt.

This comparison isolates the stated charging rules at equal funded volume.
Repayment obligations, collateral, and default risk are outside its modeled
variables. Together, the growth and charge comparisons characterize
additional service under capital scarcity and the cost of specified terms
at fixed volume. Investor transfers use the observation-window convention;
secondary-market valuation is a separate analysis.
\FloatBarrier

\subsection{Protection: Execution Evidence and Analytical Reserve Sensitivity}
\label{app:eval-protection}
The main text reports the fault-frequency, premium, and reserve study.
The initial controlled and retail interventions below report contract
eligibility, approval, and settlement. The final subsection develops an
analytical reserve-sensitivity study with its own stated scenario parameters.
Alternative policy wallets are evaluated separately, with each wallet's
payments reported for the corresponding policy.

\subsubsection{Controlled Recovery and Loss Allocation}
\begin{table}[htbp]\centering\small
\caption{Twelve scheduled cases per fault arm. Reached counts injected incidents. Funded and scarce columns are actual service-credit payments, not cash income.}\label{tab:controlled-protection}
\begin{tabular}{@{}lrrrrr@{}}\toprule
Arm & Success & Reached & Loss & Funded & Scarce\\\midrule
Own / no recovery & 0 & 12 & 0.5600 & 0.5600 & 0.1400\\
Own / recovery & 9 & 12 & 0.1400 & 0.1400 & 0.1400\\
Right / no recovery & 0 & 12 & 0.5400 & 0.5400 & 0.1200\\
Right / recovery & 9 & 12 & 0.1400 & 0.1400 & 0.1400\\
\bottomrule\end{tabular}\end{table}
Local recovery achieves 9/12 successful services in both recovery-enabled
arms through alternate paths to the same Provider. Funded policies settle
all approved compensation, totaling 1.3800 in service credits across the
four arms. Scarce-reserve policies pay 0.5400 and leave 0.8400 approved
but unpaid. Eligibility is restricted to covered terminal injected
failures; task execution and operating capital remain fixed within each
protection comparison. Claims, reserves, and wallet credits are reconciled
separately, and each payment passes a whole-claim funding check.

\begin{figure}[htbp]\centering
\includegraphics[width=\linewidth]{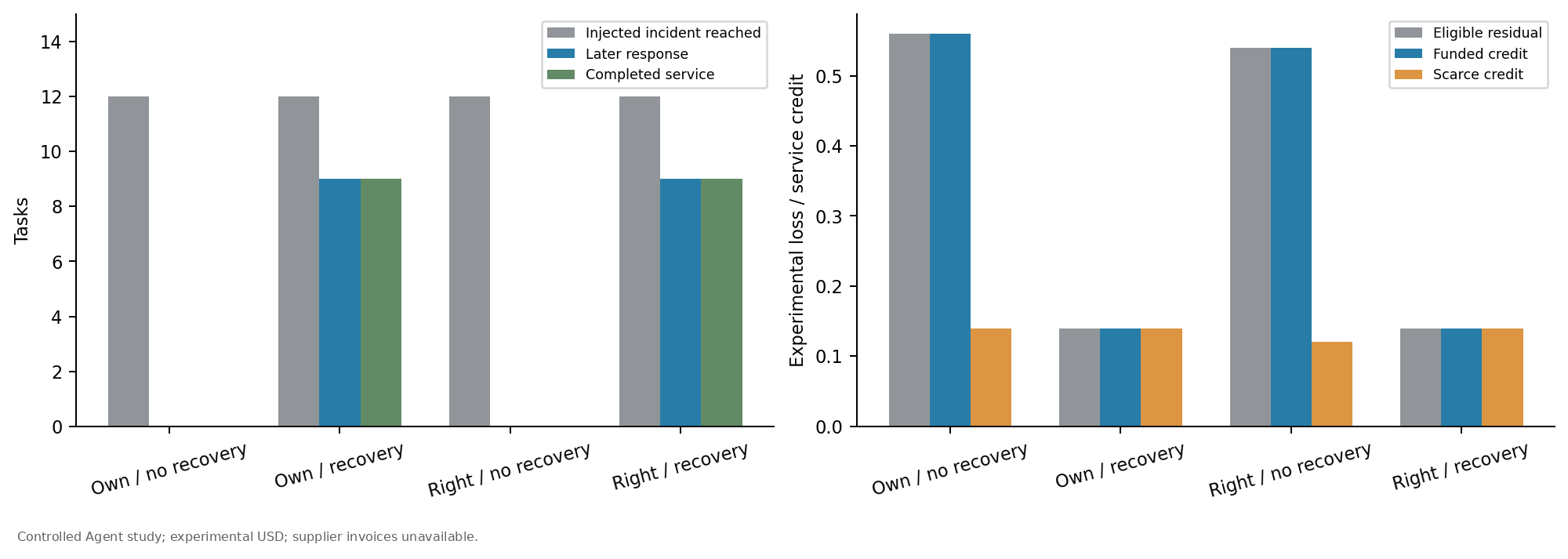}
\caption{Completed services and reserve-funded loss allocation under controlled local faults. Monetary values use the stated laboratory tariff.}\label{fig:controlled-protection}
\end{figure}
\subsubsection{Retail Recovery and Protection Interventions}
Four fresh fault arms cross financing off/on with recovery off/on. They start
with ample operating capital ($100\mu$, plus $A$ for financing) to separate
recovery from capital scarcity; any admission failure is still reported.
The schedule repeats timeout, rate limit, primary-route outage, common-route
outage, and no fault across ten tasks. Injections require the third Agent call
to be reached. For transient or primary-only adapter faults, recovery permits
a real continuation request to the same configured Nexus/SiliconFlow endpoint;
common-route outages terminate the task. Recovery is a controlled continuation
branch at the same configured endpoint. The model, supplier, and bounded
transport-retry settings remain identical across arms; the recovery switch
controls only the injected local incident.
Post-incident responses, scoreable completions, official task success,
and inference expense are reported as separate outcomes.

Protection fees and payouts use a separate research wallet, leaving
operating capital fixed. Accordingly, S=0 and S=1 share the same execution
trajectory. Terminal injected failures may generate a
claim for known incurred experimental Agent expense, rounded down to cents;
wrong answers, missing usage, and infrastructure errors do not automatically
qualify. Claims pass actual coverage review and reserve checks. The funded
and scarce-reserve plans share per-claim terms but use separate financial
accounts. Whole-claim settlement requires sufficient remaining reserve budget
and available account balance. Payouts are
Billing Wallet service credits, reported separately from cash.

\begin{table}[tb]\centering\small
\setlength{\tabcolsep}{4pt}
\caption{Retail fault interventions and service-credit settlements.
Off/on denotes local recovery. Eligible loss excludes semantic task errors
and unknown usage.}
\label{tab:live-protection}
\begin{tabular}{@{}lrrrrrr@{}}\toprule
Arm & Successes & Faults & Residual cost & Eligible & Funded paid & Scarce paid\\\midrule
Own / off & 0 & 8 & 0.356410 & 0.080000 & 0.080000 & 0.030000\\
Own / on & 1 & 8 & 1.511219 & 0.020000 & 0.020000 & 0.020000\\
Right / off & 0 & 8 & 0.426050 & 0.080000 & 0.080000 & 0.030000\\
Right / on & 0 & 8 & 1.063093 & 0.020000 & 0.020000 & 0.020000\\
\bottomrule\end{tabular}\end{table}

\paragraph{Recovery and coverage have different observed effects.}
Each recovery-enabled arm encounters eight scheduled incidents: six permit a
returned post-incident Agent response and two common incidents terminate the
task. Both recovery-disabled arms terminate at all eight incidents. This
measures execution continuation under the controlled adapter policy.
Official successes change from zero to one in the own-capital pair and
remain zero in the revenue-right pair. Known Agent expenses are
0.356410/1.600688 and 0.426050/1.063093 for recovery off/on, respectively,
providing the execution-cost component of the financial comparison.

Without recovery, each funded plan pays eight claims totaling 0.08 service
credits. Its scarce-reserve counterpart pays 0.03 and leaves five approved
claims totaling 0.05 unpaid; no partial settlement occurs. With recovery,
both reserve conditions pay the two common-incident claims totaling 0.02 per
arm. The 0.01 premium is paid once per plan and also increases reserve budget.
These matched plans quantify how reserve funding and local recovery
jointly determine paid compensation. Coverage is limited to the specified
terminal injected failures; semantic task-error expense is reported
separately under the controlled protocol.
\begin{figure}[tb]\centering
\includegraphics[width=\linewidth]{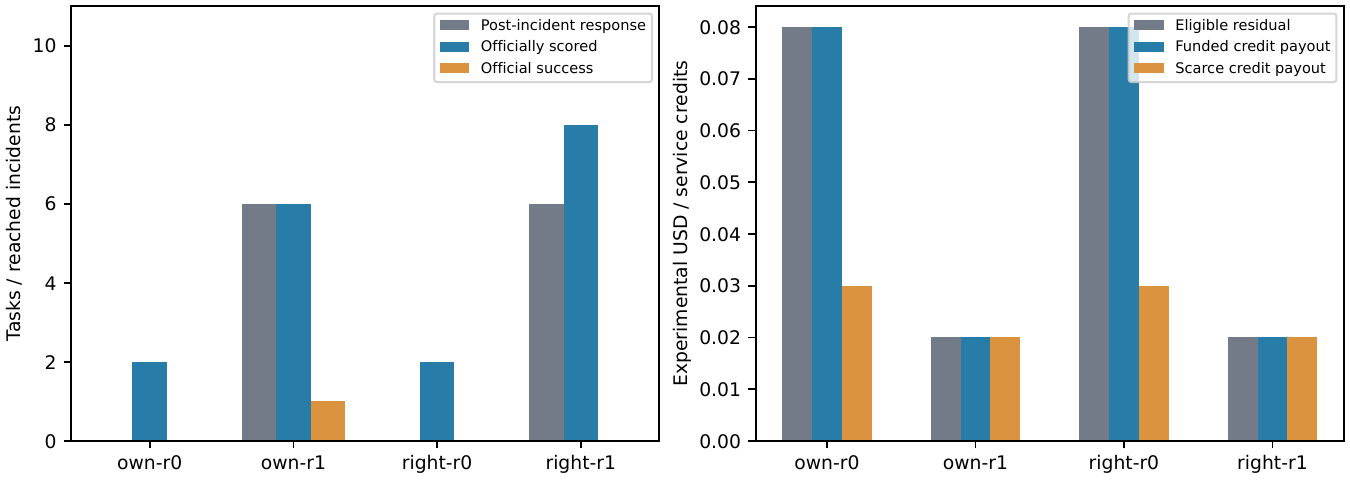}
\caption{Post-incident responses, scored completions, official task success,
and redistribution of covered residual expense.
Recovery changes executions; protection transfers service-credit value.
Amounts are reconciled across task outcomes, claim decisions, reserve
accounts, and service-credit records.}
\label{fig:live-protection}\end{figure}

\paragraph{Economic interpretation.}
Under the explicit assumption that issued service credits are usable at face
value, let $L$ be total residual economic loss after recovery, $p$ the premium
and $X$ the actual payout for its covered and eligible portion, the operator bears $L+p-X$ and the protection provider bears
$X-p$ before administration costs. Their combined loss remains $L$: protection
transfers economic loss, while recovery changes its occurrence. This
participant-level conservation relation connects the conditional economic
analysis to the separately reconciled intervention outcomes.

\subsubsection{Analytical Reserve Sensitivity}
\paragraph{Recovery and portfolio-level coverage.}
Protection comparisons use identical retry settings. Primary failures use
zero-output trace records; alternate failures use the calibration rate
of 0.2190846\%. A parameter $\rho$ specifies the fraction of primary failures
attributed to a persistent shared cause that alternate attempts cannot
remove. The remaining failures use the assumed alternate failure rate.
These parameters define the analytical shared-failure and recovery scenario.

In the fixed-volume control with $\rho=0.75$, one alternate attempt costs
\$25.91 and increases expected successful API responses from 883,304 to
887,329.41. Adding financial protection leaves that count unchanged by
construction. The modeled plan charges a 1\% premium, meaning the protection
fee, and uses an initial reserve and total liability cap each equal to 10\%
of the calibration-predicted 42-day invoice. Claims become due after seven
days, unpaid amounts carry forward, and replay continues through day 49.
At \$0.10 loss per unresolved call, the pool collects \$34.22 and pays
\$220.73. The \$186.51 reduction in buyer expense is matched by payments
exceeding protection fees by the same amount, before reserve capital and
operating costs. This reserve-funded transfer is evaluated at fixed
execution, separating loss allocation from computational expenditure.

\paragraph{Reserve sufficiency and claim timing.}
With $\rho=0.75$, a 1\% fee, and no claim delay, an initial reserve equal
to 10\% of the predicted invoice funds all capped claims. This result
identifies a sufficient capitalization level for the stated scenario.
At 5\%, the first shortfall occurs on day 10 and \$76.41 remains unpaid;
with zero initial reserve, \$186.77 remains unpaid. The chronological
sensitivity varies initial reserves, fees, and claim lags of zero, seven,
or thirty days, carrying unpaid obligations beyond the final request.
The liability cap is fixed from calibration. Initial capital and premium
receipts are tracked separately in the reserve accounting.

\begin{figure}[htbp]
\centering
\includegraphics[width=\linewidth]{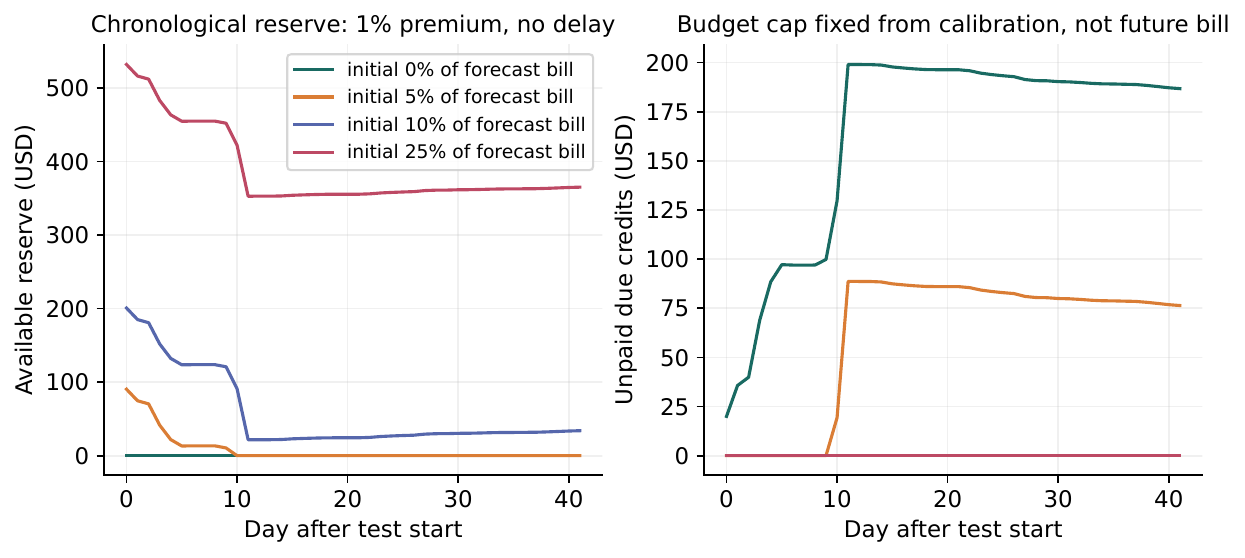}
\caption{Reserve sufficiency under different initial-capital levels, with
$\rho=0.75$, a 1\% protection fee, and no claim delay. Panels report
available reserves and unpaid capped claims; initial capital is accounted
for separately from premium receipts.}
\label{fig:reserve}
\end{figure}

The analytical estimates are conditional on the portfolio-level cap,
claim-lag schedule, and carried-forward unpaid obligations specified above.
Zero-output records supply the failure indicator; claim eligibility and
credit utilization are scenario assumptions. Compensation reduces modeled
economic expense under the assumption that the corresponding credit is
usable at its recorded amount.
\FloatBarrier

\subsection{Combined Financial Mechanisms}
\label{app:eval-combined}
Section~\ref{sec:eval-combined} reports the controlled combinations. The retail
and fixed-volume request comparisons below retain their own execution,
pricing, and valuation assumptions. Forward and protection settlements
are applied with the associated operational-admission trajectory held fixed.

\paragraph{Retail workload: measured execution with paired financial combinations.}
The eight D/F/S cells reuse the recovery-enabled own-capital and revenue-right
executions for F=0 and F=1. Financing has fresh executions; D and S settle in
separate wallets after or outside operational admission. At the declared 1.5
price scenario and $h=0.5$, Table~\ref{tab:live-combined} reports conditional
contribution excluding credits,
$R_{\rm retained}-1.5C+P_{\rm fwd}-f-p$, where $P_{\rm fwd}$ is the signed
forward payment, $f$ the forward fee, and $p$ the protection premium.
Service-credit compensation is reported separately from cash contribution.
The scenario price is applied after execution with admission fixed, so the
eight cells characterize paired financial accounting on the two observed
financing trajectories. Each contribution is calculated from its component
cash flows under the same scenario.

\begin{table}[tb]\centering\small
\setlength{\tabcolsep}{4pt}
\caption{Eight combinations: actual task outcomes and contract payments under an assumed price scenario.}
\label{tab:live-combined}
\begin{tabular}{@{}lrrr@{}}\toprule
DFS & Successes & Known-cost contribution & Service credit\\\midrule
000 & 1 & -2.036122 & 0.000000\\
001 & 1 & -2.046122 & 0.020000\\
100 & 1 & -1.586435 & 0.000000\\
101 & 1 & -1.596435 & 0.020000\\
010 & 0 & -1.594640 & 0.000000\\
011 & 0 & -1.604640 & 0.020000\\
110 & 0 & -1.144952 & 0.000000\\
111 & 0 & -1.154952 & 0.020000\\
\bottomrule\end{tabular}\end{table}

At the rising index, D adds a net 0.449687 after fees, while S exchanges
a 0.01 premium for 0.02 in service credits. All eight known-cost
contributions remain negative, as reported in Table~\ref{tab:live-combined}.
Within each financing trajectory, the paired contrasts isolate these
payment components. Cross-financing contrasts also include differences
in task outcomes, measured usage, and unresolved reservations. The common
accounting representation preserves these distinct contributions to the
combined outcome.

\paragraph{Fixed-volume request replay.}
The eight $D/F/S$ combinations cross five price paths and two forward
discounts, yielding 80 configurations with the same requests and recovery
policy. Funding uses modeled own-capital opportunity cost when $F=0$ and the
stated revenue-sharing charges when $F=1$. Economic expense is defined by
Equation~\eqref{eq:eval-economic-cost}; it is distinct from a commercial cash
invoice or the real-model contribution excluding service credits.

\begin{table}[htbp]
\centering
\small
\setlength{\tabcolsep}{4pt}
\caption{Fixed-volume comparison under constant prices. Codes indicate
whether $D$, $F$, and $S$ are enabled.}
\label{tab:eval-combined}
\begin{tabular}{@{}llr@{}}
\toprule
Configuration & Forward discount & Economic expense (\$)\\
\midrule
000: baseline & --- & 3,438.59\\
100: forward only & 0\% & 3,459.88\\
100: forward only & 10\% & 3,282.46\\
111: all three & 10\% & 3,605.09\\
\bottomrule
\end{tabular}
\end{table}

Under fixed-volume demand, the assumed 10\%-discount forward reduces
economic expense from \$3,438.59 to \$3,282.46. The all-mechanism
configuration has expense \$3,605.09, or 4.84\% above baseline
(Table~\ref{tab:eval-combined}), because financing and protection charges
are not accompanied by additional admitted work. The decomposition
separates individual-contract terms and combination-dependent effects;
these sum to the total 111-minus-000 difference. This fixed-volume
control complements the capital-constrained growth experiments by
identifying the demand condition needed to offset financing charges.
\FloatBarrier

\section{Supplementary Execution Cohorts and Measurement Accounting}
\label{app:qwen-development}

This appendix records preliminary execution cohorts separately from the
retail evaluation. The cohorts use distinct configurations or repeat
previously inspected tasks. Their configuration, completion counts, and
available usage records establish the provenance of the supplementary
measurements; they are excluded from evaluation denominators.

\subsection{Preliminary Retail Cohorts}
Table~\ref{tab:integration-diagnostics} summarizes the preliminary retail
cohorts and the scope of their recorded outcomes. A scored trial has an
official task outcome. An unscored trial retains its execution records
without an imputed task score. Unattempted scheduled trials are recorded
separately. Final evaluation task selection remains unchanged.

\begin{table}[htbp]\centering\small
\caption{Preliminary retail cohorts and recorded measurement scope.
These cohorts are separate from the final baseline estimate.}
\label{tab:integration-diagnostics}
\begin{tabular}{@{}p{.25\linewidth}p{.66\linewidth}@{}}
\toprule Configuration & Recorded measurement scope\\\midrule
Direct-provider pilot & Ten scored trials and no successful tasks.\\
Gateway-format pilot & Ten Agent requests and ten returned simulated-customer
responses; the customer responses match gateway records.\\
Authenticated pilot & Three scored trials with no successes, three unscored
trials, and four unattempted scheduled trials; all 181 returned responses
reconcile with gateway records.\\
Fixed-schedule pilot & Two scored trials, including one success.\\
Recovery configuration pilot & Two scored trials, including one success.\\
Shortened pilot & No completed trial.\\
Local-host pilot & Forty-nine returned model responses; all ten scheduled
trials remain unscored.\\\bottomrule
\end{tabular}\end{table}

The corresponding request records and configuration details are retained
in the supplementary artifact. Preliminary observations are reported at
their own cohort scope and are not pooled with completed evaluation trials.

\subsection{Grounding and Reasoning Configurations}
Four additional batches reuse the same two retail calibration tasks,
retaining every scheduled attempt. Both the Agent and simulated customer
use Qwen/Qwen3-8B through Nexus and SiliconFlow. The official retail tools
and ALL completion criterion remain unchanged. Additional Agent and
customer instructions define a custom execution protocol.

\begin{table}[htbp]\centering\small
\setlength{\tabcolsep}{5pt}
\caption{Supplementary configuration batches, each scheduling two tasks.
Request counts include all model roles; scored tasks have an official
outcome. Returned usage is accounted for separately from task scores.}
\label{tab:qwen-development}
\begin{tabular}{@{}lrrrr@{}}\toprule
Configuration & Requests & Responses & Scored & Successes\\\midrule
Grounding & 53 & 52 & 2 & 0\\
Reasoning, 60\,s timeout & 6 & 0 & 0 & 0\\
Reasoning, 240\,s timeout & 6 & 0 & 0 & 0\\
Reasoning, final batch & 21 & 21 & 2 & 0\\
\bottomrule\end{tabular}\end{table}

The grounding configuration adds general reminders about identifier
provenance and customer-scenario fidelity. It uses temperature zero,
a 600-second task deadline, and 30 requests per role. The reasoning
configuration uses temperature 0.6, a 1,024-token reasoning budget,
a 2,048-token output setting, a 1,200-second task deadline, and 40 requests
per role. These are joint configuration comparisons rather than an isolated
reasoning-mode intervention. All 21 returned usage records in the final
batch match Nexus records exactly.

\subsection{Usage Accounting and Study Populations}
Across the four batches, 86 physical requests yield 73 complete responses,
four scored outcomes, and no successful tasks. Thirteen requests have no
returned usage, including one in the grounding batch. Returned usage and
unresolved request records are accounted for separately. Monetary
interpretation follows the experimental-valuation conventions in
Section~\ref{sec:eval-measurement-scope}.

The records distinguish model responses, tool execution, official task
outcomes, and financial settlement. The controlled study in
Appendix~\ref{app:controlled-protocol} uses fixed customer requests and
explicit operation-level completion rules. Its observations retain a
separate study population, as do the retail evaluation, calibration passes,
and supplementary batches reported here.
\FloatBarrier

\end{document}